# Analyzing Search Topology Without Running Any Search:
# On the Connection Between Causal Graphs and $h^+$

**Jörg Hoffmann**　　　　　　　　　　　　　　　　　　JOERG.HOFFMANN@INRIA.FR
*INRIA*
*Nancy, France*

## Abstract

The ignoring delete lists relaxation is of paramount importance for both satisficing and optimal planning. In earlier work, it was observed that the optimal relaxation heuristic $h^+$ has amazing qualities in many classical planning benchmarks, in particular pertaining to the complete absence of local minima. The proofs of this are hand-made, raising the question whether such proofs can be lead automatically by domain analysis techniques. In contrast to earlier disappointing results – the analysis method has exponential runtime and succeeds only in two extremely simple benchmark domains – we herein answer this question in the affirmative. We establish connections between causal graph structure and $h^+$ topology. This results in low-order polynomial time analysis methods, implemented in a tool we call TorchLight. Of the 12 domains where the absence of local minima has been proved, TorchLight gives strong success guarantees in 8 domains. Empirically, its analysis exhibits strong performance in a further 2 of these domains, plus in 4 more domains where local minima may exist but are rare. In this way, TorchLight can distinguish "easy" domains from "hard" ones. By summarizing structural reasons for analysis failure, TorchLight also provides diagnostic output indicating domain aspects that may cause local minima.

## 1. Introduction

The ignoring delete lists relaxation has been since a decade, and still is, of paramount importance for effective satisficing planning (e.g., McDermott, 1999; Bonet & Geffner, 2001; Hoffmann & Nebel, 2001a; Gerevini, Saetti, & Serina, 2003; Helmert, 2006; Richter & Westphal, 2010). More recently, heuristics making this relaxation have also been shown to boost optimal planning (Karpas & Domshlak, 2009; Helmert & Domshlak, 2009). The planners using the relaxation approximate, in a variety of ways, the optimal relaxation heuristic $h^+$ which itself is **NP**-hard to compute (Bylander, 1994). As was observed in earlier work (Hoffmann, 2005), $h^+$ has some rather amazing qualities in many classical planning benchmarks. Figure 1 gives an overview of these results.[1]

The results divide domains into classes along two dimensions. We herein ignore the horizontal dimension, pertaining to dead ends, for which domain analysis is already available: easy-to-test powerful criteria implying that a task is "undirected"/"harmless" are known (e.g., Hoffmann, 2005). The vertical dimension divides the domains into three classes, with respect to the behavior of exit distance, defined as $d - 1$ where $d$ is the distance to a state with strictly smaller $h^+$ value. In the "easiest" bottom class, there exist constant upper bounds on exit distance from both, states on local minima and states on benches (flat regions). In the figure, the bounds are given in square brackets. For example, in Logistics, the bound for local minima is 0 – meaning that no local minima exist at all – and the bound for benches is 1. In the middle class, a bound exists only for local minima; that bound is 0 (no local minima at all) for all domains shown. In the "hardest" top class, both local minima and benches may take arbitrarily many steps to escape.

---

1. We omit ADL domains, and we add the more recent IPC benchmarks Elevators and Transport (without action costs), for which these properties are trivial to prove based on the earlier results. Blocksworld-Arm is the classical blocksworld, Blocksworld-NoArm is a variant allowing to "move A from B to C" directly.

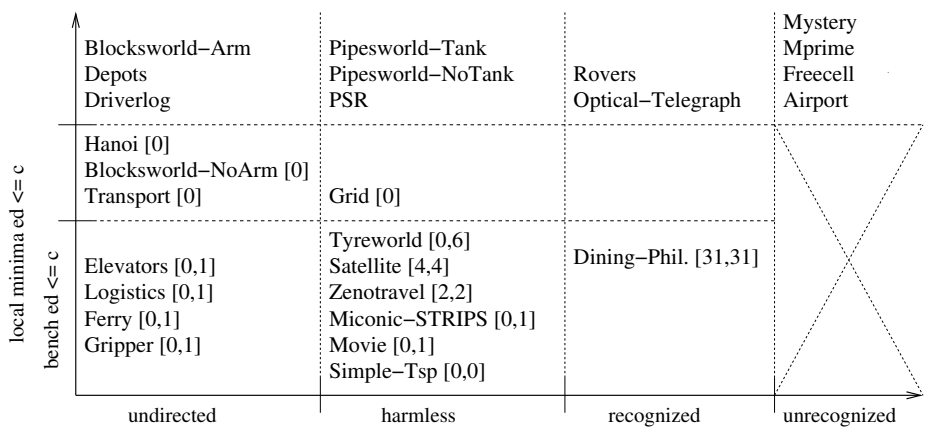

Figure 1: Overview of $h^+$ topology (Hoffmann, 2005).

The proofs underlying Figure 1 are hand-made. For dealing with unseen domains, the question arises whether we can design domain analysis methods leading such proofs automatically. The potential uses of such analysis methods are manifold; we discuss this at the end of the paper. For now, note that addressing this question is a formidable challenge. *We are trying to automatically infer properties characterizing the informativeness (or lack thereof) of a heuristic function.* We wish to do this based on a static analysis, not actually running any search. Formally characterizing the informativeness of a heuristic function is, in most cases, hardly possible even for experienced researchers, which explains perhaps why no-one so far has even attempted to do it automatically. The single exception, to the best of the author's knowledge, is an analysis method mentioned on the side in the author's earlier work (Hoffmann, 2005). This analysis method builds an exponentially large tree structure summarizing all ways in which relaxed plans may generate facts. The tree size, and therewith the analysis runtime, explodes quickly with task size. Worse, the analysis succeeds only in Movie and Simple-TSP – arguably the two most simplistic planning benchmarks in existence.[2]

By contrast, the TorchLight tool developed herein has low-order polynomial runtime and usually terminates in split seconds. Distinguishing between *global* (per task) and *local* (per state) analysis, it proves the global absence of local minima in Movie, Simple-TSP, Logistics, and Miconic-STRIPS. It gives a strong guarantee for local analysis – to succeed in every state – in Ferry, Gripper, Elevators, and Transport. Taking the *success rate* to be the fraction of states for which local analysis succeeds, TorchLight empirically exhibits strong performance – delivering high success rates – also in Zenotravel, Satellite, Tyreworld, Grid, Driverlog, and

---

2. Simple-TSP encodes TSP but on a fully connected graph with uniform edge cost. The domain was introduced by Fox and Long (1999) as a benchmark for symmetry detection.

Rovers. Thus TorchLight's success rates tend to be high in the "easy" domains of Figure 1, while they are low in the "hard" ones, serving to automatically distinguish between these two groups.[3] By summarizing structural reasons for analysis failure, TorchLight finally provides diagnostic output indicating problematic aspects of the domain, i.e., operator effects that potentially cause local minima under $h^+$.

What is the key to this performance boost? Consider Logistics and Blocksworld-Arm. At the level of their PDDL domain descriptions, the difference is not evident – both have delete effects, so why do those in Blocksworld-Arm "hurt" and those in Logistics don't? What does the trick is to move to the finite-domain variable representation (e.g., Jonsson & Bäckström, 1998; Helmert, 2006, 2009) and to consider the associated structures, notably the causal graph (e.g., Knoblock, 1994; Jonsson & Bäckström, 1995; Domshlak & Dinitz, 2001; Helmert, 2006) capturing the precondition and effect dependencies between variables. The causal graph of Blocksworld-Arm contains cycles. That of Logistics doesn't. Looking into this, it was surprisingly easy to derive the following basic result:

> *If the causal graph is acyclic, and every variable transition is invertible,*
> *then there are no local minima under $h^+$.*

This result is certainly interesting in that, for the first time, it establishes a connection between causal graph structure and $h^+$ topology. However, by itself the result is much too weak for domain analysis – of the considered benchmarks, it applies only in Logistics. We devise generalizations and approximations yielding the analysis results described above. Aside from their significance for domain analysis, our techniques are also interesting with respect to research on causal graphs. Whereas traditional methods (e.g., Jonsson & Bäckström, 1995; Brafman & Domshlak, 2003; Jonsson, 2009; Giménez & Jonsson, 2009a) seek execution paths solving the overall task, we seek "only" execution paths decreasing the value of $h^+$. In local analysis, this enables us to consider only small fragments of the causal graph, creating the potential to successfully analyze states in tasks whose causal graphs are otherwise arbitrarily complex.

The next section gives a brief background on planning with finite-domain variables, and the associated notions such as causal graphs and the definition of $h^+$ and its topology. Section 3 then gives an illustrative example explaining our basic result, and Section 4 provides a synopsis of our full technical results relating causal graphs and $h^+$ topology. Sections 5 and 6 present these results in some detail, explaining first how we can analyze a state $s$ provided we are given an optimal relaxed plan for $s$ as the input, and thereafter providing criteria on causal graph structure implying that such analysis will always succeed. We evaluate the domain analysis technique by proving a number of domain-specific performance guarantees in Section 7, and reporting on a large-scale experiment with TorchLight in Section 8. We point to related work within its context where appropriate, and discuss details in Section 9. We close the paper with a discussion of future work in Section 10. To improve readability, the main text omits many technical details and only outlines the proofs. The full details including proofs are in Appendix A.

---

3. To some extent, this particular result can also be achieved by simpler means (limited search probing). We discuss this along with the experiments in Section 8.

## 2. Background

We adopt the terminology and notation of Helmert (2006), with a number of modifications suiting our purposes. A (finite-domain variable) *planning task* is a 4-tuple $(X, s_I, s_G, O)$. $X$ is a finite set of *variables*, where each $x \in X$ is associated with a finite domain $D_x$. A *partial state* over $X$ is a function $s$ on a subset $X_s$ of $X$, so that $s(x) \in D_x$ for all $x \in X_s$; $s$ is a *state* if $X_s = X$. The *initial state* $s_I$ is a state. The *goal* $s_G$ is a partial state. $O$ is a finite set of *operators*. Each $o \in O$ is a pair $o = (\text{pre}_o, \text{eff}_o)$ of partial states, called its *precondition* and *effect*. As simple non-restricting sanity conditions, we assume that $|D_x| > 1$ for all $x \in X$, and $\text{pre}_o(x) \neq \text{eff}_o(x)$ for all $o \in O$ and $x \in X_{\text{pre}_o} \cap X_{\text{eff}_o}$.

We identify partial states with sets of variable-value pairs, which we will often refer to as *facts*. The *state space* $S$ of the task is the directed graph whose vertices are all states over $X$, with an arc $(s, s')$ iff there exists $o \in O$ such that $\text{pre}_o \subseteq s$, $\text{eff}_o \subseteq s'$, and $s(x) = s'(x)$ for all $x \in X \setminus X_{\text{eff}_o}$. A *plan* is a path in $S$ leading from $s_I$ to a state $s$ with $s_G \subseteq s$.

We next define the two basic structures in our analysis: domain transition graphs and causal graphs. For the former, we diverge from Helmert's definition (only) in that we introduce additional notations indicating the operator responsible for the transition, as well as the "side effects" of the transition, i.e., any other variable values set when executing the responsible operator. In detail, let $x \in X$. The *domain transition graph $DTG_x$* of $x$ is the labeled directed graph with vertex set $D_x$ and the following arcs. For each $o \in O$ where $x \in X_{\text{pre}_o} \cap X_{\text{eff}_o}$ with $c := \text{pre}_o(x)$ and $c' := \text{eff}_o(x)$, $DTG_x$ contains an arc $(c, c')$ labeled with *responsible operator* $\text{rop}(c, c') := o$, with *conditions* $\text{cond}(c, c') := \text{pre}_o \setminus \{(x, c)\}$, and with *side effects* $\text{seff}(c, c') := \text{eff}_o \setminus \{(x, c')\}$. For each $o \in O$ where $x \in X_{\text{eff}_o} \setminus X_{\text{pre}_o}$ with $c' := \text{eff}_o(x)$, for every $c \in D_x$ with $c \neq c'$, $DTG_x$ contains an arc $(c, c')$ labeled with $\text{rop}(c, c') := o$, $\text{cond}(c, c') := \text{pre}_o$, and $\text{seff}(c, c') := \text{eff}_o \setminus \{(x, c')\}$.

The reader familiar with causal graphs may have wondered why we introduced a notion of side effects, seeing as causal graphs can be acyclic only if all operators are unary (affect only a single variable). The reason is that we do handle cases where operators are non-unary. The variant of causal graphs we use can still be acyclic in such cases, and indeed this happens in some of our benchmark domains, specifically in Simple-TSP, Movie, Miconic-STRIPS, and Satellite. We define the *support graph $SG$* to be the directed graph with vertex set $X$, and with an arc $(x, y)$ iff $DTG_y$ has a relevant transition $(c, c')$ so that $x \in X_{\text{cond}(c,c')}$. Here, a transition $(c, c')$ on variable $x$ is called *relevant* iff $(x, c') \in s_G \cup \bigcup_{o \in O} \text{pre}_o$.

Our definition modifies the most commonly used one in that it uses relevant transitions only, and that it does not introduce arcs between variables co-occurring in the same operator effect (unless these variables occur also in the precondition). Transitions with side effects are handled separately in our analysis. Note that irrelevant transitions occur naturally, in domains with non-unary operators. For example, unstacking a block induces the irrelevant transition making the arm non-empty, and departing a passenger in Miconic-STRIPS makes the passenger "not-boarded".[4]

Consider now the definition of $h^+$. In the more common Boolean-variable setting of PDDL, this is defined as the length of a shortest plan solving the problem when ignoring all delete lists, i.e., the negative operator effects (Bylander, 1994; McDermott, 1999; Bonet & Geffner, 2001). This raises the question what $h^+$ actually is, in finite-domain variable planning, where there are no "delete lists". That question is easily answered. "Ignoring deletes" essentially means to act as if "what was true once will remain true forever". In the finite-domain variable setting, this simply means to not over-write any values that the variables had previously. To our knowledge, this generalization was first described by Helmert (2006). Consider the directed graph $S^+$ whose vertices are all sets $s^+$ of variable-value pairs over $X$, with an arc $(s_1^+, s_2^+)$ iff there exists $o \in O$ such that $\text{pre}_o \subseteq s_1^+$ and $s_2^+ = s_1^+ \cup \text{eff}_o$. If $s$ is a state, then a *relaxed plan* for $s$ is a path in $S^+$ leading from $s$ to $s^+$ with $s_G \subseteq s^+$. By $h^+(s)$ we denote the length of a shortest relaxed plan for $s$, or $h^+(s) = \infty$ if no such plan exists. It is easy to see that this definition corresponds to the common Boolean one: if we translate the finite-domain variables into Boolean ones by creating one Boolean variable "is-$(x, c)$-true?" for every fact $(x, c)$, then standard $h^+$ in the Boolean task is identical to $h^+$ in the finite-domain variable task.

---

4. We remark that relevant transitions correspond to what has been called "requestable values" in some works, (e.g., Jonsson & Bäckström, 1998; Haslum, 2007). In Fast Downward's implementation, the causal graph includes only precondition-effect arcs, similarly as the support graph defined here.

Bylander (1994) proved that it is intractable to compute $h^+$. Many state-of-the-art planners approximate $h^+$, in a variety of ways (e.g., McDermott, 1999; Bonet & Geffner, 2001; Hoffmann & Nebel, 2001a; Gerevini et al., 2003; Helmert, 2006; Richter, Helmert, & Westphal, 2008; Richter & Westphal, 2010). A popular approximation in satisficing planning – that gives no guarantees on the quality of the relaxed plan returned – is the so-called *relaxed plan heuristic* first proposed in the FF system (Hoffmann & Nebel, 2001a), which approximates $h^+$ in terms of the length of some not necessarily shortest relaxed plan. Such relaxed plans can be computed in low-order polynomial time using techniques inspired by Graphplan (Blum & Furst, 1997).

We next introduce the relevant notations pertaining to search space topology under $h^+$. Let $s \in S$ be a state where $0 < h^+(s) < \infty$. Then an *exit* is a state $s'$ reachable from $s$ in $S$, so that $h^+(s') = h^+(s)$ and there exists a neighbor $s''$ of $s'$ so that $h^+(s'') < h^+(s')$ (and thus $h^+(s'') < h^+(s)$). The *exit distance* $ed(s)$ of $s$ is the length of a shortest path in $S$ from $s$ to an exit, or $ed(s) = \infty$ if no exit exists. A path in $S$ is called *monotone* iff there exist no two consecutive states $s_1$ and $s_2$ on it so that $h^+(s_1) < h^+(s_2)$. We say that $s$ is a *local minimum* if there exists no monotone path to an exit.

The topology definitions, adapted from the author's previous work (Hoffmann, 2005), are specific to $h^+$ only for the sake of simplicity (we will herein not consider any heuristics other than $h^+$).[5] States with infinite heuristic value are ignored because they are correctly identified, by the heuristic, to be dead ends (relaxed-plan based approximations like that of FF do identify all these cases). If the heuristic value is 0 then we have already reached the goal, so this case can also be safely ignored. Note that we do not force exit paths to be monotone, i.e., we will also talk about exit distances in situations where $s$ may be a local minimum. This is necessary to capture the structure of domains like Satellite and Zenotravel, where local minima exist but their exit distance is bounded. Also, some of our analysis methods guarantee an upper bound on the length of an exit path only, not that the heuristic values on that path will decrease monotonically.

---

5. We remark that the original definitions are significantly more involved, e.g., defining "local minima" not based on individual states but based on strongly connected sub-graphs of the state space. None of these complications is relevant to the results herein.

Finally, let us say a few words on domain analysis. Generally speaking, domain analysis aims at automatically obtaining non-trivial information about a domain or planning task. Such analysis has a long tradition in planning (e.g., Nebel, Dimopoulos, & Koehler, 1997; Fox & Long, 1998; Gerevini & Schubert, 1998; Edelkamp & Helmert, 1999; Rintanen, 2000). Most often, the information sought pertains to reachability or relevance properties, i.e., which entities or combinations thereof are reachable from the initial state/relevant to the goal. A notable exception is the work of Long and Fox (2000) which automatically recognizes certain "generic types" of domains, like transportation. However, there exists no prior work at all trying to automatically infer topological properties of a heuristic function. The single exception are the aforementioned disappointing results reported (as an aside) in the author's previous work (Hoffmann, 2005). This method builds a structure called "fact generation tree", enumerating all ways in which facts may support each other in a non-redundant relaxed plan. If there is no "conflict" then $h^+$ is the *exact* solution distance. Clearly, this is a far too strong property to be applicable in any reasonably complex domain. Of the considered benchmarks, the property applies only in Simple-TSP. A slightly more general property, also identified in this work, applies in Movie as well as trivial Logistics tasks with 2 locations, 1 truck, and 1 package.

It is worth noting that analyzing the topology of $h^+$ is computationally hard:

**Theorem 1.** *It is* **PSPACE**-*complete to decide whether or not the state space of a given planning task contains a local minimum, and given an integer $K$ it is* **PSPACE**-*complete to decide whether or not for all states $s$ we have $ed(s) \leq K$. Further, it is* **PSPACE**-*complete to decide whether or not a given state $s$ is a local minimum, and given an integer $K$ it is* **PSPACE**-*complete to decide whether or not $ed(s) \leq K$.*

These results are hardly surprising, but have not been stated anywhere yet. The membership results in Theorem 1 are easy to prove based on guess-and-check arguments similar as given by Bylander (1994), exploiting the fact that **NPSPACE=PSPACE**. The hardness results still hold when restricting the input to solvable tasks/states. Their proofs work by reducing plan existence, respectively bounded plan existence (with a bound in non-unary representation). Given a task whose plan existence we wish to decide, we flatten $h^+$ by a new operator that can always achieve the goal but that has a fatal side effect. Then we give the planner the choice between solving this task, or solving a new alternative task. That latter task is designed so that a local minimum exists/that the exit distance exceeds the bound iff the planner must choose the alternative task, i.e., iff the original task is unsolvable/iff it cannot be solved within a given number of steps. The full proof is in Appendix A.1.

In practice, computational hardness here is particularly challenging because, in most applications of domain analysis, we are not willing to run a worst-case exponential search. After all, the analysis will not actually solve the problem. Consequently, in the present research, we restrict ourselves to analysis methods with low-order polynomial runtime.

The reader will have noticed the state-specific analysis problems in Theorem 1. We distinguish between *global* analysis per-task, and *local* analysis per-state. More precisely, we herein devise three kinds of analyses:

(I) **Guaranteed global analysis.** Taking as input the planning task description, this analysis returns "yes, $d$" *only if* the state space does not contain any local minima and the exit distance from any state is bounded by $d$.

(II) **Guaranteed local analysis.** Taking as input the planning task description and a state $s$, this analysis returns "yes, $d$" *only if $s$ is not a local minimum,* and the exit distance from $s$ is bounded by $d$.

(III) **Approximate local analysis.** Taking as input the planning task description and a state $s$, this analysis returns "yes, $d$" *to indicate* that $s$ is not a local minimum, and that the exit distance from $s$ is bounded by $d$. Both may be wrong, i.e., the analysis is not guaranteed to be sound. Compared to analysis (II), this trades soundness for the ability to successfully analyze more states.

Domain analysis traditionally considers only the global variant (I), or even more generalizing variants looking at only the PDDL domain file. While global once-and-for-all analysis is also the "holy grail" in our work, local analysis has strong advantages. If a planning task *does* contain local minima – which one would expect to typically be the case in interesting domains – then analysis (I) is useless. It will simply answer "no". By contrast, local analysis (II,III) may still detect some individual states, that we sample randomly in our experiments, to not be local minima. The percentage of such states, which we refer to as the *success rate*, can deliver useful information no matter what the structure of the planning task is. Note also that, while the contrast between a **PSPACE**-hard problem and low-order polynomial analysis runtime necessarily implies that all analyses are incomplete, the local analyses have a chance to ameliorate this by averaging their outcome over a set of sample states.

## 3. An Illustrative Example

The basic connection we identify between causal graphs and $h^+$ topology – more precisely, between support graphs, domain transition graphs, and $h^+$ topology – is quite simple. It is instructive to understand this first, before delving into the full results. Figure 2 shows fragments of the domain transition graphs (DTGs) of three variables $x_0$, $x_1$, and $x_2$. All DTG transitions here are assumed to be invertible, and to have no side effects.

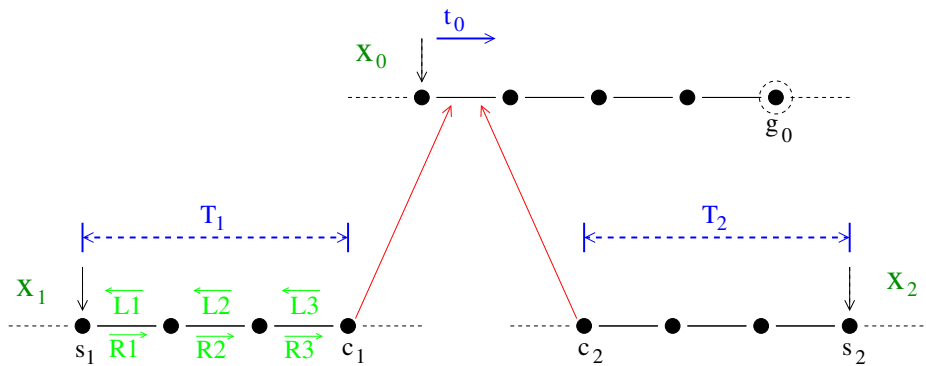

Figure 2: An example illustrating our basic result.

The imaginative reader is invited to think of $x_0$ as a car whose battery is currently empty and that therefore requires the help of two people, $x_1$ and $x_2$, in order to push-start it. The people may, to solve different parts of the task, be required for other purposes too, but here we consider only the sub-problem of achieving the goal $x_0 = g_0$. We wish to take the $x_0$ transition $t_0$, which has the two conditions $c_1$ and $c_2$. These conditions are currently not fulfilled. In the state $s$ at hand, $x_1$ is in $s_1$ and $x_2$ is in $s_2$. We must move to a different state, $s_0$, in which $x_1 = c_1$ and $x_2 = c_2$. What will happen to $h^+$ along the way?

Say that an optimal relaxed plan $P^+(s)$ for $s$ moves $x_1$ to $c_1$ along the path marked $T_1$, and moves $x_2$ to $c_2$ along the path marked $T_2$ – clearly, some such paths will have to be taken by any $P^+(s)$. Key observation (1) is similar to a phenomenon known from transportation benchmarks. When moving $x_1$ and $x_2$, whichever state $s'$ we are in, as long as $s'$ remains within the boundaries of the values traversed by $T_1$ and $T_2$, we can construct a relaxed plan $P^+(s')$ for $s'$ so that $|P^+(s')| \leq |P^+(s)|$. Namely, to obtain $P^+(s')$, we simply replace the respective move sequence $\overrightarrow{o}_i$ in $P^+(s)$, for $i = 1, 2$, with its inverse $\overleftarrow{o}_i$. For example, say we got to $s'$ by $\overrightarrow{o}_1 = \langle R1, R2, R3 \rangle$ moving $x_1$ to $c_1$, as indicated in Figure 2. Then wlog $P^+(s)$ has the form $\langle R1, R2, R3 \rangle \circ P$. We define $P^+(s') := \langle L3, L2, L1 \rangle \circ P$. The postfix $P$ of both relaxed plans is the same; at the end of the prefix, the set of values achieved for $x_1$, namely $s_1$, $c_1$, and the two values in between, is also the same. Thus $P^+(s')$ is a relaxed plan for $s'$.[6] This is true in general, i.e., $\overleftarrow{o}_1$ is necessarily applicable in $s'$, and will achieve, within relaxed execution of $P^+(s')$, the same set of facts as achieved by $\overrightarrow{o}_1$ in $P^+(s)$. Thus $h^+(s') \leq h^+(s)$ for any state $s'$, including the state $s_0$ we're after.

Key observation (2) pertains to the "leaf" variable, $x_0$. Say that $x_0$ moves only for its own sake, i.e., the car position is not important for any other goal. Then executing $t_0$ in $s_0$ does not delete anything needed anywhere else. Thus we can remove rop($t_0$) from the relaxed plan $P^+(s_0)$ for $s_0$ – constructed as per observation (1) – to obtain a relaxed plan for the state $s_1$ that results from executing $t_0$ in $s_0$. Hence $h^+(s_1) < h^+(s)$. With observation (1), the heuristic values along the path to $s_1$ are all $\leq h^+(s)$. We know that at least one state $s''$ on the path has a heuristic value strictly smaller than $h^+(s)$: this happens at the latest in $s'' = s_1$, and may happen earlier on in case the relaxed plan $P^+(s'')$ as constructed here is not optimal (cf. Footnote 6). Let $s''$ be the earliest state with $h^+(s'') < h^+(s)$ on the path, and let $s'$ be the state preceding $s''$. Then $s'$ is an exit for $s$, and the path to that exit is monotone. Thus $s$ is not a local minimum. As for the exit distance, in the worst case we have $s'' = s_1$ and $s' = s_0$, so $ed(s)$ is bounded by the length of the path up to $s_0$.

It is not difficult to imagine that the above works also if preconditions need to be established recursively, as long as no cyclic dependencies exist. A third person may be needed to first persuade $x_1$ and $x_2$, the third person may need to take a bus, and so on. The length of the path to $s_0$ may grow exponentially – if $x_1$ depends on $x_3$ then each move of $x_1$ may require several moves of $x_3$, and so forth – but we will still be able to construct $P^+(s')$ by inverting the moves of all variables individually. Further, the inverting transitions may have conditions, too, provided these conditions are the same as required by the original moves. For example, in the above, the inverting operator $L1$ may have an arbitrary condition $p$ if that condition is also required for $R1$. This is because any conditions that are required for the original moves (like $p$ for $R1$) are established in $P^+(s)$, and thus will be established in $P^+(s')$ in time for the inverse moves (like $L1$).

---

6. Note that $P^+(s')$ may not be an *optimal* relaxed plan for $s'$. If $P^+(s)$ does not move $x_1$ for anything other than attaining $c_1$, then the postfix $P$ alone is a relaxed plan for $s'$: there is no need to insert the inverted prefix $\langle L3, L2, L1 \rangle$. In cases like this, we obtain an exit state already on the path to $s_0$; we get back to this below.

Now, say that the support graph is acyclic, and that all transitions are invertible and have no side effects. Given any state $s$, unless $s$ is already a goal state, some variable $x_0$ moving only for its own sake necessarily exists. But then, within any optimal relaxed plan for $s$, a situation as above exists, and therefore we have a monotone exit path, *Q.E.D. for no local minima under $h^+$*.

The execution path construction just discussed is not so different from known results exploiting causal graph acyclicity and notions of connectedness or invertibility of domain transition graphs (e.g., Jonsson & Bäckström, 1995; Williams & Nayak, 1997). What is new here is the connection to $h^+$.

We remark that the hand-made analysis of $h^+$ (Hoffmann, 2005) uses a notion of operators "respected by the relaxation". An operator $o$ is respected by the relaxation iff, whenever $o$ starts an optimal plan for $s$, then $o$ also starts an optimal relaxed plan for $s$. A core property of many of the hand-made proofs is that all operators are respected by the relaxation. This motivated the speculation that recognizing this property automatically could be key to domain analysis recognizing the absence of local minima under $h^+$. We do not explore this option herein, however we note that even the basic result we just outlined contains cases not covered by this property. Even with acyclic support graph and invertible transitions without side effects, there are examples where an operator is not respected by the relaxation. We give such a construction in Example 1, Appendix A.4.

## 4. Synopsis of Technical Results

Our technical results in what follows are structured in a way similar to the proof argument outlined in the previous section. The results are structured into two parts, (A) and (B). In (A), Section 5, we identify circumstances under which we can deduce from an optimal relaxed plan that a monotone exit path exists. In (B), Section 6, we devise support-graph based sufficient criteria implying that analysis (A) will always succeed. Technique (B) underlies TorchLight's conservative analysis methods, i.e., guaranteed global analysis (I) and guaranteed local analysis (II) as described at the end of Section 2. By feeding technique (A) with the usual relaxed plans as computed, e.g., by FF's heuristic function, we obtain TorchLight's approximate local analysis (III). That analysis does not give a guarantee, because (and only because) FF's relaxed plans are not guaranteed to be optimal.

For ease of reading, we now give a brief synopsis of the results obtained in (A) and (B), and how they provide the analysis methods (I)–(III). The synopsis contains sufficient information to understand the rest of the paper, so the reader may choose to skip Sections 5 and 6, moving directly to the evaluation.

Each analysis method is based on a particular kind of sub-graph of the support graph. Table 1 overviews these. Their role in parts (A) and (B) is as follows:

(A) Given an optimal relaxed plan $P^+(s)$ for a state $s$, an *optimal rplan dependency graph* $oDG^+$ is a sub-graph of $SG$ with a single leaf variable $x_0$ with transition $t_0$ as in our example ($\text{rop}(t_0)$ will be frequently referred to as $o_0$). An arc $(x, x')$ is in $oDG^+$ if $P^+(s)$ relies on $x'$ to achieve the conditions of $t_0$, and $P^+(s)$ relies on $x$ for moving $x'$. We say that $oDG^+$ is *successful* if it is acyclic, all involved transitions will be usable in our exit path construction (e.g., they have no harmful side effects), and the deletes of $t_0$ are either not relevant to $P^+(s)$ at all, or are being recovered inside $P^+(s)$. The main result, Theorem 2, states that $s$ is no local minimum if there exists a successful $oDG^+$ for $s$. It also derives an exit distance bound from $oDG^+$. Approximating Theorem 2 by applying it to a relaxed plan as computed by FF's heuristic yields analysis (III).

| Name | Symbol | Analysis | Leaves | Arcs |
|------|--------|----------|--------|------|
| Support graph | $SG$ | – | All | All |
| Optimal rplan dependency graph | $oDG^+$ | Approximate local analysis (III) Theorem 2 | Single leaf $x_0$ s.t. applying $t_0$ does not affect the remainder of $P^+(s)$ | $(x, x')$ where $x$ is used in $P^+(s)$ to support $x'$ for obtaining $\text{cond}(t_0)$ |
| Local dependency graph | $lDG$ | Guaranteed local analysis (II) Theorem 3 | Single leaf $x_0 \in X_{s_G}$, $s(x_0) \neq s_G(x_0)$ and $x_0$ has no transitive $SG$ successor with same property | $(x, x_0)$ where $s(x) \neq \text{cond}(t_0)(x)$; and $(x, x')$ where $x'$ is in $lDG$ and $(x, x')$ is in $SG$ |
| Global dependency graph | $gDG$ | Guaranteed global analysis (I) Theorem 4 | Single leaf $x_0 \in X_{s_G}$ | $(x, x_0)$ where $x \neq x_0$; and $(x, x')$ where $x'$ is in $gDG$ and $(x, x')$ is in $SG$ |

Table 1: Overview of the different support graph sub-graphs underlying our results.

(B) Given a state $s$, a *local dependency graph* $lDG$ is a sub-graph of $SG$ with a single leaf variable $x_0$, whose goal value is yet unachieved, and all of whose transitive successors in $SG$ have already attained their goal values. In this setting, $x_0$ "moves for its own sake" as in the example. The graph $lDG$ simply includes all $SG$ predecessors of $x_0$, the single exception pertaining to arcs $(x, x_0)$ into $x_0$ itself, which are not inserted if the corresponding condition of $t_0$ is already satisfied in $s$. We say that $lDG$ is successful if it is acyclic, all involved transitions will be usable in our exit path construction, and $t_0$ does not have any relevant deletes. This implies that there exists a successful $oDG^+$ contained in $lDG$, and thus we have Theorem 3, stating that $s$ is no local minimum and giving a corresponding exit distance bound. This result underlies analysis (II).

A *global dependency graph* $gDG$ is a sub-graph of $SG$ that identifies any goal variable $x_0$, and includes all $SG$ predecessors of $x_0$. Being successful is defined in the same way as for $lDG$s. If all $gDG$s are successful, then Theorem 3 will apply to every state because each $lDG$ is contained in a successful $gDG$. Thus we have Theorem 4, stating that the state space does not contain any local minima. The exit distance bound is obtained by maximizing over all $gDG$s. This result underlies analysis (I).

For understanding the practical performance of TorchLight, it is important to note that (A) is not only a minimal result that would suffice to prove (B). The cases identified by Theorem 2 are much richer than what we can actually infer from support graphs. For this reason, analysis (III), while not sound due to the use of potentially non-optimal relaxed plans, is able to analyze a much larger class of states than analysis (II). In a little detail, the difference between the two methods pertains to (1) whether "$P^+(s)$ relies on values of $x$ for moving $x'$", and (2) whether "the deletes of $t_0$ are being recovered inside $P^+(s)$". Neither (1) nor (2) are visible in the support graph, because both rely on details of the form of the relaxed plan $P^+(s)$. For example, consider the Gripper domain. Notion (1) is important because the support graph contains the arcs ("carry-ball-$b$", "free-gripper") – due to dropping ball $b$ – and ("free-gripper", "carry-ball-$b$") – due to picking up ball $b$. Thus, looking only at $SG$, it seems that "carry-ball-$b$" may support itself (free the gripper by dropping the ball we want to pick up). Of course, that doesn't happen in an optimal relaxed plan. Notion (2) is important because some operators (picking up a ball) do have harmful side effects (making the gripper hand non-empty), but these side effects are always recovered inside the relaxed plan (when dropping the ball again later on). It remains future work to extend analyses (I,II) so that they can detect these kinds of phenomenona.

## 5. Analyzing Optimal Relaxed Plans

We consider a state $s$ and an optimal relaxed plan $P^+(s)$ for $s$. To describe the circumstances under which a monotone exit path is guaranteed to exist, we will need a number of notations pertaining to properties of transitions etc. We will introduce these notations along the way, rather than up front, in the hope that this makes them easier to digest.

Given $o_0 \in P^+(s)$, by $P^+_{<0}(s)$ and $P^+_{>0}(s)$ we denote the parts of $P^+(s)$ in front of $o_0$ and behind $o_0$, respectively. By $P^+(s, x)$ we denote the sub-sequence of $P^+(s)$ affecting $x$. We capture the dependencies between the variables used in $P^+(s)$ for achieving the precondition of $o_0$, as follows:

**Definition 1.** *Let $(X, s_I, s_G, O)$ be a planning task, let $s \in S$ with $0 < h^+(s) < \infty$, let $P^+(s)$ be an optimal relaxed plan for $s$, let $x_0 \in X$, and let $o_0 \in P^+(s)$ be an operator taking a relevant transition of the form $t_0 = (s(x_0), c)$.*

*An* optimal rplan dependency graph for $P^+(s)$, $x_0$ and $o_0$, *or* optimal rplan dependency graph for $P^+(s)$ in brief, *is a graph $oDG^+ = (V, A)$ with unique leaf vertex $x_0$, and where $x \in V$ and $(x, x') \in A$ if either: $x' = x_0$, and $\text{pre}_{o_0}(x) \neq s(x)$; or $x \neq x' \in V \setminus \{x_0\}$ and there exists $o \in P^+_{<0}(s)$ taking a relevant transition on $x'$ so that $x \in X_{\text{pre}_o}$ and $\text{pre}_o(x) \neq s(x)$.*

*For $x \in V \setminus \{x_0\}$, by $oDTG^+_x$ we denote the sub-graph of $DTG_x$ that includes only the values true at some point in $P^+_{<0}(s, x)$, the relevant transitions $t$ using an operator in $P^+_{<0}(s, x)$, and at least one relevant inverse of such $t$ where a relevant inverse exists. We refer to the $P^+_{<0}(s, x)$ transitions as* original, *and to the inverse transitions as* induced.

The transition $t_0$ with responsible operator $o_0$ will be our candidate for reaching the exit state, like $t_0$ in Figure 2. $oDG^+$ collects all variables $x$ connected to a variable $x'$ insofar as $P^+_{<0}(s)$ uses an operator preconditioned on $x$ in order to move $x'$. These are the variables we will need to move, like $x_1$ and $x_2$ in Figure 2, to obtain a state $s_0$ where $t_0$ can be taken. For any such variable $x$, $oDTG^+_x$ captures the domain transition graph fragment that $P^+_{<0}(s)$ traverses and within which we will stay, like $T_1$ and $T_2$ in Figure 2.

Note that there is no need to consider the operators $P^+_{>0}(s)$ behind $o_0$, simply because these operators are not used in order to establish $o_0$'s precondition. This is of paramount importance in practice. An example is the Gripper situation mentioned above. if $o_0$ picks up a ball $b$ in Gripper, then $P^+(s)$ will also contain – behind $o_0$, i.e., in $P^+_{>0}(s)$ – an operator $o'$ dropping $b$. If we considered $o'$ in Definition 1, then $oDG^+$ would contain the mentioned cycle assuming that $o'$ is used for making the gripper hand free for picking up $b$. In TorchLight's approximate local analysis, whenever we consider an operator $o_0$, before we build $oDG^+$ we re-order $P^+(s)$ by moving operators behind $o_0$ if possible. This minimizes $P^+_{<0}(s)$, and $oDG^+$ thus indeed contains only the necessary variables and arcs.

Under which circumstances will $t_0$ actually "do the job"? The sufficient criterion we identify is rather complex. To provide an overview of the criterion, we next state its definition. The items in this definition will be explained below.

**Definition 2.** *Let* $(X, s_I, s_G, O)$, $s$, $P^+(s)$, $x_0$, $o_0$, $t_0$, *and* $oDG^+ = (V, A)$ *be as in Definition 1. We say that* $oDG^+$ *is* successful *if all of the following holds:*

*(1)* $oDG^+$ *is acyclic.*

*(2) We have that either:*

    *(a) the $oDG^+$-relevant deletes of $t_0$ are $P^+_{>0}(s)$-recoverable; or*

    *(b) $s(x_0)$ is not $oDG^+$-relevant, and $t_0$ has replaceable side effect deletes; or*

    *(c) $s(x_0)$ is not $oDG^+$-relevant, and $t_0$ has recoverable side effect deletes.*

*(3) For $x \in V \setminus \{x_0\}$, all $oDTG^+_x$ transitions either have self-irrelevant deletes, or are invertible/induced and have irrelevant side effect deletes and no side effects on $V \setminus \{x_0\}$.*

As already outlined, our exit path construction works by staying within the ranges of $oDTG^+_x$, for $x \in V \setminus \{x_0\}$, until we have reached a state $s_0$ where the transition $t_0$ can be taken. To make this a little more precise, consider a topological order $x_k, \ldots, x_1$ of $V \setminus \{x_0\}$ with respect to $oDG^+$ – such an order exists due to Definition 2 condition (1). (If there are cycles, then moving a variable may involve moving itself in the first place, which is not covered by our exit path construction.) Now consider, for $0 \leq d \leq k$, the *$d$-abstracted* task. This is like the original task except that, for every transition $t$ of one of the graphs $oDTG^+_{x_i}$ with $i \leq d$, we remove each condition $(x_j, c) \in \text{cond}(t)$ where $j > d$. The exit path construction can then be understood as an induction over $d$, proving the existence of an execution path $\overrightarrow{o}$ at whose end $t_0$ can be taken. We construct $\overrightarrow{o}$ exclusively by operators responsible for transitions in $oDTG^+_x$, for $x \in V \setminus \{x_0\}$. For the base case, in the 0-abstracted task, $t_0$ is directly applicable. For the inductive case, if we have constructed a suitable path $\overrightarrow{o}_d$ for the $d$-abstracted task, then a suitable path $\overrightarrow{o}_{d+1}$ for the $d + 1$-abstracted task can be constructed as follows. Assume that $o$ is an operator in $\overrightarrow{o}_d$, and that $o$ has a precondition $(x_{d+1}, c)$ that is not true in the current state. Then, in $\overrightarrow{o}_{d+1}$, in front of $o$ we simply insert a path through $oDTG^+_{x_{d+1}}$ that ends in $c$. Note here that, by construction, $(x_{d+1}, c)$ is a condition of a transition $t$ in $oDTG^+_{x_i}$, for some $i < d + 1$. If $t$ is taken in $P^+_{<0}(s, x)$, then $(x_{d+1}, c)$ must be achieved by $P^+_{<0}(s)$ and thus $c$ is a node in $oDTG^+_{x_{d+1}}$. If $t$ is an induced transition – inverting a transition taken in $P^+_{<0}(s, x)$ – then the same is the case unless the inverse may introduce new outside conditions. We thus need to exclude this case, leading to the following definition of "invertibility":

- Let $t = (c, c')$ be a transition on variable $x$. We say that $t$ is *invertible* iff there exists a transition $(c', c)$ in $DTG_x$ so that $\text{cond}(c', c) \subseteq \text{cond}(c, c')$.

A transition is invertible if we can "go back" without introducing any new conditions (e.g., driving trucks in Logistics). There are subtle differences to previous definitions of "invertible operators", like the author's (Hoffmann, 2005). We do not allow new conditions even if they are actually established by the operator $\text{rop}(t)$ responsible for $t$. This is because, on $\overrightarrow{o}$, we do not necessarily execute $t$ before executing its inverse – we may have got to the endpoint of $t$ via a different path in $oDTG^+_x$. On the other hand, our definition is also more generous than common ones because, per se, it does not care about any side effects the inverse transition may have (side effects are constrained separately as stated in Definition 2).

Consider Definition 2 condition (3). Apart from the constraints on conditions of induced transitions, for the $oDTG_x^+$ transitions taken by $\overrightarrow{o}$, we must also make sure that there are no harmful side effects. Obviously, this is the case if, as in the example from Section 3, the transitions have no side effects at all. However, we can easily generalize this condition. Let $t = (c, c')$ be a transition on variable $x$.

- The *context* of $t$ is the set $\mathrm{ctx}(t)$ of all facts that may be deleted by side effects of $t$. For each $(y, d) \in \mathrm{seff}(t)$, $(y, \mathrm{cond}(t)(y)) \in \mathrm{ctx}(t)$ if a condition on $y$ is defined; else all $D_y$ values $\neq d$ are inserted.

- We say that $t$ has *irrelevant side effect deletes* iff $\mathrm{ctx}(t) \cap (s_G \cup \bigcup_{o \in O} \mathrm{pre}_o) = \emptyset$.

- We say that $t$ has *self-irrelevant side effect deletes* iff $\mathrm{ctx}(t) \cap (s_G \cup \bigcup_{\mathrm{rop}(t) \neq o \in O} \mathrm{pre}_o) = \emptyset$.

- We say that $t$ has *self-irrelevant deletes* iff it has self-irrelevant side effect deletes and $(x, c) \notin s_G \cup \bigcup_{\mathrm{rop}(t) \neq o \in O} \mathrm{pre}_o$.

Irrelevant side effect deletes capture the case where no side effect delete occurs in the goal or in the precondition of any operator. Self-irrelevant side effect deletes are slightly more generous in that they allow to delete conditions needed only for the responsible operator $\mathrm{rop}(t)$ itself. Self-irrelevant deletes, finally, extend the latter notion also to $t$'s "own delete". In a nutshell, we need to postulate irrelevant side effect deletes for transitions that may be executed again, on our path. Examples of irrelevant side effect deletes are transitions with no side effects at all, or a move in Simple-TSP, whose side effect, when $x_0$="at", deletes the target location's being "not-visited". An example of an operator with self-irrelevant side effect deletes, but no irrelevant side effect deletes, is departing a passenger in Miconic-STRIPS, whose side effect, when $x_0$="served", deletes "boarded(passenger)" which is used only for the purpose of this departure. In fact, this transition has self-irrelevant deletes because its own effect deletes "not-served(passenger)" which obviously is irrelevant. Another example of self-irrelevant deletes is inflating a spare wheel in Tyreworld – the wheel is no longer "not-inflated".

Clearly, if all $oDTG_x^+$ transitions $t$ we may be using on $\overrightarrow{o}$ have irrelevant side effect deletes, then, as far as not invalidating any facts needed elsewhere is concerned, this is just as good as having no side effects at all. To understand why we need to require that $t$'s side effect is not used to move another variable $x' \in V \setminus \{x_0\}$, recall that, for the states $s'$ visited by $\overrightarrow{o}$, we construct relaxed plans $P^+(s')$ with $|P^+(s')| \leq |P^+(s)|$ by inverting such transitions $t$. Now, say that $t$'s side effect is used to move another variable $x' \in V \setminus \{x_0\}$. Then we may have to invert both transitions separately (with different operators), and thus we would have $|P^+(s')| > |P^+(s)|$.

Regarding the own delete of $t$, this may be important for two reasons. First, the deleted fact may be needed in the relaxed plan for $s'$. Second, $x$ may have to traverse $oDTG_x^+$ several times, and thus we may need to traverse the deleted value again later on. Both are covered if $t$ is invertible, like we earlier on assumed for all transitions. Now, what if $t$ is not invertible? This does not constitute a problem in case that $t$ has self-irrelevant deletes: in that case, all deletes of $t$ are irrelevant except maybe for the responsible operator itself. Therefore, to obtain $P^+(s')$, we can simply remove rop($t$) from the relaxed plan constructed for the predecessor state $s''$. Thus $|P^+(s')| < |P^+(s)|$ so we have reached an exit and there is no need to continue the construction of $\overrightarrow{o}$. For example, consider $t$ that inflates a spare wheel $W$ in Tyreworld. This deletes only "not-inflated(W)", and thus has self-irrelevant deletes ("not-inflated(W)" is irrelevant for the goal and any other operator). Say that we are in a state $s''$ with relaxed plan $P^+(s'')$ constructed as described. We have $|P^+(s'')| \leq |P^+(s)|$. We also have rop($t$) ="inflate-W"$\in P^+(s'')$, because "inflate-W"$\in P^+(s)$, and because "inflate-W" was not executed as yet on our path, and was hence not removed from the relaxed plan. Applying "inflate-W" to $s''$, we get to a state $s'$ identical to $s''$ except that $W$ is now inflated. Clearly, the relaxed plan for $s'$ no longer needs to apply "inflate-W", and the rest of the relaxed plan $P^+(s'')$ still works unchanged. Thus $P^+(s')$ can be obtained by removing "inflate-W" from $P^+(s'')$, yielding $|P^+(s')| < |P^+(s)|$ as desired.

Consider now our endpoint transition $t_0$ and its responsible operator $o_0$. We previously demanded that $x_0$ "moves for its own sake", i.e., that $x_0$ has a goal value and is not important for achieving any other goal. This is unnecessarily restrictive. For example, in Miconic-STRIPS, if we board a passenger then $h^+$ decreases because we can remove the boarding operator from the relaxed plan. However, boarding is only a means for serving the passenger later on, so this variable $x_0$ has no own goal. In Driverlog, a driver may have its own goal *and* be needed to drive vehicles, and still $t_0$ moving the driver results in decreased $h^+$ if the location moved away from is not actually needed anymore. The latter example immediately leads to a definition capturing also the first one: all we want is that "any deletes of $t_0$ are not needed in the rest of the relaxed plan". We can then remove $o_0$ from the relaxed plan for $s_0$, and have reached an exit as desired.

To make this precise, recall the situation we are addressing. We have reached a state $s_0$ in which $t_0 = (s(x_0), c)$ can be applied, yielding a state $s_1$. We have a relaxed plan $P^+(s_0)$ for $s_0$ so that $|P^+(s_0)| \leq |P^+(s)|$, where $P^+(s_0)$ is constructed from $P^+(s)$ by replacing some operators of $P^+_{<0}(s)$ with operators responsible for induced $oDTG^+_x$ transitions for $x \in V \setminus \{x_0\}$. We construct $P^+_1$ by removing $o_0$ from $P^+(s_0)$, and we need $P^+_1$ to be a relaxed plan for $s_1$. What are the facts possibly needed in $P^+_1$? A safe approximation is the union of $s_G$, the precondition of any $o_0 \neq o \in P^+(s)$, and any $oDTG^+_x$ values needed by induced $oDTG^+_x$ transitions.[7] Denote that set with $R^+_1$. The values potentially deleted by $t_0$ are contained in $C_0 := \{(x_0, s(x_0))\} \cup \text{ctx}(t_0)$. Thus if $R^+_1 \cap C_0 = \emptyset$ then we are fine. Simple examples for this have been given above already. In Miconic-STRIPS, the only delete of $o_0$ boarding passenger "P" is "not-boarded(P)", which is not contained in any operator precondition or the goal and thus the intersection of $R^+_1$ with $C_0 = \{$"not-boarded(P)"$\}$ is empty. In Driverlog, $C_0 = \{$"at(D,A)"$\}$ is the delete of $o_0$ moving driver "D" away from location "A". If that location is irrelevant to the rest of the task, then we will have "at(D,A)"$\notin R^+_1$ and thus, again, $R^+_1 \cap C_0 = \emptyset$.

We can sharpen this further. Consider the set of facts $F_0 := s \cup \bigcup_{o \in P^+_{<0}(s)} \text{eff}_o$ that are true after relaxed execution of $P^+_{<0}(s)$. Say that $p \notin F_0$. Then $p$ is not needed for $P_1^+$ to be a relaxed plan for $s_1$. To see this, note first that $p$ is not needed in the part of $P_1^+$ pertaining to $P_{<0}^+(s)$. More precisely, $p$ cannot be an operator precondition in $P_{<0}^+(s)$ because this condition would not be satisfied in (relaxed) execution of $P^+(s)$. Also, $p$ cannot be the start value of an induced $oDTG_x^+$ transition because, by definition, all such values are added by operators in $P_{<0}^+(s)$. Now, what about the part of $P_1^+$ pertaining to $P_{>0}^+(s)$? Assume that $p$ is either a goal, or is an operator precondition in $P_{>0}^+(s)$. Then, since $p \notin F_0$ and $P^+(s)$ is a relaxed plan, either $o_0$ or an operator in $P_{>0}^+(s)$ must establish $p$. As for $o_0$, all its effects are true in $s_1$ anyway. As for $P_{>0}^+(s)$, this remains unchanged in $P_1^+$ and thus this part is covered, too. Altogether, it thus suffices if $R_1^+ \cap C_0 \cap F_0 = \emptyset$. An example where this helps is the Satellite domain. Say that $o_0$ switches on instrument "I". This deletes calibration, i.e., "calibrated(I)"$\in C_0$. The only purpose of switching "I" on can be to take images with it, and thus "calibrated(I)"$\in R_1^+ \cap C_0$. However, the instrument may not actually be calibrated in $s$. If that is so, then we need to switch "I" on before it can be calibrated – because the calibration operator requires to have power in "I" – and thus "calibrated(I)" will be false in the relaxed execution of $P^+(s)$, up to at least $o_0$. In particular, we have "calibrated(I)"$\notin F_0$ and thus $R_1^+ \cap C_0 \cap F_0 = \emptyset$.

---

7. To understand the latter two items, note first that operators preceding $o_0$ in $P^+(s)$, i.e., operators from $P^+_{<0}(s)$, may still be contained in $P^+_1$ and thus it does not suffice to include the preconditions only of operators $o \in P^+_{\geq 0}(s)$. As for $oDTG^+_x$ values needed by induced $oDTG^+_x$ transitions, these may be needed in $P^+_1$ but not in $P^+_{<0}(s)$.

Even the condition $R_1^+ \cap C_0 \cap F_0 = \emptyset$ can still be sharpened. Say that there exists a (possibly empty) sub-sequence $\overrightarrow{o_0}$ of $P_{>0}^+(s)$ so that $\overrightarrow{o_0}$ is guaranteed to be applicable at the start of $P_1^+$, and so that $\overrightarrow{o_0}$ re-achieves all facts in $R_1^+ \cap C_0 \cap F_0$ (both are easy to define and test). Then moving $\overrightarrow{o_0}$ to the start of $P_1^+$ does the job. We say in this case that *the $oDG^+$-relevant deletes of $t_0$ are $P_{>0}^+(s)$-recoverable* – Definition 2 condition (2a). For example, consider $o_0$ that picks up a ball $b$ in the Gripper domain. This operator deletes a fact $p =$ "free-gripper" which may be needed in the remainder of the relaxed plan, and thus $p \in R_1^+ \cap C_0 \cap F_0$. However, $P_{>0}^+(s)$ will necessarily contain a sub-sequence $\overrightarrow{o_0}$ that moves to another room and then puts $b$ down again. We can re-order $P_1^+$ to put $\overrightarrow{o_0}$ right at the start, re-achieving $p$. Similar patterns occur in any transportation domain with capacity constraints, or more generally in domains with renewable resources.

Finally, we have identified two simple alternative sufficient conditions under which $t_0$ is suitable, Definition 2 conditions (2b) and (2c). For the sake of brevity, we only sketch them here. Both require that $s(x_0)$, i.e., the start value of $t_0$, is not contained in $R_1^+$ as defined above. We say in this case that $s(x_0)$ *is not $oDG^+$-relevant*. Note that, then, $R_1^+ \cap C_0 = \emptyset$ unless $t_0$ has side effects. Side effects do not hurt if $t_0$ has *replaceable side effect deletes*, i.e., if any operator whose precondition may be deleted can be replaced with an alternative operator $o'$ that is applicable and has the same effect (this happens, e.g., in Simple-TSP). Another possibility is that where $t_0$ has *recoverable side effect deletes*: there exists an operator $o'$ that is necessarily applicable directly after execution of $t_0$, and that recovers all relevant side effect deletes. This happens quite frequently, for example in Rovers where taking a rock/soil sample fills a "store", but we can free the store again simply by emptying it anywhere. We can replace $o_0$ with $o'$ to obtain a relaxed plan $P_1^+$ for $s_1$ (and thus $h^+(s_1) \le h^+(s)$). Then we can apply $o'$, yielding a state $s_2$ which has $h^+(s_2) < h^+(s)$ because we can obtain a relaxed plan for $s_2$ by removing $o'$ from $P_1^+$.

What will the length of the exit path be? We have one move for $x_0$. Each non-leaf variable $x$ must provide a new value at most once for every move of a variable $x'$ depending on it, i.e., where $(x, x') \in A$. The new value can be reached by a $oDTG_x^+$ traversal. Denote the maximum length of such a traversal, i.e., the diameter of $oDTG_x^+$, by $\operatorname{diam}(oDTG_x^+)$.[8]  Now, we may have $\operatorname{diam}(oDTG_x^+) > \operatorname{diam}(DTG_x)$ because $oDTG_x^+$ removes not only vertices but also arcs. There may be "short-cuts" not traversed by $P^+(s)$. Under certain circumstances it is safe to take these short-cuts, namely if:

(*) all $oDTG_x^+$ transitions are invertible/induced and have irrelevant side effect deletes and no side effects on $V \setminus \{x_0\}$, and all other $DTG_x$ transitions either are irrelevant, or have empty conditions and irrelevant side effect deletes.

When traversing a short-cut under this condition, as soon as we reach the end of the short-cut, we are back in the region of states $s'$ where a relaxed plan $P^+(s')$ can be constructed as before. The rest of our exit path construction remains unaffected. Thus, denote by $V^*$ the subset of $V \setminus \{x_0\}$ for which (*) holds. We define $\operatorname{cost}^{d*}(oDG^+) := \sum_{x \in V} \operatorname{cost}^{d*}(x)$, where $\operatorname{cost}^{d*}(x) :=$

$$
\begin{cases}
1 & x = x_0 \\
\operatorname{diam}(oDTG_x^+) * \sum_{x':(x,x')\in A} \operatorname{cost}^{d*}(x') & x \neq x_0, x \notin V^* \\
\min(\operatorname{diam}(oDTG_x^+), \operatorname{diam}(DTG_x)) * \sum_{x':(x,x')\in A} \operatorname{cost}^{d*}(x') & x \neq x_0, x \in V^*
\end{cases}
$$

Note that $\operatorname{cost}^{d*}(.)$ is exponential in the depth of the graph. This is not an artifact of our length estimation. It is easy to construct examples where exit distance is exponential in that parameter. This is because, as hinted, a variable may have to move several times for each value required by other variables depending on it. See Example 6 in Appendix A.4 for such a construction (following an earlier construction in Domshlak & Dinitz, 2001).

That said, of course $\operatorname{cost}^{d*}(.)$ may over-estimate the length of a shortest exit path. It assumes that, whenever a variable $x'$ with $(x,x') \in A$ makes a move, then $x$ must move through its entire $oDTG^+$ respectively $DTG$. This is very conservative: (1) it may be that the move of $x'$ does not actually have a condition on $x$; (2) even if such a condition exists, $x$ may need less steps in order to reach it. One might be able to ameliorate (1) by making more fine-grained distinctions which part of $\operatorname{cost}^{d*}(x')$ pertains to moves conditioned on $x$. We leave this open for future work. For now, we note that the over-estimation can be exponential even just due to (2), i.e., $\operatorname{cost}^{d*}(oDG^+)$ may be exponentially larger than the length of a shortest exit path even if, for all $(x,x') \in A$, all moves of $x'$ depend on $x$. This can be shown by a simple variant of Example 6; we discuss this in Appendix A.4.

Exit paths using short-cuts in the described way may be non-monotone. Example 5 in Appendix A.4 contains a construction showing this. For an intuitive understanding, imagine a line $l_0, \ldots, l_n$ where our current task, to achieve the precondition of another operator, is to move from $l_0$ to $l_n$. Say that all locations on the line need to be visited, in the relaxed plan, e.g. because we need to load or unload something at all of these locations. Say further that there is a shortcut via $l'$ that needs not be visited. If we move to $l'$ then $h^+$ increases because we have made it 1 step more costly – for the relaxed plan – to reach all the locations $l_0, \ldots, l_n$. For the same reason, $\operatorname{cost}^{d*}(oDG^+)$ is *not* an upper bound on the length of a shortest *monotone* exit path. This is also shown in Example 5, where we construct a situation in which the shortest monotone exit path is longer than $\text{cost}^{\text{d*}}(oDG^+)$.[9] To obtain a bound on monotone exit paths, we can simply set $V^* := \emptyset$ in the definition of $\text{cost}^{\text{d*}}$.

---

8. More precisely, $\operatorname{diam}(.)$ is not the diameter of a graph but the maximum distance from vertex $v$ to vertex $v'$ where there exists a path from $v$ to $v'$.

If we have Definition 2 condition (2a) or (2b), then the exit distance is bounded by $\text{cost}^{\text{d*}}(oDG^+) - 1$ because $\text{cost}^{\text{d*}}(oDG^+)$ counts the last step reducing $h^+$. If we have Definition 2 condition (2c), then after that last step we need 1 additional operator to reduce $h^+$, and so the exit distance is bounded by $\text{cost}^{\text{d*}}(oDG^+)$. Putting the pieces together yields our main result of this section:

**Theorem 2.** *Let $(X, s_I, s_G, O)$, $s$, $P^+(s)$, and $oDG^+$ be as in Definition 1. If $oDG^+$ is successful, then $s$ is not a local minimum, and $ed(s) \leq \text{cost}^{\text{d*}}(oDG^+)$. If we have Definition 2 condition (2a) or (2b), then $ed(s) \leq \text{cost}^{\text{d*}}(oDG^+) - 1$.*

The full proof is in Appendix A.2. As pointed out earlier, for approximate local analysis (III) we simply feed Theorem 2 with the relaxed plans returned by FF's heuristic function (Hoffmann & Nebel, 2001a). It is important to note that, this way, we do not give any guarantees, i.e., Theorem 2 does *not* hold if $P^+(s)$ is not optimal, and even if $P^+(s)$ is non-redundant and parallel-optimal like those computed by FF. At the end of the "exit path" we may obtain a relaxed plan shorter than $P^+(s)$ but not shorter than $h^+(s)$. In a nutshell, the reason is that a parallel-optimal relaxed plan – more generally, a relaxed plan not minimizing the number of operators – may take very different decisions than a sequentially-optimal relaxed plan, thus constructing an "exit path" leading into the wrong direction. Example 8 in Appendix A.4 gives a full construction proving this.

Feeding Theorem 2 with non-optimal relaxed plans can of course also be imprecise "in the other direction", i.e., Theorem 2 may not apply although it does apply for an optimal relaxed plan. Thus "good cases" may go unrecognized. We demonstrate this with a simple modification of Example 8, explained below the example in Appendix A.4. Importantly, as we will point out in Section 8, our empirical results suggest that this weakness does not tend to occur in practice, at least as far as represented by the benchmarks.

## 6. Conservative Approximations

We now identify sufficient criteria guaranteeing that the prerequisites of Theorem 2 hold true. We consider both the local case where a particular state $s$ is given, and the global case where the criterion implies the prerequisites of Theorem 2 for *every* state $s$ in the task at hand. We approximate optimal rplan dependency graphs as follows:

**Definition 3.** *Let $(X, s_I, s_G, O)$ be a planning task, let $s \in S$ with $0 < h^+(s) < \infty$, let $x_0 \in X_{s_G}$, and let $t_0 = (s(x_0), c)$ be a relevant transition in $DTG_{x_0}$ with $o_0 := \text{rop}(t_0)$.*

*A local dependency graph for $s$, $x_0$, and $o_0$, or local dependency graph in brief, is a graph $lDG = (V, A)$ with unique leaf vertex $x_0$, and where $x \in V$ and $(x, x') \in A$ if either: $x' = x_0$, $x \in X_{\text{pre}_{o_0}}$, and $\text{pre}_{o_0}(x) \neq s(x)$; or $x' \in V \setminus \{x_0\}$ and $(x, x')$ is an arc in $SG$.*

*A global dependency graph for $x_0$ and $o_0$, or global dependency graph in brief, is a graph $gDG = (V, A)$ with unique leaf vertex $x_0$, and where $x \in V$ and $(x, x') \in A$ if either: $x' = x_0$ and $x_0 \neq x \in X_{\text{pre}_{o_0}}$; or $x' \in V \setminus \{x_0\}$ and $(x, x')$ is an arc in $SG$*

---

9. We remark that, due to the mentioned sources of over-estimation in $\text{cost}^{\text{d*}}$, constructing such an example requires fairly awkward constructs that do not appear likely to occur in practice.

If an optimal relaxed plan $P^+(s)$ for $s$ contains $o_0$, then $oDG^+$ as per Definition 1 will be a sub-graph of $lDG$ and $gDG$ as defined here. This is simply because any optimal rplan dependency graph has only arcs $(x, x')$ contained in the support graph of the task.[10] As previously indicated, the support graph may contain a lot more arcs than actually necessary. $SG$ captures what may ever support what else, not what will support what else *in an optimal relaxed plan*. Consider our earlier point that, when constructing $oDG^+$, we take into account only the operators *in front of* $o_0$ in $P^+(s)$. This information is not contained in $SG$, thus in Gripper we get the aforementioned cycle dropping a ball to support "free-gripper" for picking up the same ball.

The reader who has waded through the cumbersome details in the previous section will be delighted to hear that defining when an $lDG$ respectively $gDG$ is successful does not involve any additional notation:

**Definition 4.** *Let* $(X, s_I, s_G, O)$, $s$, $x_0$, $t_0$, $o_0$, *and* $G = lDG$ *or* $G = gDG$ *be as in Definition 3. We say that* $G = (V, A)$ *is* successful *if all of the following hold:*

*(1) $G$ is acyclic.*

*(2) If $G = lDG$ then $s_G(x_0) \neq s(x_0)$, and there exists no transitive successor $x'$ of $x_0$ in $SG$ so that $x' \in X_{s_G}$ and $s_G(x') \neq s(x')$.*

*(3) We have that $t_0$ either:*

  *(a) has self-irrelevant side effect deletes; or*

  *(b) has replaceable side effect deletes; or*

  *(c) has recoverable side effect deletes.*

*(4) For $x \in V \setminus \{x_0\}$, all $DTG_x$ transitions either are irrelevant, or have self-irrelevant deletes, or are invertible and have irrelevant side effect deletes and no side effects on $V \setminus \{x_0\}$.*

Consider first only local dependency graphs $G = lDG$; we will discuss $G = gDG$ below. Assume that we have an optimal relaxed plan $P^+(s)$ for $s$ that contains $o_0$, and thus $oDG^+$ is a sub-graph of $lDG$. Then condition (1) obviously implies Definition 2 condition (1). Condition (4) implies Definition 2 condition (3) because $oDTG_x^+$ does not contain any irrelevant transitions. Condition (2) implies that (*) $s(x_0)$ *is not $oDG^+$-relevant*, i.e., $s(x_0)$ is not needed in the rest of the relaxed plan. This is simply because no other un-achieved goal depends on $x_0$. With (*), condition (3a) implies Definition 2 condition (2a), because $R_1^+ \cap C_0 = \emptyset$, in the notation introduced previously. Conditions (3b) and Definition 2 condition (2b), respectively (3c) and Definition 2 condition (2c), are equivalent given (*).

Regarding exit distance, we do not know which parts of the domain transition graphs of the variables $x \in V \setminus \{x_0\}$ will be traversed by $P^+(s)$. An obvious bound on $\text{diam}(oDTG_x^+)$ is the length $\text{maxPath}(DTG_x)$ of a longest non-redundant path through the graph (a path visiting each vertex at most once). Unfortunately, we cannot compute $\text{maxPath}(.)$ efficiently. A Hamiltonian path (Garey & Johnson, 1979) exists in a graph $G = (V, A)$ iff

---

10. For $gDG$, note that $\text{pre}_{o_0}(x_0)$, if defined, will be $= s(x_0)$ and thus $x_0$ does not need to be recorded as its own predecessor.

maxPath$(G) = |V| - 1$. Thus the corresponding decision problem is **NP**-hard. Torch-Light over-approximates maxPath$(G)$ simply by $|V| - 1$. However, we can sometimes use diam$(DTG_x)$ instead of maxPath$(DTG_x)$, namely if we are certain that $x$ is one of the variables $V^*$ used in the definition of cost$^{d*}(oDG^+)$. This is certain if:

> (**) *all $DTG_x$ transitions either are irrelevant, or are invertible and have empty conditions, irrelevant side effect deletes, and no side effects on $V \setminus \{x_0\}$.*

Note that this is a strictly stronger requirement than Definition 4 condition (4). Clearly, it implies Definition 2 condition (3) as well as condition (*) in Section 5. Denote by $V^{**}$ the subset of $V \setminus \{x_0\}$ for which (**) holds. We define cost$^{D*}(G) := \sum_{x \in V}$ cost$^{D*}(x)$, where cost$^{D*}(x) :=$

$$\begin{cases} 1 & x = x_0 \\ \text{maxPath}(DTG_x) * \sum_{x':(x,x') \in A} \text{cost}^{D*}(x') & x \neq x_0, x \notin V^{**} \\ \text{diam}(DTG_x) * \sum_{x':(x,x') \in A} \text{cost}^{D*}(x') & x \neq x_0, x \in V^{**} \end{cases}$$

Because $x_0$ must move – to attain its own goal – every optimal relaxed plan must take at least one transition leaving $s(x_0)$. Thus, with Theorem 2 and the above, we have that:

**Theorem 3.** *Let $(X, s_I, s_G, O)$ be a planning task, and let $s \in S$ be a state with $0 < h^+(s) < \infty$. Say that $x_0 \in X$ so that, for every $o_0 = \text{rop}(s(x_0), c)$ in $DTG_{x_0}$ where $(s(x_0), c)$ is relevant, $lDG_{o_0}$ is a successful local dependency graph. Then s is not a local minimum, and $ed(s) \leq \max_{o_0} \text{cost}^{D*}(lDG_{o_0})$. If, for every $lDG_{o_0}$, we have Definition 4 condition (3a) or (3b), then $ed(s) \leq \max_{o_0} \text{cost}^{D*}(lDG_{o_0}) - 1$.*

Theorem 3 is our tool for guaranteed local analysis (II). For guaranteed global analysis (I), we simply look at the set of *all* global dependency graphs $gDG$, requiring them to be successful. In particular, all $gDG$ are then acyclic, from which it is not difficult to deduce that any non-goal state $s$ will have a variable $x_0$ fulfilling Definition 4 (2). For that $x_0$, we can apply Theorem 3 and thus get:

**Theorem 4.** *Let $(X, s_I, s_G, O)$ be a planning task. Say that all global dependency graphs $gDG$ are successful. Then $S$ does not contain any local minima and, for any state $s \in S$ with $0 < h^+(s) < \infty$, $ed(s) \leq \max_{gDG} \text{cost}^{D*}(gDG)$. If, for every $gDG$, we have Definition 4 condition (3a) or (3b), then $ed(s) \leq \max_{gDG} \text{cost}^{D*}(gDG) - 1$.*

The full proofs of Theorems 3 and 4 are in Appendix A.3. If $SG$ is acyclic and all transitions are invertible and have no side effects, then Theorem 4 applies, whereby we have now in particular proved our basic result. Vice versa, note that, if Theorem 4 applies, then $SG$ is acyclic. As far as local minima are concerned, one may thus reformulate Theorem 4 in simpler terms not relying on a notion of "successful dependency graphs". Apart from allowing to also determine an exit distance bound, the present formulation already paves the way for future research: a $gDG$ is defined relative to a concrete variable $x_0$ and operator $o_0$, and may thus allow for more accurate analysis of which other variables may actually become important for $x_0$ and $o_0$, in an optimal relaxed plan.

The use of diam$(DTG_x)$ instead of maxPath$(DTG_x)$ in cost$^{D*}(.)$, for the variables in $V^{**}$, has a rather significant effect on the quality of the bounds computed in many benchmarks. A typical example is a transportation domain where vehicle positions are leaf variables in $SG$ whose transitions have no side effects. Such variables qualify for $V^{**}$. Using $\text{maxPath}(DTG_x)$ instead, we would obtain exceedingly large bounds even for trivial road maps. For example, consider Logistics where the road map is fully connected. We have $\text{diam}(DTG_x) = 1$ and thus $\text{cost}^{\text{D}*}(.)$ delivers the correct bound 1. Using $\text{maxPath}(DTG_x)$ we instead get the bound $N - 1$, $N$ being the total number of locations in $DTG_x$.

Note that, within the scope of Theorem 4, i.e., the class of planning tasks to which Theorem 4 applies, plan existence is tractable. Namely, there exists a plan for the task iff there exists a relaxed plan for the initial state. This is because, starting from an optimal relaxed plan, we are guaranteed to be able to construct an exit path; iterating this argument gets us to the goal. In our view, this tractability is a *weakness* of this form of global analysis. The analysis does not apply in intractable classes of tasks that do not contain local minima. Note that such classes do exist, cf. Theorem 1. On the other hand, plan existence is tractable in all known benchmark domains where local minima are absent, so in practice this does not appear to be a major limitation. Also, note that plan construction, as well as optimal planning, are still intractable within the scope of Theorem 4. Plan construction is intractable because the plans may be exponentially long, cf. Example 6 in Appendix A.4. As for optimal planning, just consider Logistics and Miconic-STRIPS. We will see shortly (Proposition 1, next section) that these are fully covered by Theorem 4. However, in both of them, deciding bounded plan existence is **NP**-hard (Helmert, 2003).

Interestingly, the fact that Theorem 2, and therewith indirectly also Theorem 4, rely on *optimal* relaxed plans is *not* a source of intractability of plan construction here. If Theorem 4 applies, then any non-redundant relaxed plan $P^+$ has a successful $oDG^+$, enabling us to construct a path to a state where that particular relaxed plan (although not necessarily an optimal relaxed plan) can be shortened. Iterating this argument gives us a constructive method for obtaining a plan, where the only worst-case exponential behavior lies in the length of the individual path segments. That said, of course the plan constructed in this way may be highly non-optimal. Indeed, as is shown in Example 7 in Appendix A.4, this plan may be exponentially longer than an optimal plan. Thus, even if Theorem 4 applies and we do not need an optimality guarantee, running a planner still makes sense.

We will discuss the relation of the scope of Theorem 4 to known tractable classes in Section 9. A basic fact is that one can construct local minima even in very small examples involving only two variables and complying with our basic result except that either the support graph is cyclic (Example 2, Appendix A.4), or there is a non-invertible transition whose own delete is relevant (Example 3, Appendix A.4), or there is a transition with a relevant side effect delete (Example 4, Appendix A.4). These examples are contained in many known tractable classes, thus underlining that the automatic analysis of $h^+$ topology and the identification of tractable classes are different (although related) enterprises.

## 7. Benchmark Performance Guarantees

We now state some guarantees that our analyses (I)–(III) give in benchmark domains. The underlying finite-domain variable formalizations are straightforward, and correspond to formulations that can be found automatically by Fast Downward. They are listed in Appendix A.5, where we also give the proofs of the following two simple observations.[11]

In four of our benchmark domains, guaranteed global analysis (I) will always succeed :

**Proposition 1.** *Let $(X, s_I, s_G, O)$ be a planning task from the Logistics, Miconic-STRIPS, Movie, or Simple-TSP domain. Then Theorem 4 applies, and the bound delivered is at most 1, 3, 1, and 1 respectively.*

It follows trivially from Proposition 1 that guaranteed local analysis (II) succeeds in these domains as well. If $s$ is any state in one of the four listed domains, then Theorem 3 applies to $s$, and the bound delivered is as stated.

Note that the bounds for Logistics and Movie are the correct ones, i.e., they are tight. For Miconic-STRIPS, the over-estimation of the actual bound (which is 1, not 3) arises because the analysis does not realize that boarding a passenger can be used as the leaf variable $x_0$. For Simple-TSP, the correct bound is 0 (since $h^+$ is the exact goal distance). The over-estimation arises because, in every goal variable $x_0 =$"visited(location)", the $gDG$ includes also the variable "at", not realizing that the value of "at" does not matter because any location can be visited from any other one.

For the transportation benchmarks involving capacity constraints, approximate local analysis (III) will always succeed, if provided with suitable optimal relaxed plans:

**Proposition 2.** *Let $(X, s_I, s_G, O)$ be a planning task from the Elevators, Ferry, Gripper, or Transport domain, and let $s \in S$. In Ferry and Gripper, for every optimal relaxed plan $P^+(s)$ there exists $oDG^+$ so that Theorem 2 applies, the bound being at most 1. In Elevators and Transport, there exists at least one $P^+(s)$ and $oDG^+$ so that Theorem 2 applies, the bound being at most 1 in Elevators and at most the road map diameter in Transport.*

The relevant deletes of $t_0$, in all these cases, are due to the effects decreasing the remaining vehicle capacity, like "free-gripper" in the Gripper domain. A decrease of capacity is always due to a "load" type of operator, which is matched by an "unload" type of operator later on inside the relaxed plan. Thus these deletes are always recovered inside $P^+(s)$ (we have Definition 2 condition (2a)). Further, relaxed plans never use an "unload" action to free a capacity for "load"ing the same object, thus the $oDG^+$s are cycle-free. Hence the $oDG^+$ are successful, and Theorem 2 applies. For Elevators and Transport, Proposition 2 is slightly weaker because a vehicle may have capacity $> 1$, allowing – but not forcing – relaxed plans to use unloading operators recovering a capacity not actually present.

We note that similar patterns are likely to occur in any domain with renewable resources, and will be recognized by Definition 2 condition (2a) in the same way.

Proposition 2 does not hold for Theorems 3 and 4, i.e., for $lDG$s and $gDG$s. This is due to two deficiencies (cf. the discussion at the end of Section 4). First, $SG$ contains cycles "unload"ing an object in order to free the capacity for "load"ing it. Second, Definition 2 condition (3a) is more restrictive than Definition 2 condition (2a), postulating the deletes of $t_0$ to be entirely irrelevant. If we had a way of removing these deficiencies, then the guaranteed analyses (I,II) would succeed in the four domains from Proposition 2.

---

11. We say "can be found automatically" here because Fast Downward's translator is not deterministic, i.e., it may return different finite-domain variable encodings even when run several times on the same planning task. Some but not all of these encodings correspond to our domain formalizations. For Elevators, we do not give a full definition because, without action costs, this is merely a variant of Transport.

# 8. Experiments

We report on a large-scale experiment with TorchLight. We fill in a few details on TorchLight's implementation, and we describe a simple alternative analysis technique based on search probing. We explain the experiments set-up, report runtime results for the different stages of TorchLight, and describe TorchLight's analysis results on a per-domain basis. We assess the quality of that analysis in terms of its predictive capability. We finally summarize the outcome of TorchLight's diagnosis facility in our benchmarks.

## 8.1 TorchLight

TorchLight is implemented in C based on FF.[12] TorchLight currently handles STRIPS only, i.e., no ADL domains. It uses Fast Downward's translator (Helmert, 2009) to find the finite-domain variables. Establishing the correspondence between these variables (respectively their values) and FF's internally used ground facts is mostly straightforward. There are a few details to take care of; we omit these for brevity.

After parsing Fast Downward's variables, TorchLight creates data structures representing the support graph and the domain transition graphs. It then enters a phase we refer to as *static analysis*, where it determines fixed properties such as, for every transition $t$, whether $t$ is irrelevant, invertible, etc. The next step is guaranteed global analysis (I), checking the preconditions of Theorem 4 by enumerating all global dependency graphs and testing whether they are successful. To be able to report the percentage of successful $gDG$s, we do not stop at the first unsuccessful one.

The local analysis techniques – guaranteed local analysis (II) using Theorem 3 and approximate local analysis (III) using Theorem 2 – are run on a set $LS$ of states comprising the initial state as well as a number $R$ of sample states obtained by random walks starting in $s_I$. The set $LS$ is identical for both analyses, and we run each technique on each state $s \in LS$ regardless of what the outcome of running the respective other technique on $s$ is. Given $s$, analysis (II) checks Theorem 3 by constructing the local dependency graph for every suitable variable $x_0$ and every transition $t_0$ leaving $s(x_0)$. If we find a non-successful $t_0$, we stop considering $x_0$. We minimize exit distance bounds across different $x_0$.

Analysis (III) checks Theorem 2 on a relaxed plan $P^+(s)$ computed by FF's heuristic function. In case that no relaxed plan exists for $s$, the analysis reports failure. Otherwise, the analysis proceeds over all operators $o_0$ in $P^+(s)$, from start to end, and over all variables $x_0$ affected by $o_0$. For each pair $o_0, x_0$ we build the optimal rplan dependency graph $oDG^+$ as per Definition 1. We skip variables $x_0$ where $\mathrm{eff}_{o_0}(x_0)$ is not actually used as a precondition or goal, in the rest of $P^+(s)$. If $oDG^+$ is successful, we stop. (Relaxed plans can be big in large examples, so continuing the analysis for exit bound minimization was sometimes costly.) As mentioned in Section 5, before we build $oDG^+$ we re-order $P^+(s)$ by moving operators behind $o_0$ if possible. This is of paramount importance because it avoids including unnecessary variables into $oDG^+$. The re-ordering process is straightforward. It starts at the direct predecessor $o$ of $o_0$, and tests whether $P^+(s)$ is still a relaxed plan when moving $o$ directly behind $o_0$. If yes, this arrangement is kept. Then we iterate to the predecessor of $o$, and so forth. It is easy to see that, this way, $oDG^+$ will contain exactly the variables and transitions used in $P^+(s)$ to achieve $\text{pre}_{o_0}$. Finally, when we check whether the $oDG^+$-relevant deletes of $t_0$ are $P_{>0}^+(s)$-recoverable, we use a simple technique allowing to recognize situations where failure due to one operator can be avoided by replacing with an alternative operator. For example, if in Transport $o_0$ is a loading operator reducing capacity level $k$ to $k-1$, then $P^+(s)$ may still contain an unloading operator relying on level $k$. Thus level $k$ will be contained in $R_1^+ \cap C_0$, causing failure. However, the unloading can just as well be performed based on capacity level $k-1$, removing this difficulty. We catch cases like this during construction of $R_1^+$. Whenever we find $o$ whose precondition overlaps $C_0$, we test whether we can replace $o$ with a similar operator.

---

12. The source code of TorchLight is an online appendix to this paper. It is available for download also at `http://www.loria.fr/~hoffmanj/TorchLight.zip`.

The local analyses return simple statistics, namely the minimum, mean, and maximal exit distance bound found, as well as the *success rate*, i.e., the fraction of sample states where guaranteed local analysis (II)/approximate local analysis (III) succeeded. Analysis (III) success rates will be a main focus, because these turn out to be very informative.

We run $R = 1, 10, 100, 1000$ in the experiment. The length of each random walk is chosen uniformly between 0 and $5 * h^{\text{FF}}(s_I)$, i.e., 5 times the FF heuristic value for the initial state. We do not play with the parameter 5. It is important, however, that this parameter is not chosen too small. In domains with many dead ends – where one may do things that are fatally wrong – it is likely that the "bad" things will happen only if doing a sufficiently large number of random choices. Consequently, the *dead-end rate*, i.e., the fraction of sample states for which no relaxed plan exists, tends to be larger for longer random walks. Since analysis (III) fails on states that have no relaxed plan, this exerts an important influence on analysis (III) success rates. We illustrate this below by comparing some results for sampled states to results obtained using the initial states only.

## 8.2 Search Probing

For approximate analysis of sample states, there exists a simple (and rather obvious) alternative to TorchLight's causal graph based technology. One can use search to determine whether or not a given sample state $s$ is a local minimum, and what its exit distance is. Since we cannot compute $h^+$ effectively, such a search-based analysis is necessarily approximate. The straightforward method is to replace $h^+$ with a relaxed-plan based approximation. Herein, we replace $h^+$ with $h^{\text{FF}}$, i.e., with FF's heuristic function. Precisely, given a state $s$, we run a single iteration of FF's Enforced Hill-Climbing, i.e., a breadth-first search for a state with better heuristic value. In this search, like FF does, we use helpful actions pruning to avoid huge search spaces. Unlike FF, to focus on the detection of states not on local minima, we allow only monotone paths (thus restricting the search space to states $s'$ where $h^{\text{FF}}(s') = h^{\text{FF}}(s)$). We refer to this technique as *search probing*, *SP* in brief. We also experiment with a variant imposing a 1 second runtime cut-off on the search. We refer to this as *limited search probing*, $SP^{1s}$ in brief. SP and $SP^{1s}$ are run on the same set $LS$ of states as TorchLight's local analyses (II,III).

As it turns out, empirically – in the present benchmarks – SP and $SP^{1s}$ are very competitive with TorchLight's analysis (III). Since that analysis is a main focus of our experiments, it is relevant to understand the commonalities and differences between these techniques.

As far as analysis quality guarantees are concerned, all 3 techniques – analysis (III), SP, $SP^{1s}$ – have similar properties: there are no guarantees whatsoever. Each may report success although $s$ is a local minimum (false positives), and each may fail although $s$ is not a local minimum (false negatives). In all cases, false positives are due to the use of non-optimal relaxed plans ($h^{FF}$ instead of $h^+$). False negatives are inherent in analysis (III) because this covers only certain special cases; they are inherent in SP$^{1s}$ due to the search limit. SP can have false negatives due to helpful actions pruning, however that could in principle be turned off; the more fundamental source of false negatives are the non-optimal relaxed plans. These are also responsible for a lack of connections across the techniques. The only implication is the trivial one that SP$^{1s}$ success on a state $s$ implies SP success on $s$. In particular, if analysis (III) correctly identifies $s$ to not be a local minimum, then this does *not* imply that SP will do so as well. The causal graph analysis may be less affected by irregularities in the $h^{FF}$ surface. This happens, for example, in the Transport domain of IPC 2008, resulting in higher success rates for analysis (III).

There are some obvious – but important – differences regarding runtime performance and the danger of false negatives. SP runtime is worst-case exponential in the size of the (grounded) input, whereas analysis (III) and SP$^{1s}$ runtime is low-order polynomial in that size. For SP, decreasing the number $R$ of sample states merely reduces the chance of hitting a "bad" state (a sample state on a large flat region), whereas analysis (III) and SP$^{1s}$ scale linearly in $R$. On the other hand, both analysis (III) and SP$^{1s}$ buy their efficiency with incompleteness, i.e., increased danger of false negatives. Analysis (III) simply recognizes only special cases. SP$^{1s}$ effectively bounds the lookahead depth, i.e., the search depth in which exit states can be detected.

As indicated, SP and SP$^{1s}$ turn out to be competitive in the benchmarks. Large search spaces are rare for SP. The success rates of SP and SP$^{1s}$ are similar, and as far as predictive capability is concerned are similarly informative as those of analysis (III). Thus good-quality success rates can be obtained with much simpler techniques than TorchLight.[13] This notwithstanding, (a) TorchLight has other functions – the guaranteed analyses (I,II) as well as diagnosis – that cannot be simulated, and (b) results in benchmarks only ever pertain to these examples. TorchLight's analysis (III) offers unlimited lookahead depth at low-order polynomial cost. This does not appear to matter much in the present benchmarks, but there are natural cases where it does matter. We get back to this below.

## 8.3 Experiments Set-Up

We run experiments in a set of 37 domains. These include the domains investigated in the hand-made analysis of $h^+$ topology (Hoffmann, 2005), as shown in Figure 1, which include all domains from the international planning competitions (IPC) up to IPC 2004. Our remaining domains are the STRIPS (versions of the) domains from IPC 2006 and IPC 2008, except IPC 2008 Cyber-Security which we omit due to parsing difficulties.[14] The test instances were collected from the IPC collection(s) where applicable (removing action cost constructs from the IPC 2008 domains), and randomly generated elsewhere. In total, our test set contains 1160 instances.

---

13. In particular, search probing appears to be a rather useful technique, raising the question why such techniques have not yet been used for performance prediction purposes. Roberts and Howe (2009), for example, use very simple features only. We get back to this in the conclusion.

14. The instances are too large for FF's parser in its standard configuration. When tweaking bison to allow larger parse trees, we obtained a segmentation fault even in the smallest instance of IPC 2008.

| tool/phase | single-shot/$R = 1$ | | $R = 10$ | | $R = 100$ | | $R = 1000$ | |
|---|---|---|---|---|---|---|---|---|
| | mean | max | mean | max | mean | max | mean | max |
| FD Translator | 6.12 | 690.59 | | | | | | |
| SG/DTG | 0.14 | 6.91 | | | | | | |
| Static Analysis | 0.25 | 31.42 | | | | | | |
| Analysis (I) | 0.40 | 53.29 | | | | | | |
| Sample States | 0.01 | 0.53 | 0.08 | 4.81 | 0.76 | 50.35 | 7.50 | 491.20 |
| Analysis (II) | 0.00 | 0.18 | 0.01 | 1.11 | 0.10 | 9.56 | 0.98 | 94.59 |
| Analysis (III) | 0.02 | 1.31 | 0.03 | 2.46 | 0.23 | 20.09 | 2.15 | 194.79 |
| TorchLight total | 6.92 | 727.63 | 7.04 | 736.98 | 8.00 | 807.70 | 17.57 | 1510.74 |
| TorchLight (III) | 6.52 | 724.54 | 6.64 | 732.98 | 7.51 | 795.16 | 16.19 | 1413.23 |
| TorchLight (III) no FD | 0.40 | 33.95 | 0.49 | 40.50 | 1.37 | 103.67 | 10.04 | 719.27 |
| SP | 0.06 | 58.02 | 0.23 | 138.54 | 5.47 | — | 26.24 | — |
| SP total | 0.07 | 58.03 | 0.32 | 138.59 | 6.23 | — | 33.74 | — |
| SP$^{1s}$ | 0.01 | 1.08 | 0.07 | 4.46 | 0.66 | 56.18 | 5.89 | 391.59 |
| SP$^{1s}$ total | 0.01 | 1.48 | 0.15 | 9.27 | 1.42 | 106.53 | 13.39 | 882.79 |
| FF | 268.20 | — | | | | | | |
| LAMA | 185.05 | — | | | | | | |

Table 2: Summary of runtime data. Mean/max is over all instances of all domains. For empty fields, the respective tool/phase is "single-shot", i.e., does not depend on $R$. A dash means time-out, 1800 seconds, which is inserted as the runtime for each respective instance into the mean computation. Rows "FD Translator" ... "Analysis (III)" time the different stages of TorchLight. "TorchLight total" is overall runtime, "TorchLight (III)" does not run analyses (II) and (III), "TorchLight (III) no FD" is the latter when disregarding the translation costs. "SP" determines a success rate (fraction of sample states deemed to not be on local minima) via search probing, i.e., search around each sample state; "SP$^{1s}$" imposes a 1 second time-out on these searches. "SP total" and "SP$^{1s}$ total" include the time for generating the sample states.

All experiments are run on a 1.8 GHZ CPU, with a 30 minute runtime and 2 GB memory cut-off. We run 4 different planners/tools. Apart from TorchLight (and SP/SP$^{1s}$), these include FF (Hoffmann & Nebel, 2001a), and LAMA (Richter et al., 2008; Richter & Westphal, 2010). The purpose of running these planners is to assess to what extent TorchLight's output – in particular analysis (III) success rate – can predict planner success or failure. To examine this also for a very plain planner, we also run a version of FF that uses no goal ordering techniques, and that runs only Enforced Hill-Climbing, without resorting to best-first search if that fails. We will refer to this planner as *EHC* in what follows.

## 8.4 Runtime

Our code is currently optimized much more for readability than for speed. Still, TorchLight is fast. Up to $R = 100$, the bottleneck is Fast Downward's translator. With $R = 1, 10, 100$, the actual analysis takes at most as much time as the translator in 99.74%, 99.74%, and 96.21% of the instances respectively. To assess this in more detail, consider Table 2 which gives the timing of the different stages of TorchLight, and of the other planners/tools.

The translation runtime sometimes hurts considerably, with a peak of 690.59 seconds in the most costly instance of the Scanalyzer domain. This is rather exceptional, however. The second most costly domain is Blocksworld-NoArm, with a peak of 138.33 seconds. In 20 of the 37 domains, the most costly instance is translated in less than 10 seconds. In 57.24% of the instances, Fast Downward's translator takes at most 1 second.

For static analysis, the peak behavior of 31.42 seconds (also in Scanalyzer) is even more exceptional: in 95.34% of the instances, static analysis takes at most 1 second. The second highest domain peak is 7.88 seconds in Pipesworld-Tankage. Similarly, while analysis (I) takes a peak of 53.29 seconds – in Blocksworld-NoArm – in 96.12% of the instances it completes in at most 1 second. The only domain other than Blocksworld-NoArm where the peak instance takes more than 10 seconds is Airport, with a peak of 41.71 seconds; the next highest domain peaks are Pipesworld-Tankage (6.8), Scanalyzer (2.91), Logistics (1.89), and Woodworking (1.17). In all other domains, analysis (I) always completes within a second.

Turning focus on the local analyses, we see that they are even more effective. In particular, we will concentrate below mostly on approximate local analysis (III). We will see that $R = 1000$ does not offer advantages over $R \leq 100$ as far as the information obtained goes, so we will mostly concentrate on $R \leq 100$. For $R = 1, 10, 100$, analysis (III) completes in at most 1 second for 99.66%, 99.40%, 95.60% of the instances respectively. For $R = 1000$ this still holds for 76.55% of the instances. The peak runtime of 20.09 seconds for $R = 100$ occurs in Scanalyzer. The next highest domain peaks are Blocksworld-NoArm (9.23), Pipesworld-Tankage (4.24), Ferry(3.21), Logistics (2.99), Blocksworld-Arm (2.77), Optical-Telegraph (1.97), and Airport (1.41). In all other 29 domains, analysis (III) with $R = 100$ always completes within a second.

The bottleneck in local analysis is the generation of sample states. This can be costly because it involves the repeated computation of applicable operators during the random walks. Its $R \leq 100$ peak of 50.35 seconds is in the Scanalyzer domain. However, once again, this peak behavior is exceptional. With $R = 1, 10, 100$, the sampling completes within at most 1 second for 100%, 98.28%, 87.41% of the instances respectively.

The main competitor of TorchLight analysis (III) success rates is search probing, i.e., SP and SP[1s]. Consider for the moment only the analysis methods themselves, i.e., row "Analysis (III)" vs. rows "SP" and "SP[1s]" in Table 2. Compared to SP[1s], analysis (III) is consistently in the advantage (except for maximum runtime with $R = 1$), but the difference is not dramatic. This is to be expected, given that SP[1s] trades completeness against a small fixed maximum runtime. Compared to the complete search in SP, analysis (III) consistently has a significant advantage. However, for $R \leq 10$ the mean runtime of SP is tolerable, and even the maximum runtime is not too bad. Further, bad runtime behavior is exceptional. For $R = 1, 10$, SP completes in at most 1 second for 99.83% and 98.45% of the instances respectively. In 35 ($R = 1$) respectively 32 ($R = 10$) of the 37 domains even the maximum runtime is below 1 second. With $R = 100$, SP has two time-outs, both in Blocksworld-Arm. With $R = 1000$, there are 11 time-outs, in Blocksworld-Arm, Blocksworld-NoArm, Freecell, and Pipesworld-NoTankage. With $R = 100$, the maximum runtime is above 10 seconds in 7 domains; with $R = 1000$, in 12. However, with $R = 100, 1000$, SP still completes in at most 1 second for 92.33% and 71.98% of the instances respectively (compared to 95.60% and 76.55% for analysis (III), cf. above).

Neither analysis (III) nor search probing are stand-alone methods. The former requires all of TorchLight except analyses (I,II). The latter requires the sampling of random states. The respective total data is given in rows "TorchLight (III)" and "SP total" / "SP[1s] total" in Table 2. Here the picture changes dramatically in favor of SP and especially SP[1s]. It should be noted, though, that this is mostly due to the overhead for the translation to finite-domain variables. This overhead is an artifact of the implementation. Our approach is defined for finite-domain variables, while the benchmarks are not, even though the finite-domain representation is in most cases more natural than the Boolean one. Further, many planners (notably Fast Downward and its quickly growing set of derivatives) use the translation anyway. The runtimes without translation are given in the row "TorchLight (III) no FD".

As one would hope and expect, the analysis methods are much faster than actual planners. LAMA has 112 time-outs in our test suite, FF has 173.

## 8.5 Analyzing Domains

We now discuss the actual analysis outcomes, on a per-domain basis. We first consider only TorchLight, then give some details on the comparison of analysis (III) success rates to those obtained by search probing. Before we begin, a few words are in order regarding the comparison between SP and $SP^{1s}$. With $R = 1, 10, 100, 1000$, the success rates are identical in 99.83%, 99.14%, 97.5%, 94.66% of our 1160 benchmark instances respectively; in 99.83%, 99.14%, 99.31%, 98.97% of the instances, the success rates differ by at most 5%. Thus, a small runtime cut-off does not adversely affect the success rates of search probing (because long searches are rare). This being so, we henceforth do not discuss the data for SP vs. $SP^{1s}$ separately. We compare TorchLight's analysis (III) success rates to those of SP only.

The guarantees of Proposition 1 are confirmed, i.e., guaranteed global analysis (I) succeeds as described in Logistics, Miconic-STRIPS, Movie, and Simple-TSP. It never succeeds in any other domain, though. In some domains, fractions of the $gDG$s are successful. Precisely, the maximum fraction of successful $gDG$s is 97% in Satellite, 50% in Ferry, 33.33% in TPP, 22.22% in Driverlog, 20% in Depots, 13.33% in Tyreworld, and 12.5% in Blocksworld-Arm. However, if the fraction is below 100% then nothing is proved, so this data may at best be used to give an indication of which aspects of the domain are "good-natured". Guaranteed local analysis (II) generally is not much more applicable than global analysis. Thus we now concentrate on approximate local analysis (III) exclusively.

Proposition 2 is backed up impressively. Even with $R = 1000$, analysis (III) succeeds in every single sample state of Ferry, Gripper, Elevators, and Transport.[15] This indicates strongly that the potentially sub-optimal relaxed plans do not result in a loss of information here. Indeed, the analysis yields high success rates in almost all domains where local minima are non-present or limited. This is not the case for the other domains, and thus TorchLight can distinguish domains with "easy" $h^+$ topology from the "hard" ones. Consider Figure 3, showing mean analysis (III) success rates per-domain with $R = 1$. (The picture is similar for $R = 10, 100, 1000$; cf. Table 3 below.)

The domains whose $h^+$ topology is not known are shown separately on the right hand side in Figure 3. For the other domains, we see quite nicely that "harder" domains tend to have lower success rates. In particular, the easiest domains in the bottom class all have 100% success rates (95% in the case of Zenotravel), whereas the hardest domains in the top right corner only have around 50% or less. In the latter domains, to some extent the low success rates result from the recognition of dead ends by FF's heuristic function. For example, if during random sampling we make random vehicle moves consuming fuel, like in Mystery and Mprime, then of course chances are we will end up in a state where fuel is so scarce that even a relaxed plan does not exist anymore. This is most pronounced in Airport, where *all* sample states here have infinite heuristic values. However, the capabilities of the analysis go far beyond counting states on recognized dead ends. In Blocksworld-Arm, for example, there are no dead ends at all and still the success rate is only 30%, clearly indicating this as a domain with a difficult topology.

---

15. Historically, this observation preceded Proposition 2, as well as the $h^+$ topology categorization of Elevators and Transport as per Figure 1. That is, these hand-made analyses were motivated by observing TorchLight's analysis outcome.

Figure 3: Overview of TorchLight domain analysis results. "*": guaranteed global analysis (I) always succeeds. "+": approximate local analysis (III) always succeeds if provided an optimal relaxed plan. Numbers shown are mean success rates per domain, for approximate local analysis (III) with $R = 1$, i.e., when sampling a single state per domain instance.

To some extent, based on the success rates we can even distinguish Pipesworld-Tankage from Pipesworld-NoTankage, and Mprime from Mystery (in Mprime, fuel can be transferred between locations). The relatively high success rate in Depots probably relates to its transportation aspects. In Grid, in 20% of cases our analysis is not strong enough to recognize the reasons behind non-existence of local minima; these reasons can be quite complicated (Hoffmann, 2003). Dining-Philosophers does not really have a favorable $h^+$ topology. Its rather excessive bound 31 is due to the very particular domain structure where philosophers behave in strictly symmetrical ways (Hoffmann, 2005). Apart from this, the only strong outliers are Driverlog, Rovers, Hanoi, and Blocksworld-NoArm. All of these are more problems of the hand-made analysis than of TorchLight's. In Driverlog and Rovers, deep local minima do exist, but only in awkward situations that don't tend to arise in the IPC instances. Thus the hand-made analysis, which is of a worst-case nature, is too pessimistic here. The opposite happens in Hanoi and Blocksworld-NoArm, where the absence of local minima is due to rather idiosyncratic reasons. For example, in Hanoi the reason is that $h^+$ is always equal to the number of discs not yet in goal position – in the relaxation, one can always accomplish the remaining goals one-by-one, regardless of the constraints entailed by their positioning. Hanoi and Blocksworld-NoArm are not actually "easy to solve" for FF, and in that sense, from a practical perspective, the low success rates of TorchLight's analysis (III) provide the more accurate picture.

| domain | $s_I$ (III) | $R=1$ (III) | SP | $R=10$ (III) | SP | $R=100$ (III) | SP | $R=1000$ (III) | SP | DE |
|---|---|---|---|---|---|---|---|---|---|---|
| Airport | 96.0 | 0.0 | 0.0 | 2.0 | 2.0 | 2.8 | 2.9 | 2.9 | 3.0 | 97.0 |
| Blocks-Arm | 38.3 | 30.0 | 93.3 | 28.2 | 94.5 | 26.9 | 91.7 | 26.5 | 82.1 | 0 |
| Blocks-NoArm | 70.0 | 56.7 | 100 | 57.2 | 100 | 55.9 | 99.9 | 56.2 | 98.3 | 0 |
| Depots | 100 | 81.8 | 100 | 85.9 | 99.1 | 86.3 | 99.7 | 86.2 | 99.6 | 0 |
| Din-Phil | 100 | 24.1 | 27.6 | 22.8 | 23.1 | 22.8 | 22.9 | 22.0 | 22.3 | 77.2 |
| Driverlog | 100 | 100 | 100 | 97.5 | 100 | 97.4 | 99.9 | 97.9 | 99.8 | 0 |
| Elevators | 100 | 100 | 100 | 100 | 100 | 100 | 100 | 100 | 100 | 0 |
| Ferry | 100 | 100 | 100 | 100 | 100 | 100 | 100 | 100 | 100 | 0 |
| Freecell | 97.5 | 55.0 | 60.0 | 57.4 | 62.8 | 57.9 | 63.5 | 58.0 | 63.2 | 35.4 |
| Grid | 60.0 | 80.0 | 100 | 74.0 | 92.0 | 69.0 | 93.8 | 69.5 | 93.5 | 0 |
| Gripper | 100 | 100 | 100 | 100 | 100 | 100 | 100 | 100 | 100 | 0 |
| Hanoi | 0.0 | 0.0 | 33.3 | 11.1 | 44.4 | 10.2 | 41.9 | 10.6 | 41.9 | 0 |
| Logistics | 100 | 100 | 100 | 100 | 100 | 100 | 100 | 100 | 100 | 0 |
| Miconic | 100 | 100 | 100 | 100 | 100 | 100 | 100 | 100 | 100 | 0 |
| Movie | 100 | 100 | 100 | 100 | 100 | 100 | 100 | 100 | 100 | 0 |
| Mprime | 74.3 | 48.6 | 74.3 | 61.1 | 76.3 | 64.3 | 79.0 | 64.1 | 78.2 | 7.2 |
| Mystery | 75.0 | 39.3 | 42.9 | 37.1 | 43.9 | 37.6 | 45.6 | 36.3 | 44.4 | 46.8 |
| Opt-Tele | 0 | 7.1 | 14.3 | 1.4 | 2.9 | 0.9 | 1.4 | 1.1 | 1.7 | 98.3 |
| Pipes-NoTank | 40.0 | 76.0 | 98.0 | 75.4 | 97.4 | 75.2 | 97.4 | 75.1 | 95.4 | 0 |
| Pipes-Tank | 34.0 | 40.0 | 92.0 | 50.6 | 90.0 | 49.4 | 88.1 | 48.7 | 88.2 | 8.7 |
| PSR | 66.0 | 50.0 | 62.0 | 57.6 | 69.8 | 58.3 | 71.1 | 57.0 | 70.4 | 0 |
| Rovers | 100 | 100 | 100 | 100 | 99.5 | 100 | 99.8 | 100 | 99.8 | 0 |
| Satellite | 85 | 100 | 100 | 98.5 | 100 | 98.4 | 100 | 98.0 | 99.8 | 0 |
| Simple-TSP | 100 | 100 | 100 | 100 | 100 | 100 | 100 | 100 | 100 | 0 |
| Transport | 100 | 100 | 93.3 | 100 | 93.0 | 100 | 94.8 | 100 | 94.4 | 0 |
| Tyreworld | 100 | 100 | 100 | 95.6 | 100 | 96.3 | 100 | 95.5 | 100 | 0 |
| Zenotravel | 90 | 95 | 100 | 94.5 | 99.5 | 95.8 | 98.4 | 95.4 | 98.2 | 0 |
| Openstacks | 100 | 0 | 4.4 | 14.8 | 21.3 | 17.7 | 22.0 | 16.6 | 20.8 | 79.1 |
| Parc-Printer | 100 | 3.3 | 6.7 | 8.0 | 8.3 | 6.3 | 7.2 | 6.0 | 6.8 | 93.0 |
| Pathways | 100 | 10.0 | 10.0 | 6.0 | 6.0 | 5.4 | 5.4 | 4.6 | 4.6 | 95.3 |
| Peg-Sol | 0 | 0 | 10 | 13.3 | 22.7 | 13.1 | 22.3 | 12.6 | 22.2 | 75.2 |
| Scanalyzer | 0 | 30.0 | 96.7 | 33.0 | 99.7 | 33.5 | 97.9 | 33.9 | 98.5 | 0 |
| Sokoban | 30.0 | 13.3 | 33.3 | 20.3 | 38.3 | 19.1 | 38.2 | 18.5 | 37.7 | 54.2 |
| Storage | 100 | 93.3 | 96.7 | 89.0 | 96.3 | 89.8 | 96.8 | 89.3 | 96.9 | 0 |
| TPP | 100 | 80.0 | 80.0 | 68.0 | 67.0 | 65.4 | 63.8 | 65.5 | 63.9 | 34.5 |
| Trucks | 56.3 | 0 | 0 | 2.5 | 3.1 | 1.9 | 2.9 | 1.4 | 2.7 | 97.3 |
| Woodworking | 100 | 13.3 | 13.3 | 14.3 | 14.3 | 15.3 | 15.4 | 15.3 | 15.4 | 84.6 |

Table 3: Mean success rates per domain. Upper part: domains whose $h^+$ topology was previously examined by hand (Hoffmann, 2005) or is trivial to examine based on these results; lower part: IPC 2006/2008 domains where that is not the case. Columns "$s_I$" show data for analyzing the initial state only, columns "$R = 1, 10, 100, 1000$" for analyzing the respective number of sample states. Columns "(III)" give data for approximate local analysis (III), columns "SP" give data for search probing, column "DE" gives dead-end rates for $R = 1000$.

Table 3 gives a complete account of per-domain averaged success rates data, including all domains, all values of $R$, the rates obtained on initial states, and using SP instead of TorchLight. This serves to answer three questions:

(1) *Is it important to sample random states, rather than only analyzing the initial state?*

(2) *Is it important to sample many random states?*

(3) *How competitive is analysis (III) with respect to a search-based analysis?*

The answer to question (1) is a clear "yes". Most importantly, this pertains to domains with dead ends, cf. our brief discussion above. It is clear from Table 3 that, in such domains, analyzing $s_I$ results in a tendency to be too optimistic. To see this, just consider the entries for Airport, Dining-Philosophers, Freecell, Mystery, Openstacks, Parc-Printer, Pathways, TPP, Trucks, and Woodworking. All these domains have dead ends, for a variety of reasons. The dead ends do not occur frequently at initial state level, but do occur frequently during random walks – cf. column "DE" in Table 3. (Interestingly, in a few domains – most notably the two Pipesworlds – the opposite happens, i.e., success rates are lower for $s_I$ than for the sample states. It is not clear to us what causes this phenomenon.)

If we simply compare the $s_I$ column with the $R = 1000$ column for analysis (III), then we find that the result is "a lot" different – more than 10% – in 22 of the 37 domains. To some extent, this difference between initial states and sample states may be just due to the way these benchmarks are designed. Often, the initial states of every instance are similar in certain ways (no package loaded yet, etc). On the other hand, it seems quite natural, at least for offline problems, that the initial state is different from states deeper down in the state space (consider transportation problems or card games, for example).

The answer to question (2) is a clear "no". For example, compare the $R = 1$ and $R = 1000$ columns for analysis (III). The difference is greater than 10% in only 6 of the 37 domains. The peak difference is in Openstacks, with 16.6% for $R = 1000$ vs. 0% for $R = 1$. The average difference over all domains is 4.17%. Similarly, comparing the $R = 1$ and $R = 1000$ columns for SP results in only 5 of 37 domains where the difference is greater than 10%, the peak being again in Openstacks, 20.8% for $R = 1000$ vs. 4.4% for $R = 1$. The average difference over all domains is 3.7%.

The answer to question (3) is a bit more complicated. Look at the columns for analysis (III) respectively SP with $R = 1000$. The number of domains where the difference is larger than 10% is now 11 out of 37, with a peak of 64.6% difference in Scanalyzer. On the one hand, this still means that in 26 out of 37 domains the analysis result we get is very close to that of search (average difference 2.18%), without actually running any search! On the other hand, what happens in the other 11 domains? In all of these, the success rate of SP is higher than that of TorchLight. This is not surprising – it basically means that TorchLight's analysis is not strong enough here to recognize all states that are not on local minima.

Interestingly, this weakness can turn into an unexpected advantage. Of the 11 domains in question, 8 domains – Blocksworld-Arm, Depots, Mprime, Pipesworld-Tankage, Pipesworld-NoTankage, PSR, Scanalyzer, and Sokoban – do contain deep local minima.[16] Thus, in these 8 domains, we would wish our analysis to return small success rates. TorchLight grants this wish much more than SP does. Consider what happens when using SP instead of analysis (III) in Figure 3. For Mystery, PSR, and Sokoban, the change is not dramatic. However, Blocksworld-Arm is marked with average success rate 93 instead of 30, putting it almost on par with the very-simple-topology domains in the bottom class. Similarly, Pipesworld-Tankage, Pipesworld-NoTankage, and Scanalyzer are put almost on par with these. Depots actually receives a 100, putting it exactly on par with them. Thus the SP analysis outcome actually looks quite a bit worse, in 5 of the domains.

---

16. Sokoban has unrecognized dead-ends (in the relaxation, blocks can be pushed across each other) and therefore local minima. In Scanalyzer, analyzing plants misplaces them as a side effect, and bringing them back to their start position, across a large circle of conveyor belts, may take arbitrarily many steps. See Figure 3 for the other 6 domains.

What causes these undesirably high success rates for SP? The author's best guess is that, in many domains, the chance of randomly finding a state on a local minimum is low. In large-scale experiments measuring statistics on the search space surface under FF's heuristic function (Hoffmann, 2003), it was observed that many sampled states were not local minima themselves, but where contained in "valleys". Within a valley, there is no monotonically decreasing path to a goal state. Such a state may not be a local minimum because, and *only* because, one can descend deeper into the valley. It seems that SP correctly identifies most valley states to not be local minima, thus counting as "good" many states that actually are located in difficult regions of the search space. This is a weakness not of SP, but of success rate as a search space feature.[17] Why does this weakness not manifest itself as much in analysis (III)? Because that analysis is "more picky" – it takes as "good" only states that qualify for particular special cases. These tend to not occur as often in the difficult domains.

Of course, it is easy to construct examples turning the discussed "strength" into a real weakness of TorchLight's analysis quality. This just does not seem to happen a lot in the present benchmarks. Now, having said that, the present benchmarks aren't well suited to bring out the theoretical advantage of analysis (III) either. The analysis offers unlimited lookahead depth at low-order polynomial cost. However, even with $R = 1000$, in 23 of the 37 domains the highest exit distance bound returned is 0, i.e., every exit path identified consists of a single operator. These cases could be handled with a much simpler variant of analysis (III), looking only at operators $o_0$ that are directly applicable in $s$, and thus removing the entire machinery pertaining to $SG$ predecessors of $x_0$. Still, that machinery does matter in cases that are quite natural. The highest exit distance bound returned is 10 in Grid and 7 in Transport. More generally, in any transportation domain with a non-trivial road-map, it is easy to construct relevant situations. For example, say the road map in Transport forms $N$ "cities", each with diameter $D$ and at least one vehicle, distances between cities being large relative to $D$. Then, in a typical state, around $N$ vehicle moves will be considered helpful by FF: at least 1 per city since local vehicles will be preferred by the relaxed plan. All successor states will have identical $h^+$ until a package can be loaded/unloaded. The typical number of steps required to do so will grow with $D$. If, for example, the vehicle is in the "outskirts" and the packages are in the "city center", then around $D/2$ steps are required, and finding an exit takes runtime around $N^{D/2}$. Then small values of $N$ and $D$ already render search probing either devoid of information (if the runtime cut-off is too small), or computationally infeasible (recall that the probing should be a "quick" pre-process to the actual planning). By contrast, analysis (III) easily delivers the correct success rate 100%.

## 8.6 Predicting Planner Performance

As a direct measure of the "predictive quality" of success rates, we conducted preliminary experiments examining the behavior of primitive classifiers, and of runtime distributions for large vs. small success rates. We consider first the classifiers. They predict, given a planning task, whether EHC/FF/LAMA will succeed in solving the task, within the given time and memory limits. The classifiers answer "yes" iff the success rate is $\geq$ a threshold $T$ in $0, 10, \ldots, 100$. Obviously, to do this, we need $R > 1$. We consider in what follows only $R = 10$ and $R = 100$ because, as shown above, $R = 1000$ can be costly.

---

17. Note that we cannot use "valley rate" instead, in a cheap domain analysis, since determining whether or not $s$ lies on a valley implies finding a plan for $s$ and thus solving the task as a side effect.

For EHC, both TorchLight analysis (III) and SP deliver fairly good-quality predictions, considering that no actual machine learning is involved. The prediction quality of Torch-Light is just as good as – sometimes slightly better than – that of search. Whether we use $R = 10$ or $R = 100$ does not make a big difference. EHC solves 60.69% of the instances, so that is the rate of correct predictions for a trivial baseline classifier always answering "yes". For $R = 10$, the best rate of correct predictions is 71.90% for TorchLight (with $T = 80$) and 70.17% for SP (with $T = 90$). For $R = 100$, these numbers are 71.76% ($T = 60$) and 71.16% ($T = 100$). Dead-end rate is a very bad predictor. Its best prediction is for the baseline classifier $T = 0$, and the second best classifier ($T = 100$) is only 36.79% correct.

Interestingly, there are major differences between the different sets of domains. On the domains previously analyzed by hand (Hoffmann, 2005; as in Figure 1 but without Elevators and Transport), the best prediction is 75.75% correct for TorchLight with $T = 70$, and 74.07% correct for SP with $T = 100$, vs. a baseline of 63.81%. On the IPC 2006 domains, these numbers are 57.98% and 61.34% vs. baseline 55.46%, and $T = 10$ in both cases, i.e., the best classifier is very close to the baseline. IPC 2008, on the other hand, appears to be exceptionally good-natured, the numbers being 79.52% ($T = 60$) and 82.38% ($T = 80$) vs. baseline 51.90%. It is not clear to us what causes these phenomena.[18]

In summary, the quality of prediction is always clearly above the baseline, around 10% when looking at all domains, and even up to 30% when looking at the IPC 2008 domains only. For comparison, using state-of-the-art classification techniques but only simple features, Roberts and Howe (2009) get 69.47% correctness vs. baseline 74% (for saying "no"), on unseen testing domains for FF. Having said that, if setting $T$ in the above is considered to be the "learning", then the above does not actually distinguish between learning data and testing data. Roberts and Howe's unseen testing domains are those of IPC 2006 (in a different setting than ours including also all ADL test suites). If we set $T$ on only the domains from before 2006 (Figure 1 without Elevators and Transport), then we get the best prediction at $T = 70$ for TorchLight and $T = 100$ for SP. With this setting of $T$, the prediction correctness on our IPC 2006 suite is 29.41% respectively 51.26% only, vs. the baseline 55.46%. On the other hand, this seems to pertain only to IPC 2006 specifically. For IPC 2008, $T = 70$ respectively $T = 100$ are good settings, giving 76.67% respectively 76.19% correctness vs. the baseline 51.90%.

Importantly, Roberts and Howe are not predicting the performance of EHC but that of FF, which is a more complex algorithm. For FF and LAMA, the prediction quality of both TorchLight and SP is rather bleak, using the described primitive classifiers. In all cases, the best prediction correctness is obtained when always answering "yes". The best that can be said is that success rate still predicts much better than dead-end rate. To give some example data, with $R = 10$ across all domains for FF, the baseline is 85.09% correct. With $T = 10$, this goes down to 77.50% for TorchLight, 79.31% for SP, and 34.57% for dead-end rate. For LAMA, the baseline is 90.26% correct, and with $T = 10$ this goes down to 81.81% for TorchLight, 83.97% for SP, and 29.91% for dead-end rate. For both FF and LAMA, with growing $T$ the prediction quality decreases monotonically in all cases.

---

18. The bad prediction quality in IPC 2006 domains might be related to the fact that these are fully grounded, potentially impeding the ability of Fast Downward's translator to find useful finite-domain variables.

Why is prediction quality so much worse for FF than for EHC, which after all is the main building block of FF? Whereas EHC typically fails on tasks whose $h^+$ topology is not favorable, FF's and LAMA's complete search algorithms are able to solve many of these cases, too. For example, with TorchLight success rates and $R = 10$, EHC solves only 34.07% of the tasks with success rate 0, and solves less than 50% up to success rate 70%. By contrast, FF and LAMA solve 74.18% respectively 76.92% of the tasks with success rate 0, and solve at least 70% for all success rates.

Despite this, success rates are far from devoid of information for FF and LAMA. Setting the threshold $T$ in $10, \ldots, 100$, we look at the distribution of planner runtime in the instance subset (A) where success rate is $< T$, vs. instance subset (B) where success rate is $\geq T$. Taking the null hypothesis to be that the means of the two runtime distributions are the same, we run the Student's T-test for unequal sample sizes to determine the confidence with which the null hypothesis can be rejected. That is, we determine the confidence with which distribution (B) has a lower mean than distribution (A). Using TorchLight's success rate on FF runtimes, with both $R = 10$ and $R = 100$, and in all 10 settings of $T$, we get a confidence of at least 99.9%. The difference between the means in our data, i.e., the mean runtime of (A) minus the mean runtime of (B), tends to grow over $T$. It peaks at 336 respectively 361 seconds for $R = 10$ respectively $R = 100$; the average difference over all values of $T$ is 239 respectively 240. Likewise, for LAMA runtimes all settings of $T$ and $R$ yield a confidence of 99.9%, with average differences 242 respectively 235. The results for SP are comparable for LAMA. They are slightly worse for FF, though. With $R = 10$ the confidence is 99.9% only for $T = 10, 20$; the confidence is 95% for all other values of $T$. The difference peaks at 241 seconds (vs. 336 for TorchLight), with an average of 150 seconds (vs. 239). With $R = 100$, thresholds $T = 30, 40, 50, 100$ yield 99.9% confidence, the average difference being 160.

Again perhaps a little surprisingly, for the simpler planner EHC the runtime distributions behave very differently. For TorchLight success rates, we do get several cases with confidence $< 95\%$, and average differences of around 80 seconds. For SP, in most cases we get a 99.9% confidence that the mean of (B) is *larger* than that of (A). Again, the reason is simple. On many tasks with unfavorable $h^+$ topology, enforced hill-climbing quickly exhausts the space of states reachable by FF's helpful actions. EHC then gives up on solving the task, although it has consumed only little runtime – a peculiar behavior that one would certainly not expect from a planner trying to be competitive.

Summing up, success rates as a planning task feature provide a very good coverage predictor for EHC even without any significant learning. For FF and LAMA, things are not that easy, however the consideration of runtime distributions clearly shows that the feature is highly informative. Exploiting this informativeness for predicting planner performance presumably requires combination with other features, and actual machine learning techniques, along the lines of Roberts and Howe (2009). This is a topic for future research.

## 8.7 Diagnosis

Let us finally consider TorchLight's diagnosis facility. The idea behind this facility is to summarize the reasons for analysis failure. Testing sufficient criteria for the absence of local minima, such diagnosis is not guaranteed to identify domain features causing their presence. Still, at least for analysis using Theorem 2, the diagnosis can be quite accurate.

The current diagnosis facility is merely a first-shot implementation based on reporting all pairs (operator $o_0$, variable $x$) that caused an $oDG^+$ for $o_0$ to not be successful. That is, we report the pair $(o_0, x)$ if $o_0$ has an effect on $x$, and a context fact $(x, c)$ of the transition $t_0$ taken by $o_0$ is contained in $R_1^+ \cap C_0 \cap F_0$, and is not recoverable by a sub-sequence of $P_{>0}^+(s)$. In brief, we record $(o_0, x)$ if $o_0$ has a harmful effect on $x$. We perform a test whether the "main" effect of $o_0$, i.e., that on $x_0$, is invertible; in this case we do not record $x_0$ since the problem appear to be the side effects. To avoid redundancies in the reporting, we record not the grounded operator $o_0$ but only the name of the action schema ("load" instead of "load(package1 truck7)"). Similarly, as an option we record not $x$ but the name of the predicate underlying the fact $(x, c)$. In that configuration, the diagnosis comes in the form of "action-name, predicate-name", which has a direct match with the high-level PDDL input files. To have some measure of which parts of the diagnosis are "more important", we associate each pair with a count of occurrences, and weigh the pairs by frequency.

In Zenotravel, the diagnosis output always has the form "fly, fuel-level" and "zoom, fuel-level", indicating correctly that it's the fuel consumption which is causing the local minima. In Mprime and Mystery, the cause of local minima is the same, however the diagnosis is not as reliable because of the specific structure of the domain, associating fuel with locations instead of vehicles. This sometimes causes the diagnosis to conclude that it is the effect changing locations which is causing the trouble. Concretely, with $R = 1000$ in Mystery, fuel consumption is the top-weighted diagnosis in 17 out of the 28 tasks; in Mprime, this happens in 30 out of the 35 tasks. In Satellite and Rovers, the diagnosis always takes the form "switch-on, calibrated" respectively "take-image, calibrated", thus reporting the problem to be that switching on an instrument, respectively taking an image, deletes calibration. This is precisely the only reason why local minima exist here.[19] In Tyreworld, most often the diagnosis reports the problem to be that jacking up a hub results in no longer having the jack (which is needed elsewhere, too). While this does not actually cause local minima (there are none), it indeed appears to be a crucial aspect of the domain. Similarly, in Grid the most frequent diagnosis is that picking up a key results in the arm no longer being empty – again, not actually a cause of local minima, but a critical resource in the domain. In Blocksworld-Arm, the dominant diagnoses are that a block is no longer clear if we stack something on top of it, and that the hand is no longer empty when picking up a block. Similarly, in Freecell, the dominant diagnoses are "send-to-free, cellspace" and "send-to-new-col, colspace".

One could make the above list much longer, however it seems clear already that this diagnosis facility, although as yet primitive, has the potential to identify interesting aspects of the domain. Note that we are making use of only one of the information sources in TorchLight. There are many other things to be recorded, pertaining to other reasons for analysis failure, like support graph cycles etc, and also to reasons for analysis success, like successful $gDG$s and $x_0, o_0$ pairs yielding successful $oDG^+$s. It appears promising to try to improve diagnosis by combining some of these information sources. A combination with other domain analysis techniques, like landmarks or invariants extraction, could also be useful. This is a direction for future work.[20]

---

19. Since analysis failure is rare in these two domains, often diagnosis does not give any output at all. With $R = 1000$, the output is non-empty in 10 instances of Satellite and in 8 instances of Rovers. For $R = 100$ this reduces to 4 instances in Satellite, and not a single one in Rovers.

## 9. Related Work

There is no prior work – other than the aforementioned one of the author (Hoffmann, 2005) – trying to automatically infer topological properties of a heuristic function. Thus our work does not relate strongly to other domain analysis techniques. The closest relation is to other techniques relying on causal graphs. In what follows we discuss this in some detail, along with some other connections arising in this context.

If local analysis succeeds, then we can construct a path to the exit identified. In this, our work relates to work on macro-actions (e.g., Botea, Müller, & Schaeffer, 2004; Vidal, 2004). Its distinguishing feature is that this macro-action is (would be) constructed in a very targeted and analytical way, even giving a guarantee, in the conservative case, to make progress towards the goal. The machinery behind the analysis is based on causal graphs, and shares some similarities with known causal-graph based execution path generation methods (e.g., Jonsson & Bäckström, 1995; Williams & Nayak, 1997; Brafman & Domshlak, 2003). The distinguishing feature here is that we focus on $h^+$ and individual states rather than the whole task. This allows us to consider small fragments of otherwise arbitrarily complex planning tasks – we look at $oDG^+$ instead of $SG$. Note that this ability is quite powerful as far as applicability goes. As we have seen in Section 8, the success rate of (local) approximate analysis – and therewith the fraction of states for which we would be able to generate a macro-action – is non-zero in almost all benchmark domains. Of course, this broad applicability comes with a prize. While traditional causal graph methods guarantee to reach the goal, in the worst case the macro-actions may only lead into $h^+$ local minima. Still, it may be interesting to look into whether other, traditional, causal-graph based methods can be "localized" in this (or a similar) manner as well.

Global analysis, where we focus on the whole planning task and thus the whole causal graph, is even more closely related to research on causal graphs based tractability analysis. The major difference between tractability analysis and $h^+$ topology analysis, in principle, is that tractability and absence of local minima are orthogonal properties – in general, neither one implies the other. Now, as we pointed out at the end of Section 6, our global analysis does imply tractability (of plan existence). Vice versa, do the restrictions made in known tractable classes imply the absence of local minima? In many cases, we can answer this question with a definite "no"; some interesting questions are open; in a single case – corresponding to our basic result – the answer is "yes".

Example 3 in Appendix A.4 shows that one can construct a local minimum with just 2 variables of domain size 3, 1-arc $SG$, unary operators, and strongly connected DTGs with a single non-invertible transition. This example (and various scaling extensions not breaking the respective conditions) falls into a variety of known tractable classes. The example is in the tractable class $\mathbb{F}_n^\vee$ identified by Domshlak and Dinitz (2001), because every transition of the dependent variable depends on the other variable. The example is in Helmert's (2004, 2006) SAS$^+$-1 class with strongly connected DTGs. The example is "solved", i.e., reduced to the empty task, by Haslum's (2007) simplification techniques (also, these techniques solve tasks from the Satellite domain, which do contain local minima). The example has a fork and inverted fork causal graph, with bounded domain size and 1-dependent actions only (actions with at most 1 prevail condition), thus it qualifies for the tractable classes identified by Katz and Domshlak (2008b). The example's causal graph is a chain, and thus in particular a polytree with bounded indegree, corresponding to the tractable class identified by Brafman and Domshlak (2003) except that, there, variables are restricted to be binary (domain size 2). It is an open question whether plan existence with chain causal graphs and domain size 3 is tractable; the strongest known result is that it is **NP**-hard for domain size 5 (Giménez & Jonsson, 2009b).[21] Similarly, the example fits the prerequisites stated by Katz and Domshlak (2008a) except that these are for binary variables only; it is an open question whether local minima exist in the tractable classes identified there. Finally, the example, and a suitable scaling extension, obviously qualifies for two theorems stated by Chen and Gimenez (2010). Their Theorem 3.1 (more precisely, the first part of that theorem) requires only a constant bound on the size of the connected components in the undirected graph induced by the causal graph. The first part of their Theorem 4.1 requires a constant bound on the size of the strongly connected components in the causal graph, and pertains to a notion of "reversible" tasks requiring that we can always go back to the initial state.

---

20. In particular, Fast Downward's translator is not always perfect in detecting the finite-domain variables underlying benchmarks. For example, in Satellite it often does not detect that electricity is available in exactly one of the instruments mounted on a satellite. This can lead to pointless diagnosis output, which for now is handled using a simple notion of predicates "exchanged" by every operator. For doing things like this in a more principled manner, further invariants analysis would be useful.

Next, consider the line of works restricting not the causal graph but the DTGs of the task (Bäckström & Klein, 1991; Bäckström & Nebel, 1995; Jonsson & Bäckström, 1998). The simplest class identified here, contained in all other classes, is SAS$^+$-PUBS where each fact is achieved by at most one operator ("post-unique", "P"), all operators are unary ("U"), all variables are binary ("B"), and all variables have at most one value required in the condition of a transition on any other variable ("single-valued", "S"). Now, Example 2 in Appendix A.4 shows a local minimum in an example that has the U and S properties. The example has two variables, $x$ and $y$, and the local minimum arises because a cyclic dependency prevents $y$ from attaining its goal value $d_n$ via the shortest path as taken by an optimal relaxed plan. If we remove all but two values from the domain of $y$, and remove the alternative way of reaching $d_n$,[22] then the example still contains a local minimum and also has the P and B properties. We remark that the modified example is unsolvable. It remains an open question whether solvable SAS$^+$-PUBS tasks with local minima exist; more generally, this question is open even for the larger SAS$^+$-PUS class, and for the (yet larger) SAS$^+$-IAO class identified by Jonsson and Bäckström (1998).

Another open question is whether the "3S" class of Jonsson and Bäckström (1995) contains local minima. The class works on binary variables only; it requires unary operators and acyclic causal graphs, however it allows facts to be "splitting" instead of reversible. If $p$ is splitting then, intuitively, the task can be decomposed into three independent sub-tasks with respect to $p$; it is an open question whether local minima can be constructed while satisfying this property. Disallowing the "splitting" option in 3S, we obtain the single "positive" case, where a known tractable class does not contain any local minima. This class corresponds to our basic result – acyclic causal graphs and invertible transitions – except that the variables are restricted to be binary. Williams and Nayak (1997) mention restrictions (but do not make formal claims regarding tractability) corresponding exactly to our basic result except that they allow irreversible "repair" actions. The latter actions are defined relative to a specialized formal framework for control systems, but in spirit they are similar to what we term "transitions with self-irrelevant deletes" herein.

---

21. Although, of course, it is clear that, if the DTGs are strongly connected as in our case, then deciding plan existence is tractable no matter what the domain size is.
22. This modification is given in detail below the example in Appendix A.4.

Finally, it is easy to see that, of Bylander's (1994) three tractability criteria, those two allowing several effects do not imply the absence of local minima. For his third criterion, restricting action effects to a single literal and preconditions to positive literals (but allowing negative goals), we leave it as an open question whether or not local minima exist. We remark that this criterion does not apply in any benchmark we are aware of.

To close this section, while we certainly do not wish to claim the identification of tractable classes to be a contribution of our work, we note that the scope of Theorem 4 – which is a tractable class, cf. the above – is not covered by the known tractable classes.[23] The tractable cases identified by Bylander (1994) obviously do not cover any of Logistics, Miconic-STRIPS, Movie, and Simple-TSP. Many causal graph based tractability results require unary operators (Jonsson & Bäckström, 1995; Domshlak & Dinitz, 2001; Brafman & Domshlak, 2003; Helmert, 2004, 2006; Katz & Domshlak, 2008a, 2008b; Jonsson, 2009; Giménez & Jonsson, 2008, 2009a), which does not cover Miconic-STRIPS, Movie, and Simple-TSP. In the work of Chen and Gimenez (2010), Theorem 4.1 requires reversibility which is not given in either of Movie, Miconic-STRIPS, or Simple-TSP, and Theorem 3.1 requires a constant bound on the size of the connected components in the undirected graph induced by the causal graph, which is given in none of Logistics, Miconic-STRIPS, and Simple-TSP. Other known tractability results make very different restrictions on the DTGs (Bäckström & Klein, 1991; Bäckström & Nebel, 1995; Jonsson & Bäckström, 1998). Even the most general tractable class identified there, SAS$^+$-IAO, covers none of Miconic-STRIPS, Logistics, and Simple-TSP (because vehicle variables are not "acyclic with respect to requestable values"), and neither does it cover Movie (because rewinding a movie is neither unary nor "irreplaceable": it has a side effect un-setting the counter, while not breaking the DTG of the counter into two disjoint components).

As far as coverage of the benchmarks goes, the strongest competitor of Theorem 4 are Haslum's (2007) simplification techniques. These iteratively remove variables where all paths relevant for attaining required conditions are "free", i.e., can be traversed using transitions that have neither conditions nor side effects. Haslum's Theorem 1 states that such removal can be done without jeopardizing solution existence, i.e., a plan for the original task can be reconstructed easily from a plan for the simplified task. In particular, if the task is "solved" – simplified completely, to the empty task – then a plan can be constructed in polynomial time. Haslum combines this basic technique with a number of domain reformulation techniques, e.g., replacing action sequences by macros under certain conditions. The choice which combination of such techniques to apply is not fully automated, and parts of these techniques are not fully described, making a comparison to Theorem 4 difficult. Haslum reports his techniques to solve tasks from Logistics, Miconic-STRIPS, and Movie, plus Gripper and Satellite. Haslum does not experiment with Simple-TSP. His Theorem 1, in its stated form, does not solve Simple-TSP, because there the transitions of the root variable have side effects (with irrelevant deletes). Extending the theorem to cover such irrelevant deletes should be straightforward. A more subtle weakness of Haslum's Theorem 1 relative to our Theorem 4 pertains to reaching required values from externally caused values. Haslum requires these moves to be free, whereas, in the definition of recoverable side effect deletes, Theorem 4 allows the recovering operators to affect several variables and to take their precondition from the prevails and effects of $o_0$.

---

23. This is not true of our basic result, which as just explained is essentially covered by the works of Jonsson and Bäckström (1995) and Williams and Nayak (1997). Formally, its prerequisites imply those of (the first part of) Theorem 4.1 in the work of Chen and Gimenez (2010), namely, the postulated bound is 1.

## 10. Conclusion

We identified a connection between causal graphs and $h^+$, and devised a tool allowing to analyze search space topology without actually running any search. The tool is not yet an "automatic Hoffmann", but its analysis quality is impressive even when compared to unlimited search probing.

At a very generic level, a conclusion of this work is that, sometimes, it is *possible* to automatically infer topological properties of a heuristic function. An interesting question for future work is whether this can also be done for heuristics other than $h^+$ (cf. also the comments regarding causal graph research below). Methodologically, it is noteworthy that the analysis is based on syntactic restrictions on the problem description, which has traditionally been used to identify tractable fragments (of planning and other computationally hard problems). The present work showcases that very similar techniques can apply to the analysis of the search spaces of general problem solvers.

A main open question is whether global analysis can more tightly approximate the scope of Theorem 2. As indicated, a good starting point appears to be trying to include, in a $gDG$ for operator $o_0$, only variable dependencies induced by operators $o$ that may actually precede $o_0$ in an optimal relaxed plan. An approach automatically recognizing such operators could possibly be developed along the lines of Hoffmann and Nebel (2001b), or using a simplified version of the aforementioned "fact generation tree" analysis technique (Hoffmann, 2005). Additionally, it would be great to recognize situations in which harmful side effects of $o_0$ – like making the hand non-empty if we pick up a ball in Gripper – will necessarily be recovered inside the relaxed plan. Possibly, such analysis could be based on a variant of action landmarks (Hoffmann, Porteous, & Sebastia, 2004; Karpas & Domshlak, 2009).

Another interesting line of research is to start from results given for individual states $s$ by local analysis, then extract the reasons for success on $s$, and generalize those reasons to determine a generic property under which success is guaranteed. Taken to the extreme, it might be possible to automatically identify domain sub-classes, i.e., particular combinations of initial state and goal state, in which the absence of local minima is proved.

This work highlights two new aspects of causal graph research. First, it shows that, in certain situations, one can "localize" the causal graph analysis, and consider only the causal graph fragment relevant for solving a particular state. Second, one can use causal graphs for constructing paths not to the global goal, but to a state where the value of a heuristic $h$ is decreased. The former enables the analysis to succeed in tasks whose causal graphs are otherwise arbitrarily complex, and thus has the potential to greatly broaden the scope of applicability. The latter is not necessarily limited to only $h^+$ – as a simple example, it is obvious that similar constructions can be made for the trivial heuristic counting the number of unsatisfied goals – and thus opens up a completely new avenue of causal graph research.

Another possibility is planner performance prediction, along the lines of Roberts and Howe (2009). Our experimental results indicate that TorchLight's problem features, and also those of search probing, are highly informative. This has the potential to significantly improve the results of Roberts and Howe for unseen domains – they currently use only very simple features, like counts of predicates and action schemes, that hardly capture a domain-independent structure relevant to planner performance. Like limited search probing ($SP^{1s}$), TorchLight generates its features without jeopardizing runtime, thus enabling automatic planner configuration. Unlike for search probing, this may even work on-line during search: a single relaxed plan can already deliver interesting information. For example, one might make the search more or less greedy – choosing a different search strategy, switching helpful actions on or off, etc. – depending on the outcome of checking Theorem 2.

As mentioned in Section 9, a direction worth trying is to use local analysis for generating macro-actions. In domains with high success rate, it seems likely that the macro-actions would lead to the goal with no search at all. It is a priori not clear, though, whether such an approach would significantly strengthen, at least in the present benchmarks, existing techniques for executing (parts of) a relaxed plan (e.g., Vidal, 2004).

One could use TorchLight's diagnosis facility as the basis of an abstraction technique for deriving search guidance, much as is currently done with other relaxation/abstraction techniques. The diagnosis can pin-point which operator effects are causing problems for search. If we remove enough harmful effects to end up with a task to which Theorem 4 applies, then the abstracted problem is tractable. For example, in transportation domains, this process could abstract away the fuel consumption. If we do not abstract that much, then the information provided may still outweigh the effort for abstract planning, i.e., for using an actual planner inside the heuristic function. For example, in Grid the abstract task could be a problem variant allowing to carry several keys at once. One could also focus the construction of different heuristics – not based on ignoring deletes – on the harmful effects.

Finally, an interesting research line is domain reformulation. As is well known, the domain formulation can make a huge difference for planner performance. However, it is very difficult to choose a "good" formulation, for a given planner. This is a black art even if the reformulation is done by the developer of the planner in question. The lack of guidance is one of the main open problems identified by Haslum (2007) for his automatic reformulation approach. The most frequent question the author has been asked by non-expert users is how to model a domain so that FF can handle it more easily.

TorchLight's diagnosis facility, pin-pointing problematic effects, might be instrumental for addressing these difficulties. For the case where the reformulation is done by a computer, one possibility to use the analysis outcome could be to produce macro-actions "hiding" within them the operators having harmful effects. Another possibility could be to pre-compose variable subsets touched by the harmful effects.

For the case where the reformulation is done by a human user, the sky is the limit. To name just one example, the local minima in Satellite could be removed by allowing to switch on an instrument only when pointing in a direction where that instrument can be calibrated. More generally, note that end-user PDDL modeling – writing of PDDL by a non-expert user wanting to solve her problem using off-the-shelf planners – is quite different from the PDDL modeling that planning experts do when developing benchmarks. For example, if an expert models a transportation benchmark with fuel consumption, then it may seem quite pointless for TorchLight to determine that fuel consumption will hurt planner performance. Indeed this may be the reason why the fuel consumption was included in the first place. By contrast, for an end-user (a) this information may come as a surprise, and (b) the user may actually choose to omit fuel consumption because this may yield a better point in the trade-off between planner performance and plan usability. Generally speaking, such an approach could give the user guidance in designing a natural hierarchy of increasingly detailed – and increasingly problematic – domain formulations. This could help making planning technology more accessible, and thus contribute to a challenge that should be taken much more seriously by the planning community.

## Acknowledgments

I would like to thank the anonymous reviewers of both, the article at hand and the ICAPS 2011 short version, for their constructive comments. In particular, one of the reviewers proved the completeness results in Theorem 1, and another reviewer suggested the future research line trying to generalize the reasons for success in local analysis.

I thank Carmel Domshlak for discussions, feedback on early stages of this work – contributing in particular the "$d$-abstracted task" construction in the proof of Lemma 3 – and an executive summary of the status quo of causal graph research.

A very special thanks goes to Carlos Areces and Luciana Benotti, for inspiring this work in the first place. I had long ago given up on this problem. It was Carlos' and Luciana's insistence that finally made me see the connection to causal graphs – while trying to convince them that an analysis like this is impossible.

## Appendix A. Technical Details and Proofs

We give the full proofs and, where needed, fill in some technical definitions. We first prove our complexity result (Appendix A.1, Theorem 1), then the result pertaining to the analysis of optimal relaxed plans (Appendix A.2, Theorem 2), then the result pertaining to conservative approximations (Appendix A.3, Theorems 3 and 4). We construct a number of examples relevant to both kinds of analysis (Appendix A.4), before giving the proofs of domain-specific performance guarantees (Appendix A.5, Propositions 1 and 2).

### A.1 Computational Complexity

**Theorem 1.** *It is* **PSPACE**-*complete to decide whether or not the state space of a given planning task contains a local minimum, and given an integer $K$ it is* **PSPACE**-*complete to decide whether or not for all states $s$ we have $ed(s) \leq K$. Further, it is* **PSPACE**-*complete to decide whether or not a given state $s$ is a local minimum, and given an integer $K$ it is* **PSPACE**-*complete to decide whether or not $ed(s) \leq K$.*

*Proof.* Throughout the proof, since **PSPACE** is closed under complementation, we do not distinguish the mentioned **PSPACE**-complete decision problems from their complements.

The membership results are all easy to prove. Note first that, given a state $s$, we can compute $h^+(s)$ within polynomial space: generate a potentially non-optimal relaxed plan, of length $n$, with the known methods; then iteratively decrement $n$ and test for each value whether a relaxed plan of that length still exists; stop when that test answers "no". The test for bounded relaxed plan existence is in **NP** and thus in **PSPACE**. From here, we can prove the membership results by simple modifications of the guess-and-check argument showing that PLANSAT, the problem of deciding whether a given planning task is solvable, is in **NPSPACE** and hence in **PSPACE** (Bylander, 1994). That argument works by starting in the initial state, guessing actions, and terminating successfully if a goal state is reached. Unsuccessful termination occurs if the guessed path is longer than the trivial upper bound $B := \Pi_{x \in X} |D_x|$ on the number of different states. To be able to check this condition in polynomial space, the path length is maintained in a binary counter.

To decide whether a given state $s$ is (not) a local minimum, we run this guess-and-check algorithm from $s$, modified to: compute $h^+$ for each encountered state; to terminate unsuccessfully if the bound $B$ is exceeded or if $h^+$ increases after an operator application; and to terminate successfully if $h^+$ decreases after an operator application. To decide whether $ed(s) \leq K$, we use the same algorithm except that the bound $B$ is replaced by the bound $K$, increases of $h^+$ are permitted, and success occurs if $h^+$ decreases from $h^+(s)$ to $h^+(s)-1$. To decide whether the state space of an entire planning task contains local minima, or whether all states $s$ in the state space have $ed(s) \leq K$, we simply run Bylander's guess-and-check algorithm as a way of enumerating all reachable states, then for each individual state $s$ we run the modified guess-and-check algorithms just described. Clearly, all these algorithms run in non-deterministic polynomial space, which shows this part of the claim.

We now show the **PSPACE**-hardness results. We first consider the problem of deciding whether or not a given state $s$ is a local minimum. The proof works by reducing PLANSAT, which is known to be **PSPACE**-hard for propositional STRIPS (Bylander, 1994), from which it trivially follows that PLANSAT is **PSPACE**-hard also for the finite-domain variable planning tasks we use herein.

Let $(X, s_I, s_G, O)$ be the planning task whose solvability we wish to decide. We design a modified task $(X', s'_I, s'_G, O')$ by starting with $(X, s_I, s_G, O)$ and making the following modifications:

- Add a new variable $ChooseTask$ to $X'$, with domain $\{nil, org, alt\}$, $s'_I(ChooseTask) = nil$, and $s'_G(ChooseTask)$ undefined.

  The role of this variable will be to give the planner a choice whether to solve the original task $(X, s_I, s_G, O)$, or whether to solve an alternative task custom-designed for this proof.

- Add a new variable $DistAlt$ to $X'$, with domain $\{0, 1\}$, $s'_I(DistAlt) = 1$, and $s'_G(DistAlt) = 1$.

  This variable simply serves to control the length of the solution of the alternative task. That solution length will be 1 plus the number of steps needed to bring $DistAlt$ from

195

value 0 to its goal value. (Here, only 1 step will be needed for doing so; later on in this proof, we will increase this distance.)

- Add two new operators $o_{Org} = (\{(ChooseTask, nil)\}, \{(ChooseTask, org)\})$ and $o_{Alt} = (\{(ChooseTask, nil)\}, \{(ChooseTask, alt), (DistAlt, 0)\})$.

  This implements the choice of planning task. Note that, if we choose the alternative task, then $DistAlt$ is set to 0, thus forcing the solution to bridge this distance. By contrast, for the original task, this variable keeps residing in its goal value as was already assigned by $s'_I(DistAlt)$.

- Add a new operator $o_{DistAlt} = (\{(ChooseTask, alt), (DistAlt, 0)\}, \{(DistAlt, 1)\})$.

  This allows to bridge the distance intended for the solution of the alternative task.

- Add a new operator $o_{s_G Alt} = (\{(ChooseTask, alt), (DistAlt, 1)\}, s_G)$.

  This allows us to accomplish the original goal, as the final step in solving the alternative task.

- Add $(ChooseTask, org)$ as a new precondition into all original operators, i.e., those taken from $O$.

  This forces the planner to choose the original task, for executing any of its operators.

- Add a new variable $StillAlive$ to $X$, with domain $\{yes, no\}$, $s'_I(StillAlive) = yes$, and $s_G(StillAlive) = yes$. Add a new operator $o_{s_G Dead} = (\emptyset, s_G \cup \{(StillAlive, no)\})$.

  The $o_{s_G Dead}$ operator allows us to accomplish the original goal in a single step, no matter which task we have chosen to solve, and also in the new initial state $s'_I$ already. However, the operator also sets the new variable $StillAlive$ to value $no$, whereas the goal value of that variable is $yes$. That value cannot be re-achieved, and thus the operator leads into a dead-end. Its function in the proof is to flatten the value of $h^+$ in the original task, and in $s'_I$, to be constantly 1 unless we are in a goal state. This extreme flattening does not happen in the alternative task because, there, the distance variable $DistAlt$ also needs to be handled.

In summary, $(X', s'_I, s'_G, O')$ is designed by setting:

- $X' := X \cup \{ChooseTask, DistAlt, StillAlive\}$

- $s'_I := s_I \cup \{(ChooseTask, nil), (DistAlt, 1), (StillAlive, yes)\}$

- $s'_G := s_G \cup \{(DistAlt, 1), (StillAlive, yes)\}$

- $O' := \{(pre \cup \{(ChooseTask, org)\}, eff) \mid (pre, eff) \in O\} \cup \{o_{Org}, o_{Alt}, o_{DistAlt}, o_{s_G Alt}, o_{s_G Dead}\}$

Now consider the new initial state $s'_I$. It has exactly three successor states: $s_{Dead}$ produced by $o_{s_G Dead}$, $s_{Org}$ produced by $o_{Org}$, and $s_{Alt}$ produced by $o_{Alt}$. We have $h^+(s_{Dead}) = \infty$ because $s_{Dead}(StillAlive) = no$. We have $h^+(s'_I) = h^+(s_{Org}) = 1$ due to the relaxed plan $\langle o_{s_G Dead} \rangle$. Finally, we have $h^+(s_{Alt}) = 2$ because $o_{Alt}$ sets the $DistAlt$ variable to 0 whereas its goal is 1. Thus a shortest relaxed plan for $s_{Alt}$ is $\langle o_{DistAlt}, o_{s_G Alt} \rangle$.

From this, it clearly follows that $s'_I$ is not a local minimum iff $s_{Org}$ has a monotone path to a state $s$ with $h^+(s) < h^+(s_{Org})$. Since $h^+(s_{Org}) = 1$, the latter is equivalent to the existence of a monotone path from $s_{Org}$ to a goal state, i.e., a path to a goal state on which $h^+$ is constantly 1. Since, for all states reachable from $s_{Org}$, the single-step sequence $\langle o_{s_G Dead} \rangle$ is a relaxed plan, this is equivalent to the existence of a path from $s_{Org}$ to a goal state. Clearly, the latter is equivalent to solvability of the original task $(X, s_I, s_G, O)$. Thus $s'_I$ is not a local minimum iff $(X, s_I, s_G, O)$ is solvable, which shows this part of the claim.

We next prove **PSPACE**-hardness of deciding whether or not a given planning task contains a local minimum. This follows easily from the above. Observe that the alternative task does not contain any local minima. As described, we have $h^+(s_{Alt}) = 2$. If we apply $o_{DistAlt}$ to $s_{Alt}$, then we obtain a state $s_{AltDist}$ where $h^+(s_{AltDist}) = 1$ because of the relaxed plan $\langle o_{s_G Alt} \rangle$. Applying $o_{s_G Alt}$ in $s_{AltDist}$ yields a goal state, and thus both $s_{Alt}$ and $s_{AltDist}$ have better evaluated neighbors. Any other states descending from $s_{Alt}$ must be produced by $o_{s_G Dead}$ and thus have $h^+$ value $\infty$. So, $(X', s'_I, s'_G, O')$ contains a local minimum iff the part of its state space descended from $s_{Org}$ does. Since all those states have $h^+$ value 1 unless they are goal states, cf. the above, the latter is equivalent to unsolvability of $(X, s_I, s_G, O)$ which shows this part of the claim.

Assume now that we are given an integer $K$ and need to decide for an individual state $s$ whether or not $ed(s) \leq K$. We reduce Bounded-PLANSAT, the problem of deciding whether any given planning task is solvable within a given number of steps. Bounded-PLANSAT is known to be **PSPACE**-complete if the bound is given in non-unary representation. We modify the task $(X', s'_I, s'_G, O')$ given above, in a way that increases the solution length of the alternative task to be $K$. We introduce a binary counter using $\lceil \log_2(K-2) \rceil$ new binary variables $Bit_i$ that are all at 0 in $s_I$. We introduce an operator for each bit, allowing to set the bit to 1 if all the lower bits are already 1, and in effect setting all these lower bits back to $O$. Each such operator has the additional precondition $(ChooseTask, alt)$, but has no effect other than modifying the bits. We then modify the operator $o_{DistAlt}$ by adding new preconditions encoding counter position $K-2$. With this construction, clearly $h^+(s_{Alt}) > 1$, and the distance to goal of $s_{Alt}$ is $K$: a plan is to count up to $K-2$, then apply $o_{DistAlt}$, then apply $o_{s_G Alt}$. Thus, the shortest exit path for $s_I$ via $o_{Alt}$ has length $K+1$. But then, with the above, $ed(s_I) \leq K$ iff $(X, s_I, s_G, O)$ has a plan of length at most $K-1$, which concludes this part of the claim.

Finally, say we need to decide whether or not, for all $s \in S$, we have $ed(s) \leq K$. Note first that $s_{Alt}$ and all its successors necessarily have exit distance at most $K$ (the goal can be reached in at most that many steps), and that the exit distance of $s_{Org}$ and all its successors is equal to the length of a shortest plan for the corresponding state in $(X, s_I, s_G, O)$. The latter length may, for some states in $(X, s_I, s_G, O)$, be longer than $K$ even if the shortest plan for $(X, s_I, s_G, O)$ (i.e., for the original initial state) has length $K$. We thus introduce another binary counter, this time counting up to $K-1$, conditioned on $(ChooseTask, org)$, and with a new operator whose precondition demands the new counter to be at $K-1$ and that achieves all goals. Then, clearly, $s_{Org}$ and all its descendants have exit distance at most $K$. Thus the only state that may have exit distance greater than $K$ is $s'_I$ – precisely, we have $ed(s'_I) = K + 1$ iff the new counter is the shortest plan for $s_{Org}$, which obviously is the case iff $(X, s_I, s_G, O)$ has no plan of length at most $K - 1$. This concludes the argument. $\square$

## A.2 Analyzing Optimal Relaxed Plans

We need to fill in some notations. For the sake of self-containedness of this section, we first re-state the definitions given in Section 5:

**Definition 1.** *Let $(X, s_I, s_G, O)$ be a planning task, let $s \in S$ with $0 < h^+(s) < \infty$, let $P^+(s)$ be an optimal relaxed plan for $s$, let $x_0 \in X$, and let $o_0 \in P^+(s)$ be an operator taking a relevant transition of the form $t_0 = (s(x_0), c)$.*

*An optimal rplan dependency graph for $P^+(s)$, $x_0$ and $o_0$, or optimal rplan dependency graph for $P^+(s)$ in brief, is a graph $oDG^+ = (V, A)$ with unique leaf vertex $x_0$, and where $x \in V$ and $(x, x') \in A$ if either: $x' = x_0$, $x \in X_{pre_{o_0}}$, and $pre_{o_0}(x) \neq s(x)$; or $x \neq x' \in V \setminus \{x_0\}$ and there exists $o \in P^+_{<0}(s)$ taking a relevant transition on $x'$ so that $x \in X_{pre_o}$ and $pre_o(x) \neq s(x)$.*

*For $x \in V \setminus \{x_0\}$, by $oDTG^+_x$ we denote the sub-graph of $DTG_x$ that includes only the values true at some point in $P^+_{<0}(s, x)$, the relevant transitions $t$ using an operator in $P^+_{<0}(s, x)$, and at least one relevant inverse of such $t$ where a relevant inverse exists. We refer to the $P^+_{<0}(s, x)$ transitions as original, and to the inverse transitions as induced.*

**Definition 2.** *Let $(X, s_I, s_G, O)$, $s$, $P^+(s)$, $x_0$, $t_0$, and $oDG^+ = (V, A)$ be as in Definition 1. We say that $oDG^+$ is* successful *if all of the following holds:*

*(1) $oDG^+$ is acyclic.*

*(2) We have that either:*

    *(a) the $oDG^+$-relevant deletes of $t_0$ are $P^+_{>0}(s)$-recoverable; or*

    *(b) $s(x_0)$ is not $oDG^+$-relevant, and $t_0$ has replaceable side effect deletes; or*

    *(c) $s(x_0)$ is not $oDG^+$-relevant, and $t_0$ has recoverable side effect deletes.*

*(3) For $x \in V \setminus \{x_0\}$, all $oDTG^+_x$ transitions either have self-irrelevant deletes, or are invertible/induced and have irrelevant side effect deletes and no side effects on $V \setminus \{x_0\}$.*

We next define two general notions that will be helpful to state our proofs.

- The *prevail condition* $prev_o$ of an operator $o \in O$ results from restricting $pre_o$ to the set of variables $X_{pre_o} \setminus X_{eff_o}$.

- Let $x \in X$, let $(c, c')$ be a transition in $DTG_x$, and let $(y, d) \in seff(c, c')$ be a side effect of the transition. The *context of $(y, d)$ in $(c, c')$* is $ctx(c, c', y, d) :=$

$$\begin{cases} (y, pre_{rop(c,c')}(y)) & y \in X_{pre_{rop(c,c')}} \\ \{(y, d') \mid d' \in D_y, d' \neq d\} & y \notin X_{pre_{rop(c,c')}} \end{cases}$$

The *context of $(c, c')$* is the set $ctx(c, c')$ of all partial variable assignments $\psi$ so that, for every $(y, d) \in seff(c, c')$, $y \in X_\psi$ and $(y, \psi(y)) \in ctx(c, c', y, d)$. We identify $ctx(c, c')$ with the set of all facts that occur in any of its assignments.

Note here that the definition of $\mathrm{ctx}(c, c')$ over-writes our previous one from Section 5, but only in the sense that we now also distinguish all possible tuples of context values, rather than just collecting the overall set. We need the more fine-grained definition to precisely formulate Definition 2 condition (2c), i.e., under which conditions a transition has "recoverable side effect deletes". Namely, Definition 2 conditions (2b) and (2c) are formalized as follows:

- A transition $(c, c')$ has *replaceable side effect deletes* iff $\mathrm{ctx}(c, c') \cap s_G = \emptyset$ and, for every $\mathrm{rop}(c, c') \neq o \in O$ where $\mathrm{pre}_o \cap \mathrm{ctx}(c, c') \neq \emptyset$ there exists $o' \in O$ so that $\mathrm{eff}_{o'} = \mathrm{eff}_o$ and $\mathrm{pre}_{o'} \subseteq \mathrm{prev}_{\mathrm{rop}(c, c')} \cup \mathrm{eff}_{\mathrm{rop}(c, c')}$.

- A transition $(c, c')$ has *recoverable side effect deletes* iff the following two conditions hold:

  - Either $(c, c')$ has irrelevant side effect deletes or, for every $\psi \in \mathrm{ctx}(c, c')$, there exists a *recovering operator* $o$ so that $\mathrm{pre}_o \subseteq \mathrm{prev}_{\mathrm{rop}(c, c')} \cup \mathrm{eff}_{\mathrm{rop}(c, c')}$ and $\mathrm{eff}_o \subseteq \psi$, $\mathrm{eff}_o \supseteq \psi \cap (s_G \cup \bigcup_{\mathrm{rop}(c, c') \neq o' \in O} \mathrm{pre}_{o'})$.

  - Every $(y, d) \in \mathrm{seff}(c, c')$ is not in the goal and appears in no operator precondition other than possibly those of the recovering operators.

If $t_0$ has replaceable side effect deletes, then upon its execution we can remove $o_0$ from the relaxed plan because any operator relying on deleted facts can be replaced. If $t_0$ has recoverable side effect deletes, then, due to the first clause of this definition, no matter what the state $s_0$ in which we apply $t_0$ is – no matter which context $\psi$ holds in $s_0$ – we have a recovering operator $o$ that is applicable after $t_0$ and that re-achieves all relevant facts. Due to the second clause, $o$ will not delete any facts relevant elsewhere in the relaxed plan (note here that anything deleted by $o$ must have been a side effect of $t_0$).

Finally, to formally define the notion used in Definition 2 condition (2a) – "the $oDG^+$-relevant deletes of $t_0$ are $P_{>0}^+(s)$-recoverable" – we now assume the surroundings pertaining to Theorem 2, i.e., $(X, s_I, s_G, O)$ is a planning task, $s$ is a state, $P^+(s)$ is an optimal relaxed plan for $s$, $oDG^+ = (V, A)$ is an optimal rplan dependency graph with leaf variable $x_0$ and transition $t_0 = (s(x_0), c)$ with responsible operator $o_0$. We are considering a state $s_0$ where $t_0$ can be executed, reaching a state $s_1$, and we are examining a relaxed plan $P_1^+$ for $s_1$ constructed from $P^+(s)$ by removing $o_0$, and by replacing some operators of $P_{<0}^+(s)$ with operators responsible for induced $oDTG_x^+$ transitions for $x \in V \setminus \{x_0\}$.

- By $C_0 := \{(x_0, s(x_0))\} \cup \mathrm{ctx}(t_0)$ we denote the values potentially deleted by $t_0$.

- By $R_1^+$ we denote the union of $s_G$, the precondition of any $P^+(s)$ operator other than $o_0$, and the precondition of any operator which is the responsible operator for an induced transition in $oDTG_x^+$, with $x \in V \setminus \{x_0\}$. As discussed in Section 5, this is a super-set of the facts possibly needed in $P_1^+$.

- By $F_0 := s \cup \bigcup_{o \in P_{<0}^+(s)} \mathrm{eff}_o$ we denote the set of facts true after the relaxed execution of $P_{<0}^+(s)$ in $s$. As discussed in Section 5, if $p \notin F_0$ then $p$ is not needed in $s_1$ for $P_1^+$ to be a relaxed plan.

- By $S_1$ we denote the union of: (1) $\mathrm{prev}_{o_0} \cup \mathrm{eff}_{o_0}$; (2) the set of facts $(x, c) \in s$ where there exists no $o$ such that $x \in X_{\mathrm{eff}_o}$ and $o$ is either $o_0$ or in $P_{<0}^+(s)$ or is the responsible operator for an induced transition in $oDTG_x^+$, with $x \in V \setminus \{x_0\}$; (3) the set $F$ defined as $F := \{(x, c) \mid (x, c) \in F_0, x \in V \setminus \{x_0\}\}$ if $X_{\mathrm{eff}_{o_0}} \cap (V \setminus \{x_0\}) = \emptyset$, else $F := \emptyset$. Here, (1) and (2) are facts of which we are certain that they will be true in $s_1$; (3) is a set of facts that we will be able to achieve at the start of $P_1^+$, by appropriately re-ordering the operators.

- If $\overrightarrow{o} = \langle o_1, \ldots, o_n \rangle$ is a sub-sequence of $P^+(s)$, then the *relaxed-plan macro-precondition* of $\overrightarrow{o}$ is defined as $\mathrm{pre}_{\overrightarrow{o}}^+ := \bigcup_{i=1}^n (\mathrm{pre}_{o_i} \setminus \bigcup_{j=1}^{i-1} \mathrm{eff}_{o_j})$. The *relaxed-plan macro-effect* of $\overrightarrow{o}$ is defined as $\mathrm{eff}_{\overrightarrow{o}}^+ := \bigcup_{i=1}^n \mathrm{eff}_{o_i}$. If $\overrightarrow{o}$ is empty then both sets default to the empty set. These notions simply capture the "outside" needs and effects of a relaxed plan sub-sequence.

- *The $oDG^+$-relevant deletes of $t_0$ are $P_{>0}^+(s)$-recoverable* iff $P_{>0}^+(s)$ contains a sub-sequence $\overrightarrow{o_0}$ so that $\mathrm{pre}_{\overrightarrow{o_0}}^+ \subseteq S_1$ and $\mathrm{eff}_{\overrightarrow{o_0}}^+ \supseteq R_1^+ \cap C_0 \cap F_0$. The first condition here ensures that $\overrightarrow{o_0}$ will be applicable at the appropriate point within $P_1^+$. The second clause ensures that all facts relevant for $P_1^+$ will be re-achieved by $\overrightarrow{o_0}$.

We now proceed with our exit path construction. In what follows, we first consider the part of the path leading up to $s_0$, i.e., where we move only the non-leaf variables $x \in V \setminus \{x_0\}$. We show how to construct the relaxed plans $P^+(s')$ for the states $s'$ visited on this path.

First, note that we can assume $P^+(s)$ to be sorted according to the optimal rplan dependency graph $oDG^+ = (V, A)$. Precisely, let $x_k, \ldots, x_1$ be a topological ordering of $V \setminus \{x_0\}$ according to the arcs $A$. Due to the construction of $(V, A)$ as per Definition 1, and because previous values are never removed in the relaxed state space, we can re-order $P^+(s)$ to take the form $P_{<0}^+(s, x_k) \circ \cdots \circ P_{<0}^+(s, x_1) \circ P$. That is, we can perform all moves within each $oDTG_x^+$ up front, in an order conforming with $A$. We will henceforth assume, wlog, that $P^+(s)$ has this form.

Recall in what follows that *original $oDTG_x^+$* transitions are those taken by $P_{<0}^+(s)$, whereas *induced $oDTG_x^+$* transitions are those included as the inverse of an original transition. For a path $\overrightarrow{p}$ of invertible transitions traversing $\langle c_0, \ldots, c_n \rangle$, the *inverse path* $\overleftarrow{p}$ traverses $\langle c_n, \ldots, c_0 \rangle$ by replacing each transition with its inverse. By $\mathrm{rop}(\overrightarrow{p})$ we denote the operator sequence responsible for the path.

We say that a state $s' \in S$ is *in the invertible surroundings of $s$ according to $oDG^+$* if $s'$ is reachable from $s$ by executing a sequence $\overrightarrow{o}$ of responsible operators of invertible/induced transitions in $oDTG_x^+$ for $x \in V \setminus \{x_0\}$. The *adapted relaxed plan* for such $s'$, denoted $P^+(s \rightarrow s')$, is constructed as follows. Let $x_k, \ldots, x_1$ be a topological ordering of $V \setminus \{x_0\}$ according to $A$, and denote $P^+(s) = P^+(s, x_k) \circ \cdots \circ P^+(s, x_1) \circ P$. Initialize $P^+(s \rightarrow s') := P^+(s)$. Then, for each $x_i \in V \setminus \{x_0\}$, let $\overrightarrow{p}$ be a path of original invertible transitions in $oDTG_{x_i}^+$ leading from $s(x_i)$ to $s'(x_i)$ – clearly, such a path must exist. Remove $\mathrm{rop}(\overrightarrow{p})$ from $P^+(s \rightarrow s')$, and insert $\mathrm{rop}(\overleftarrow{p})$ at the start of $P^+(s \rightarrow s', x_i)$.

We next show that adapted relaxed plans indeed are relaxed plans, under restricting conditions that are in correspondence with Definition 2 condition (3):

**Lemma 1.** *Let $(X, s_I, s_G, O)$ be a planning task, let $s \in S$ be a state with $0 < h^+(s) < \infty$, and let $P^+(s)$ be an optimal relaxed plan for $s$. Say that $oDG^+ = (V, A)$ is an optimal rplan* *dependency graph for $P^+(s)$ where, for every $x \in V \setminus \{x_0\}$, the invertible/induced $oDTG^+_x$ transitions have irrelevant side effect deletes and no side effects on $V \setminus \{x_0\}$. Let $s' \in S$ be a state in the invertible surroundings of $s$ according to $oDG^+$. Then $P^+(s \to s')$ is a relaxed plan for $s'$, and $|P^+(s \to s')| \leq |P^+(s)|$.*

*Proof.* By definition, we know that $P^+(s)$ takes the form $P^+_{<0}(s, x_k) \circ \cdots \circ P^+_{<0}(s, x_1) \circ P$, and that $P^+(s \to s')$ takes the form $P^+_{<0}(s', x_k) \circ \cdots \circ P^+_{<0}(s', x_1) \circ P$, where $x_k, \ldots, x_0$ is a topological ordering of $V$, and $P$ is some operator sequence that is common to both, but whose content will not be important for this proof. For simplicity, we denote in the rest of the proof $P^+(s \to s')$ as $P^+(s')$, and we leave away the "$< 0$" subscripts.

Consider first the (relaxed) execution of $P^+(s, x_k)$ and $P^+(s', x_k)$. Say that $\overrightarrow{p}$ is the path in $oDTG^+_{x_k}$ considered in the definition of $P^+(s')$, i.e., a path of original invertible transitions in $oDTG^+_{x_k}$ leading from $s(x_k)$ to $s'(x_k)$. Clearly, $\langle o_1, \ldots, o_n \rangle := \text{rop}(\overrightarrow{p})$ is a sub-sequence of $P^+(s, x_k)$. Say that $\overrightarrow{p}$ visits the vertices $s(x_k) = c_0, \ldots, c_n = s'(x_k)$; denote $C := \{c_0, \ldots, c_n\}$. Assume wlog that $P^+(s, x_k)$ starts with $\langle o_1, \ldots, o_n \rangle$ – note here that we can re-order $P^+(s, x_k)$ (and relaxed plans in general) in any way we want as long as we do not violate operator preconditions. The latter is not the case here because: $\langle o_1, \ldots, o_n \rangle$ constitutes a path in $oDTG^+_{x_k}$; because all other operators depending on a value in $C$ are ordered to occur later on in $P^+(s, x_k)$; and because, since all transitions in $\overrightarrow{p}$ have no side effects on $V \setminus \{x_0\}$, by construction of $(V, A)$ as per Definition 1 the operators in $\langle o_1, \ldots, o_n \rangle$ do not support each other in any way, in $P^+(s)$, other than by affecting the variable $x_k$.

Given the above, wlog $P^+(s, x_k)$ has the form $\langle o_1, \ldots, o_n \rangle \circ P_1$. By construction, $P^+(s', x_k)$ has the form $\text{rop}(\overleftarrow{p}) \circ P_1 =: \langle \overleftarrow{o_n}, \ldots, \overleftarrow{o_1} \rangle \circ P_1$. Consider now the endpoints of the prefixes, i.e., $s_1^+ := s \cup \bigcup_{i=1}^n \text{eff}_{o_i}$ and $s_2^+ := s' \cup \bigcup_{i=n}^1 \text{eff}_{\overleftarrow{o_i}}$. Clearly, since all the transitions on $\overrightarrow{p}$ have irrelevant side effect deletes, we have that the relevant part of $s$ is contained in $s'$. But then, as far as the variables outside $V \setminus \{x_0, x_k\}$ are concerned, the relevant part of $s_1^+$ is contained in $s_2^+$: any relevant side effects of $\langle o_1, \ldots, o_n \rangle$ are already contained in $s'$; the values $C$ are obviously true in $s_2^+$; if the induced transitions have side effects, then these can only increase the fact set $s_2^+$. Further, the sequence $\langle \overleftarrow{o_n}, \ldots, \overleftarrow{o_1} \rangle$ is applicable in the relaxation. To see this, note first that the preconditions on $x_k$ itself are satisfied by definition, because $\langle \overleftarrow{o_n}, \ldots, \overleftarrow{o_1} \rangle$ constitutes a path in $DTG_{x_k}$. Any side effects, if they occur, are not harmful because old values are not over-written in the relaxation. As for preconditions on other variables, due to invertibility – the outside conditions of $\overleftarrow{o_i}$ are contained in those of $o_i$ – those are a subset of those for $\langle o_1, \ldots, o_n \rangle$. Hence, with Definition 1 and since $x_k$ has no incoming edges in $oDG^+$, all these preconditions are satisfied in $s$. They are then also satisfied in $s'$ because ($v_k$ being a root of $oDG^+$) these variables $x$ are not contained in $V$ and hence $s'(x) = s(x)$ by prerequisite – note here that precondition facts cannot have been deleted by the side effects whose deletes are irrelevant by prerequisite.

The above has shown that the relevant part of the outcome of relaxed execution of $P^+(s, x_k)$ in $s$ is contained in the outcome of relaxed execution of $P^+(s', x_k)$ in $s'$, on all variables outside $V \setminus \{x_0, x_k\}$. We can now iterate this argument. Assume as induction hypothesis that we have already shown that the relevant part of the outcome of relaxed execution of $P^+(s, x_k) \circ \ldots P^+(s, x_{i+1})$ in $s$ is contained in the outcome of relaxed execution of $P^+(s', x_k) \circ \cdots \circ P^+(s', x_{i+1})$ in $s'$, on all variables outside $V \setminus \{x_0, x_k, \ldots, x_{i+1}\}$. Now consider $P^+(s, x_i)$ and $P^+(s', x_i)$. The only thing that changes with respect to $x_k$ above is that there may be preconditions on variables $x_j$ that are not true in $s$; we have $j > i$ because such preconditions must belong to predecessors of $x_i$ in $oDG^+$ by Definition 1. Since $P^+(s) = P^+(s, x_k) \circ \cdots \circ P^+(s, x_1) \circ P$ is a relaxed plan for $s$, those conditions are established after relaxed execution of $P^+(s, x_k) \circ \cdots \circ P^+(s, x_{i+1})$ in $s$. Given this, by induction hypothesis the conditions – which are clearly not irrelevant – are established also after relaxed execution of $P^+(s', x_k) \circ \cdots \circ P^+(s', x_{i+1})$ in $s'$, which concludes the argument for the inductive case. With $i = 1$, it follows that the relevant part of the outcome of relaxed execution of $P^+(s, x_k) \circ \cdots \circ P^+(s, x_1)$ in $s$ is contained (on *all* variables) in the outcome of relaxed execution of $P^+(s', x_k) \circ \cdots \circ P^+(s', x_1)$ in $s'$. From this, the claim follows trivially because $P^+(s)$ is a relaxed plan for $s$, and the remainder $P$ of both operator sequences is identical.

The second part of the claim follows because, for any $i \neq j$, we have that the original transitions we use for $x_i$ respectively $x_j$ have no operators in common. This is because, as argued above, all the relevant operators have no side effects on $V \setminus \{x_0\}$. Since each of these operators affects the variable $x_i$, it cannot affect any other variable in $V \setminus \{x_0\}$. Thus, for each inverse transition that we introduce via an inverse operator, $P^+(s)$ contains a separate operator. From this, obviously we get that $|P^+(s \rightarrow s')| \leq |P^+(s)|$. □

Lemma 1 captures the second case of Definition 2 condition (3), transitions that are invertible/induced and have irrelevant side effect deletes and no side effects on $V \setminus \{x_0\}$. The next lemma captures the first case of Definition 2 condition (3):

**Lemma 2.** *Let $(X, s_I, s_G, O)$ be a planning task, let $s \in S$ be a state with $0 < h^+(s) < \infty$, and let $P^+(s)$ be an optimal relaxed plan for $s$. Say that $oDG^+ = (V, A)$ is an optimal rplan dependency graph for $P^+(s)$ where, for every $x \in V \setminus \{x_0\}$, the invertible/induced $oDTG_x^+$ transitions have irrelevant side effect deletes and no side effects on $V \setminus \{x_0\}$. Let $s' \in S$ be a state in the invertible surroundings of $s$ according to $oDG^+$. Let $s''$ be a state reached from $s'$ by a $P^+(s \rightarrow s', x)$ operator $o$ constituting a transition $(c, c')$ for $x \in V$, where $s'(x) = c$, that has self-irrelevant deletes. Then removing $o$ from $P^+(s \rightarrow s')$ yields a relaxed plan for $s''$.*

*Proof.* By Lemma 1, $P^+(s \rightarrow s')$ is a relaxed plan for $s'$. Now, upon execution of $o$, in $s''$, its effects are true, i.e., we have $(x, c')$ and any side effects (if present). On the other hand, obviously the only facts $(z, e)$ that are true in $s'$ but not in $s''$ are in $\text{ctx}(c, c') \cup \{(x, c)\}$. Since, by prerequisite, the transition $(c, c')$ has self-irrelevant deletes, all facts in $\text{ctx}(c, c') \cup \{(x, c)\}$ are either irrelevant or $\text{rop}(c, c')$-only relevant, meaning they are not in the goal and occur in no operator precondition other than, possibly, that of $o$ itself. The claim follows directly from that. □

We remark that a much more easily formulated, and more general, version of Lemma 2 could be proved simply by associating the notion of "self-irrelevant deletes" with operators rather than transitions, and postulating only that $o$ be used in $P^+(s)$. That argument corresponds to part (A) in the proof to Lemma 3 in the author's previous work (Hoffmann, 2005). We state the argument in the particular form above since that will be the form we need below.

We are now almost ready to prove the main lemma behind our exit path construction. We need one last notation, capturing a simpler form of the cost function $\text{cost}^{\text{d*}}(oDG^+)$ that we considered in Section 5. The simpler function does not make use of the "short-cut" construction; that construction will be introduced separately further below. We define $\text{cost}^{\text{d}}(oDG^+) := \sum_{x \in V} \text{cost}^{\text{d}}(x)$, where $\text{cost}^{\text{d}}(x) :=$

$$
\begin{cases}
1 & x = x_0 \\
\text{diam}(oDTG_x^+) * \sum_{x':(x,x') \in A} \text{cost}^{\text{d}}(x') & x \neq x_0
\end{cases}
$$

**Lemma 3.** *Let $(X, s_I, s_G, O)$ be a planning task, let $s \in S$ be a state with $0 < h^+(s) < \infty$, and let $P^+(s)$ be an optimal relaxed plan for $s$. Say that $oDG^+ = (V, A)$ is a successful optimal rplan dependency graph for $P^+(s)$. Then there exists an operator sequence $\overrightarrow{o}$ so that:*

(I) *$\overrightarrow{o}$ constitutes a monotone path in $S$ from $s$ to a state $s_1$ with $h^+(s) > h^+(s_1)$.*

(II) *The length of $\overrightarrow{o}$ is at most $\text{cost}^{\text{d}}(oDG^+)$ if we have Definition 2 condition (2a) or (2b), and is at most $\text{cost}^{\text{d}}(oDG^+) + 1$ if we have Definition 2 condition (2c).*

*Proof.* Let $x_k, \dots, x_1$ be a topological ordering of $V \setminus \{x_0\}$ according to the arcs $A$. Consider a state $s_0$ where for every $x \in V \setminus \{x_0\}$ we have that $s_0(x)$ is a vertex in $oDTG_x^+$, and for every variable $x$ outside $V \setminus \{x_0\}$ we have that $s_0(x) = s(x)$ unless $s(x)$ is irrelevant. Say that $\text{pre}_{o_0} \subseteq s_0$. Note first that such a state $s_0$ exists. By definition, we have that either $\text{pre}_{o_0}(x_0)$ is undefined or that $\text{pre}_{o_0}(x_0) = s(x_0) = s_0(x_0)$. (Note that "for every variable $x$ outside $V \setminus \{x_0\}$ we have that $s_0(x) = s(x)$ unless $s(x)$ is irrelevant" covers also the case where a transition on $V \setminus \{x_0\}$ has a side effect on $x_0$, whose delete must then by prerequisite be irrelevant and thus either the side effect is $x_0 := s(x_0)$ or $o_0$ is not actually preconditioned on $x_0$.) By Definition 1 and because $P^+(s)$ is a relaxed plan for $s$, each variable $x \in X_{\text{pre}_{o_0}}$ is contained in $V$ unless $\text{pre}_{o_0}(x) = s(x)$. For the same reasons, by construction of $oDTG_x^+$, we have that $\text{pre}_{o_0}(x)$ is a vertex in $oDTG_x^+$.

Now, consider the state $s_1$ that results from applying $o_0$ to $s_0$. We first consider the situation where $s_0$ is in the invertible surroundings of $s$ according to $oDG^+$; the opposite case will be discussed further below. We can apply Lemma 1 to $s_0$, and hence have a relaxed plan $P^+(s{\rightarrow}s_0)$ for $s_0$ that results from replacing, in $P^+(s)$, some moves of $P_{<0}^+(s,x)$, for $x \in V \setminus \{x_0\}$, with their inverses. In particular, $h^+(s) \geq h^+(s_0)$, and $P^+(s{\rightarrow}s_0, x') = P^+(s, x')$ for all $x' \notin V$. What is a relaxed plan for $s_1$? We distinguish Definition 2 condition (2) cases (a), (b), and (c).

In case (a), by definition we have that $P_{>0}^+(s)$ contains a sub-sequence $\overrightarrow{o_0}$ so that $\text{pre}_{\overrightarrow{o_0}}^{\pm} \subseteq S_1$ and $\text{eff}_{\overrightarrow{o_0}}^{\pm} \supseteq R_1^+ \cap C_0 \cap F_0$. This implies that we can remove $o_0$ from $P^+(s{\rightarrow}s_0)$ and obtain a relaxed plan $P_1^+$ for $s_1$, thus getting $h^+(s) > h^+(s_1)$. More precisely, we construct $P_1^+$ by: removing $o_0$ from $P^+(s{\rightarrow}s_0)$; if $X_{\text{eff}_{o_0}} \cap (V \setminus \{x_0\}) \neq \emptyset$ then moving $\overrightarrow{o_0}$ to occur at the start of $P_1^+$; if $X_{\text{eff}_{o_0}} \cap (V \setminus \{x_0\}) = \emptyset$ then moving $\overrightarrow{o_0}$ to occur at the start of $P_{>0}^+(s)$ (which is unchanged in $P^+(s{\rightarrow}s_0)$).

Observe first that $o_0 \in P^+(s{\rightarrow}s_0)$ and $\overrightarrow{o_0}$ is a sub-sequence of $P^+(s{\rightarrow}s_0)$ since the adaptation pertains exclusively to operators that precede $o_0$ in $P^+(s)$. Second, of course the values established by $o_0$ are true in $s_1$.

Third, $\overrightarrow{o_0}$ is applicable (in the relaxation) at its assigned point in $P_1^+$. To see this, consider first the case where $X_{\text{eff}_{o_0}} \cap (V \setminus \{x_0\}) \neq \emptyset$. Then, by definition of $S_1$, $\text{pre}_{\overrightarrow{o_0}}^{\pm}$ is contained in $(\text{prev}_{o_0} \cup \text{eff}_{o_0})$ and the set of facts $(x, c) \in s$ where there exists no $o$ such that $x \in X_{\text{eff}_o}$ and $o$ is either $o_0$ or in $P_{<0}^+(s)$ or is the responsible operator for the inverse of a transition taken by an operator $o' \in P_{<0}^+(s)$. All these facts will be true in $s_1$. This is obvious for $\text{prev}_{o_0} \cup \text{eff}_{o_0}$ and follows for the other facts because they were true in $s$ and cannot have been affected by any operator on the path to $s_1$. Consider now the case where $X_{\text{eff}_{o_0}} \cap (V \setminus \{x_0\}) = \emptyset$. By definition of $S_1$, $\text{pre}_{\overrightarrow{o_0}}^{\pm}$ is contained in the previous sets of facts, plus $\{(x, c) \mid (x, c) \in F_0, x \in V \setminus \{x_0\}\}$. The latter facts, as far as relevant, will all be true at the start of $\overrightarrow{o_0}$ in $P_1^+$. This is because execution of $o_0$ does not affect the execution of $P^+(s \rightarrow s_0)$, and thus of $P_1^+$, up to this point. But then, with what was argued in Lemma 1, we have that the outcome of such execution in $s_0$ contains, on the variables $V \setminus \{x_0\}$, the relevant part of the outcome of $P_{<0}^+(s)$ in $s$ – that is, the relevant part of $F_0$. Since $o_0$ does not affect these variables, the same is true of $s_1$, which concludes this point.

Finally, consider any facts $(z, e)$ that are true in $s_0$ but not in $s_1$, and that may be needed by $P_1^+$ behind $\overrightarrow{o_0}$, i.e., that either are in the goal or in the precondition of any of these operators. Observe that, since inverse operators are performed only for transitions on variables $V \setminus \{x_0\}$, and since they do not include any new outside preconditions, any such $(z, e)$ is contained in $R_1^+$.[24] Now, say first that $(z, e) \in F_0$. Then, with the above, $(z, e) \in (\text{ctx}(s(x_0), c) \cup \{(x_0, s(x_0))\}) \cap F_0 \cap R_1^+$ and thus $(z, e) \in \text{eff}_{\overrightarrow{o_0}}^{\pm}$ by prerequisite and we are done. What if $(z, e) \notin F_0$? Note that, then, $(z, e) \notin \text{pre}_o$ for any $o \in P_{<0}^+(s)$ – else, this precondition would not be true in the relaxed execution of $P^+(s)$ and thus $P^+(s)$ would not be a relaxed plan. Neither is $(z, e)$ added by any $o \in P_{<0}^+(s)$, and thus $(z, e)$ is not needed as the precondition of any inverse operator used in $P^+(s \rightarrow s_0)$ – these operators do not introduce new outside preconditions, and of course use only own-preconditions previously added by other operators affecting the respective variable. Thus the only reason why $(z, e)$ could be needed in $P_1^+$ is if either $(z, e) \in s_G$ or $(z, e) \in \text{pre}_o$ for some $o \in P_{>0}^+(s)$. If $(z, e) \in s_G$ then certainly, since $P^+(s)$ is a relaxed plan, it is achieved by some operator $o$ in $P^+(s)$. We cannot have $o = o_0$ since the effect of $o_0$ is true in $s_1$, and we cannot have $o \in P_{<0}^+(s)$ since $(z, e) \notin F_0$. Thus $o \in P_{>0}^+(s)$, and thus $o$ is contained in $P_1^+$ and we are done. If $(z, e) \in \text{pre}_{o'}$ for some $o' \in P_{>0}^+(s)$, the same arguments apply, i.e., there must be $o \in P_{>0}^+(s)$, ordered before $o'$, that adds $(z, e)$. This concludes the proof for case (a).

Consider now case (b), where $s(x_0) \notin R_1^+$ and the transition $(s(x_0), c)$ has replaceable side effect deletes, i.e., $\text{ctx}(s(x_0), c) \cap s_G = \emptyset$ and, for every $o_0 \neq o \in O$ where $\text{pre}_o \cap \text{ctx}(s(x_0), c) \neq \emptyset$ there exists $o' \in O$ so that $\text{eff}_{o'} = \text{eff}_o$ and $\text{pre}_{o'} \subseteq \text{prev}_{o_0} \cup \text{eff}_{o_0}$. We obtain a relaxed plan for $P_1^+$ by removing $o_0$ from $P^+(s \rightarrow s_0)$, and replacing any other operators $o$ with the respective $o'$ if needed. Precisely, say that $(z, e)$ is true in $s_0$ but not in $s_1$. If $z = x_0$ then $e = s(x_0)$ is not needed in $P_1^+$ by construction. For every other $z$, we must have $(z, e) \in \text{ctx}(s(x_0), c)$. Then $(z, e)$ is not a goal by prerequisite. For any operator $o \in P_1^+$ that has $(z, e)$ as a precondition, we can replace $o$ with the postulated operator $o_1$ that is obviously applicable in $s_1$ and has the same effect. This concludes this case.

Consider last case (c), where by definition $s(x_0) \notin R_1^+$, and the transition $(s(x_0), c)$ has recoverable side effect deletes. Here, the guarantee to decrease $h^+$ is obtained not for $s_1$ itself, but for a successor state $s_2$ of $s_1$. Namely, let $\overline{o_0}$ be the operator recovering the relevant side effect deletes of $(s(x_0), c)$. Precisely, let $\psi \in \text{ctx}(s(x_0), c)$ so that $\psi \subseteq s_0$ (such a $\psi$ exists by definition of $\text{ctx}(s(x_0), c)$). Then, let $\overline{o_0}$ be an operator so that $\text{pre}_{\overline{o_0}} \subseteq (\text{prev}_{o_0} \cup \text{eff}_{o_0})$ and $\text{eff}_{\overline{o_0}} \subseteq \psi$, $\text{eff}_{\overline{o_0}} \supseteq \psi \cap (s_G \cup \bigcup_{o_0 \neq o' \in O} \text{pre}_{o'})$ (such an operator exists by case (b)). Say that we obtain $P_1^+$ by replacing, in $P^+(s \to s_0)$, $o_0$ with $\overline{o_0}$. Then $P_1^+$ is a relaxed plan for $s_1$. To see this, note first that $\overline{o_0}$ is applicable in $s_1$ by virtue of $\text{pre}_{\overline{o_0}} \subseteq (\text{prev}_{o_0} \cup \text{eff}_{o_0})$. Further, note that the only values deleted by $o_0$ are those in $\psi$ plus $(x_0, s_0(x_0))$. Since $s_0(x_0) = s(x_0)$, by $s(x_0) \notin R_1^+$ we know that $s_0(x_0) \notin R_1^+$ and thus this delete is of no consequence. As for $\psi$, by virtue of $\text{eff}_{\overline{o_0}} \supseteq \psi \cap (s_G \cup \bigcup_{o_0 \neq o' \in O} \text{pre}_{o'})$ all facts that could possibly be relevant are re-achieved by $\overline{o_0}$. Finally, the values established by $o_0$ are true in $s_1$.

---

24. Note in particular the special case of inverse transitions on non-leaf variables $x$, which may have a precondition in $x$ that is added by, but not needed as a prerequisite of, the operators in $P^+(s, x)$. Such preconditions – and only such preconditions – may be needed in $P^+(s \rightarrow s_0)$ and thus in $P_1^+$, but not in $P^+(s)$. It is for this reason that we include these facts in the definition of $R_1^+$.

Now, say we obtain $s_2$ by applying $\overline{o_0}$ in $s_1$. Then removing $\overline{o_0}$ from $P_1^+$ yields a relaxed plan for $s_2$. This is simply because its established effects are true in $s_2$, and by virtue of $\text{eff}_{\overline{o_0}} \subseteq \psi$ the only facts it deletes are side-effects of the transition $(s(x_0), c)$. By case (c), these are not relevant for anything except possibly the recovering operators. The recovering operator $\overline{o_0}$ we have just removed from $P_1^+$. As for any other recovering operators $\bar{o}$ that could still be contained in $P_1^+$, since $\text{eff}_{\bar{o}} \subseteq \psi$ and $\text{eff}_{\overline{o_0}} \supseteq \psi \cap (s_G \cup \bigcup_{o_0 \neq o' \in O} \text{pre}_{o'})$, all relevant facts that $\bar{o}$ could possibly achieve are already true in $s_2$ and thus we can remove $\bar{o}$ as well. Hence, overall, $h^+(s) > h^+(s_2)$.

In cases (a) and (b) we can prove (I) by constructing a monotone path to $s_1$, in case (c) the same is true of $s_2$. (Of course, we will also show (II), by constructing a path that has at most the specified length; we will ignore this issue for the moment.) The only difficulty in constructing such a path is achieving the preconditions of $o_0$. These preconditions may not be satisfied in $s$, so we need to reach the state $s_0$ where they are satisfied. We need to do so without ever increasing the value of $h^+$. Note that, if we *decrease* the value of $h^+$ somewhere along the way, then we have already reached an exit on a monotone path, and are done. Thus in what follows we will only show the upper bound $h^+(s)$. With Lemma 1, this bounding can be accomplished by starting at $s$, and always taking only $oDTG_x^+$ transitions of variables $x \in V$ pertaining to the second case in Definition 2 condition (3), i.e., transitions that are invertible/induced and have irrelevant side effect deletes and no side effects on $V \setminus \{x_0\}$. In what follows we will, for brevity, refer to such transitions as "case2". Note here that, this way, we will reach only states in the invertible surroundings of $s$ according to $oDG^+$. For any such operator sequence $\vec{o}$, by Lemma 1 we know that $h^+(s) \geq h^+(s')$ for all states $s'$ along the way. Now, what if we cannot reach $s_0$ by using such a sequence, i.e., what if we would have to take a non-case2 $oDTG_x^+$ transition $(c, c')$ of variable $x$, at some state $s'$? By prerequisite we know that transition $(c, c')$ has self-irrelevant deletes. We can apply Lemma 2 because: $s'$ is in the invertible surroundings of $s$ according to $oDG^+$; since we're following a transition path, clearly $s'(x) = c$, i.e., the value of the relevant variable in $s'$ is the start value of the last transition we are taking; and by construction, $P^+(s \to s')$ changes $P^+(s)$ only in the case2 transitions, and thus the responsible operator $\text{rop}(c, c')$ (which is not case2) is guaranteed to be contained in $P^+(s \to s')$. Note here that $\text{rop}(c, c')$ cannot be used in any of the case2 transitions for any other $V \setminus \{x_0\}$ variable we might have taken on the path to $s'$, because by prerequisite all these transitions have no side effects on $V \setminus \{x_0\}$, in contradiction to $o$ constituting a transition for the variable $x$ at hand. Thus we know that $h^+(s) > h^+(s')$ so we have already constructed our desired monotone path to an exit and can stop. Else, if we can reach $s_0$ by such a sequence $\overrightarrow{o}$, then with the above, $\overrightarrow{o} \circ \langle o_0 \rangle$ (respectively $\overrightarrow{o} \circ \langle o_0, \overline{o_0} \rangle$, in case (c)) constitutes the desired path.

It remains to show how exactly to construct the operator sequence $\overrightarrow{o}$. Consider a topological ordering of $V$, $x_k, \ldots, x_1$. In what follows, we consider "depth" indices $k \geq d \geq 0$, and we say that a variable $x \in V$ "has depth" $d$, written $depth(x) = d$, iff $x = x_d$. Each $d$ characterizes the $d$-abstracted planning task which is identical to the original planning task except that all (and only) those outside preconditions, of all $oDTG_x^+$ transitions for variables $x$ where $depth(x) \leq d$, are removed that pertain to values of variables $x'$ where $depth(x') > d$. We prove by induction over $d$ that:

(*) For the $d$-abstracted task, there exists an operator sequence $\overrightarrow{o}_d$ so that:

(a) either (1) $\overrightarrow{o}_d \circ \langle o_0 \rangle$ is an execution path applicable in $s$, or (2) $\overrightarrow{o}_d$ is an execution path applicable in $s$, and the last transition $(c, c')$ for variable $x$ taken in $\overrightarrow{o}_d$ is relevant, has self-irrelevant deletes, its responsible operator is contained in the adapted relaxed plan for the state $s'$ it is applied to, and $s'(x) = c$;

(b) $\overrightarrow{o}_d$, except in the last step in case (2) of (a), uses only case2 $oDTG_x^+$ transitions for variables $x$ with $1 \leq depth(x) \leq d$;

(c) the number of operators in $\overrightarrow{o}_d \circ \langle o_0 \rangle$ pertaining to any $x \in V$ is at most $cost^d(x)$.

Our desired path $\overrightarrow{o}$ then results from setting $d := k$. To see this, note that the $k$-abstracted planning task is identical to the original planning task. The claim then follows with our discussion above: (a) and (b) together mean that $h^+$ decreases monotonically on $\overrightarrow{o}_d$ and is less than $h^+(s)$ at its end. Given (c), the length of $\overrightarrow{o}_d$ is bounded by $\sum_{x \in V, depth(x) \leq d} cost^d(x)$. This proves the claim when adding the trivial observation that, if we have Definition 2 condition (2) case (c) as discussed above, then we need to add one additional operator at the end of the path.

We now give the proof of (*). The base case, $d = 0$, is trivial. Just set $\overrightarrow{o}_0$ to be empty. By the construction of $(V, A)$ as per Definition 1, and by construction of the 0-abstracted task, all outside preconditions of $o_0$ are either true in $s$ or have been removed. All of (a) (case (1)), (b), (c) are obvious.

Inductive case, $d \rightarrow d + 1$. Exploiting the induction hypothesis, let $\overrightarrow{o}_d$ be the operator sequence as per (*). We now turn $\overrightarrow{o}_d$ into the requested sequence $\overrightarrow{o}_{d+1}$ for the $d + 1$-abstracted planning task.

For the remainder of this proof, we will consider $oDTG_x^+$, for any $x \in V \setminus \{x_0\}$, to contain also any irrelevant transitions, i.e., we omit this restriction from Definition 1. This is just to simplify our argumentation – as we will show, the $oDTG_x^+$ paths we consider do not contain any irrelevant transitions, and hence are contained in the actual $oDTG_x^+$ as per Definition 1.

Let $o$ be the first operator in $\overrightarrow{o}_d \circ \langle o_0 \rangle$. $o$ may not be applicable in $s$, in the $d + 1$-abstracted planning task. The only reason for that, however, may be a precondition that was removed in the $d$-abstracted planning task but that is not removed in the $d + 1$-abstracted planning task. By construction, that precondition must pertain to $x_{d+1}$. Say the precondition is $(x_{d+1}, c)$. By induction hypothesis, we know that $o$ is contained in $P_{<0}^+(s)$, or is responsible for an inverse transition of such an operator. In both cases, since inverse transitions introduce no new outside preconditions, $(x_{d+1}, c)$ is a precondition of an operator in $P^+_{<0}(s)$. Thus $c$ is a vertex in $oDTG^+_{x_{d+1}}$ – this is trivial if $(x_{d+1}, c)$ is true in $s$ (which actually cannot be the case here because else $o$ would be applicable in $s$ in the $d+1$-abstracted planning task), and if $(x_{d+1}, c)$ is not true in $s$ it follows because $P^+(s)$ is a relaxed plan and must thus achieve $(x_{d+1}, c)$ before it is needed as a precondition. Hence, $P^+_{<0}(s, x_{d+1})$ must contain a shortest path $\overrightarrow{q}$ in $oDTG^+_{x_{d+1}}$ from $s(x_{d+1})$ to $c$. All the transitions on the path are not irrelevant. To see this, note first that the endpoint is an operator precondition by construction, and thus the last transition $(c_1, c)$ is not irrelevant. But then, neither is the previous transition, $(c_2, c_1)$: if it was, then $(x_{d+1}, c_1)$ would be in no operator precondition; but then, rop$(c_1, c)$ – which is contained in $P^+_{<0}(s)$ by construction – would also constitute the transition $(c_2, c)$ in $oDTG^+_{x_{d+1}}$ and thus $\overrightarrow{q}$ would not be a shortest path in contradiction. Iterating the argument, $\overrightarrow{q}$ does not contain any irrelevant transitions. Thus, since depth$(x_{d+1}) = d + 1$, by Definition 1 (which includes all non-satisfied preconditions of relevant transitions) and by construction of the $d+1$-abstracted planning task, all the outside preconditions used in rop$(\overrightarrow{q})$ are either true in $s$ or have been removed. Hence we can execute rop$(\overrightarrow{q})$. We do so until either we have reached the end of the sequence, or until the last transition taken in $oDTG^+_{x_{d+1}}$ was not case2, and hence has self-irrelevant deletes by prerequisite. In the latter case, since we are following a path and since as discussed above the adapted relaxed plan exchanges only operators pertaining to case2 transitions and thus not the last one we just executed, we clearly have attained (a) case (2) and can stop – the part of rop$(\overrightarrow{q})$ that we executed is, on its own, an operator sequence $\overrightarrow{o}_{d+1}$ as desired. In the former case, we reach a state $s'$ where $s'(x_{d+1}) = c$ (and nothing else of relevance has been deleted, due to the non-existence of relevant side-effect deletes). In $s'$, $o$ can be applied, leading to the state $s''$.

Let now $o'$ be the second operator in $\overrightarrow{o}_d \circ \langle o_0 \rangle$. Like above, if $o'$ is not applicable in $s''$, then the only reason may be an unsatisfied precondition of the form $(x_{d+1}, c')$. Like above, $o'$ or its inverse is contained in $P^+_{<0}(s)$, and hence $c'$ is a vertex in $oDTG^+_{x_{d+1}}$. Likewise, $s''(x_{d+1}) = c$ is a vertex in $oDTG^+_{x_{d+1}}$. Now, we have not as yet used any non-case2 transition in $oDTG^+_{x_{d+1}}$, or else we wouldn't get here. This means that we are still in the invertible surroundings around $s(x_{d+1})$ of $oDTG^+_{x_{d+1}}$. Clearly, this implies that there exists a path in $oDTG^+_{x_{d+1}}$ from $c$ to $c'$ (we could simply go back to $s(x_{d+1})$ and move to $c'$ from there). Taking the shortest such path $\overrightarrow{q}$, clearly the path length is bounded by the diameter of $oDTG^+_{x_{d+1}}$. The path does not contain any irrelevant transitions – the endpoint $c'$ has been selected for being an operator precondition, the values in between are part of a shortest path in $oDTG^+_{x_{d+1}}$, and thus the same argument as given above applies. Thus the outside preconditions used by the operators constituting $\overrightarrow{q}$ are either true in $s$ or have been removed – this follows from the construction of $(V, A)$ as per Definition 1 and by construction of the $d+1$-abstracted planning task for operators in $P^+_{<0}(s)$, and follows for inverses thereof because inverse operators introduce no new outside preconditions. Hence we can execute $\overrightarrow{q}$ in $s''$. We do so until either we have reached the end of the path, or until the last transition taken was not case2, and hence has self-irrelevant deletes by prerequisite.

Consider the latter case. The state $s'$ just before the last transition is reached only by case2 transitions, and since the transition is in $oDTG^+_{x_{d+1}}$ but not case2, the responsible operator must be contained in $P^+(s)$ and with that in the adapted relaxed plan $P^+(s \rightarrow s')$ for $s'$ – recall here that, as pointed out above, since case2 transitions are postulated to have no side effects on $V \setminus \{x_0\}$, the responsible operator cannot be used by any of them. Further, clearly since we are following a path of transitions, we have that the value of $x_{d+1}$ in $s'$ is the start value of the transition. Hence we have attained (a) case (2) and can stop. In the former case, we have reached a state where $o'$ can be applied (and nothing of relevance has been deleted, due to the postulated non-existence of relevant side-effect deletes, for case2 transitions). Iterating the argument, we get to a state where the last operator of $\overrightarrow{o}_d \circ \langle o_0 \rangle$ can be applied, by induction hypothesis reaching a state $s_1$ as desired by (a) case (1).

Properties (a) and (b) are clear from construction. As for property (c), to support any operator of $\overrightarrow{o}_d \circ \langle o_0 \rangle$, clearly in the above we apply at most $\mathrm{diam}(oDTG^+_{x_{d+1}})$ operators pertaining to $x_{d+1}$ (or we stop the sequence earlier than that). Note further that, for all operators $o$ in $\overrightarrow{o}_d \circ \langle o_0 \rangle$ with unsatisfied preconditions on $x_{d+1}$ in the above, if $o$ pertains to variable $x$ then we have $(x_{d+1}, x) \in A$. This is a consequence of the construction of $(V, A)$ as per Definition 1, and the fact that inverse transitions do not introduce new outside preconditions. Thus, in comparison to $\overrightarrow{o}_d \circ \langle o_0 \rangle$, overall we execute at most

$$\mathrm{diam}(oDTG^+_{x_{d+1}}) * \sum_{x:(x_{d+1}, x) \in A} k(x)$$

additional operators in $\overrightarrow{o}_{d+1} \circ \langle o_0 \rangle$, where $k(x)$ is the number of operators in $\overrightarrow{o}_d \circ \langle o_0 \rangle$ pertaining to variable $x$. By induction hypothesis, property (c) of (*), we have that $k(x) \leq \mathrm{cost}^{\mathrm{d}}(x)$, for all $x$ with $\mathrm{depth}(x) < d+1$, and thus for all $x$ with $(x_{d+1}, x) \in A$. Hence we get, for the newly inserted steps affecting $x_{d+1}$, the upper bound

$$\mathrm{diam}(oDTG^+_{x_{d+1}}) * \sum_{x:(x_{d+1}, x) \in A} \mathrm{cost}^{\mathrm{d}}(x)$$

which is identical to $\mathrm{cost}^{\mathrm{d}}(x_{d+1})$. This concludes the argument. $\qquad \square$

We next note that we can improve the exit distance bound in case we do not insist on *monotone* exit paths:

**Lemma 4.** *Let $(X, s_I, s_G, O)$ be a planning task, let $s \in S$ be a state with $0 < h^+(s) < \infty$, and let $P^+(s)$ be an optimal relaxed plan for $s$. Say that $oDG^+ = (V, A)$ is a successful optimal rplan dependency graph for $P^+(s)$. Let $V^* \subseteq V \setminus \{x_0\}$ so that, for every $x \in V^*$, all $oDTG^+_x$ transitions are invertible/induced and have irrelevant side effect deletes and no side effects on $V \setminus \{x_0\}$, and all other $DTG_x$ transitions either are irrelevant, or have empty conditions and irrelevant side effect deletes. Then there exists an operator sequence $\overrightarrow{o}$ so that:*

*(I)* $\overrightarrow{o}$ *constitutes a path in $S$ from $s$ to a state $s_1$ with $h^+(s) > h^+(s_1)$.*

*(II) The length of $\overrightarrow{o}$ is at most $\mathrm{cost}^{\mathrm{d*}}(oDG^+)$ if we have Definition 2 condition (2a) or (2b), and is at most $\mathrm{cost}^{\mathrm{d*}}(oDG^+) + 1$ if we have Definition 2 condition (2c).*

*Proof.* This is a simple adaptation of Lemma 3, and we adopt in what follows the terminology of the proof of that lemma. The only thing that changes is that the bound imposed on exit path length is sharper, and that we do not insist on that path being monotone. At the level of the proof mechanics, what happens is that, whenever $x_{d+1} \in V^*$, when we choose a path $\overrightarrow{q}$ to achieve the next open precondition of an operator $o$ already chosen to participate in $\overrightarrow{o}_d \circ \langle o_0 \rangle$, then we do not restrict ourselves to paths within $oDTG^+_{x_{d+1}}$, but allow also any shortest path through $DTG_{x_{d+1}}$. Being a shortest path in $DTG_{x_{d+1}}$ to a value that occurs as an operator precondition, $\overrightarrow{q}$ contains no irrelevant transitions (same argument as in the proof of Lemma 3). Further, $\overrightarrow{q}$ will be executable because by prerequisite the alternative (non-$oDTG^+_x$) transitions in it have no outside conditions; for original/induced transitions, precondition achievement works exactly as before. Note here the important property that open preconditions to be achieved for $x_{d+1}$ will only ever pertain to values contained in $oDTG^+_{x_{d+1}}$. This is trivial to see by induction because alternative transitions do not have any outside preconditions. Since by prerequisite any deletes of the alternative transitions are irrelevant, executing them does no harm – all we need is a minor extension to Lemma 1, allowing $s'$ to be identical with a state $s''$ in the invertible surroundings of $s$, modulo a set of irrelevant values that hold in $s''$ but not in $s$; it is obvious that this extension is valid. With this extension, it is also obvious that the arguments pertaining to $s_0$ and $s_1$ remain valid. Finally, consider the case where $\overrightarrow{q}$ involves a non-case2 $oDTG^+_{x_{d+1}}$ transition. Then the state where this transition is applied is in the invertible surroundings of $s$. This holds for any $x \notin V^*$ because for these our construction remains the same. It holds for any $x \in V^*$ because, first, alternative transitions have no outside conditions, hence cause no higher-depth transitions to be inserted in between, hence the value of all lower-depth variables $x$ is in $oDTG^+_x$; second, by prerequisite, $oDTG^+_x$ does not contain any non-case2 transitions, and thus the value of $x$ we're at clearly can be reached by case2 transitions. □

**Theorem 2.** *Let $(X, s_I, s_G, O)$, $s$, $P^+(s)$, and $oDG^+$ be as in Definition 1. If $oDG^+$ is successful, then $s$ is not a local minimum, and $ed(s) \leq \mathrm{cost}^{\mathrm{d*}}(oDG^+)$. If we have Definition 2 condition (2a) or (2b), then $ed(s) \leq \mathrm{cost}^{\mathrm{d*}}(oDG^+) - 1$.*

*Proof.* This is a direct consequence of Lemmas 3 and 4. □

We note that the prerequisites of Lemma 4 could be weakened by allowing, for $x \in V^*$, outside conditions that are already true in $s$. This extension obviously does not break the proof arguments. We have omitted it here to not make the lemma prerequisite even more awkward than it already is.

As indicated, the exit path constructed in Lemma 4 is *not* necessarily monotone. Example 5 in Appendix A.4 contains a construction showing this.

### A.3 Conservative Approximations

For the sake of self-containedness of this section, we re-state the definitions given in Section 6:

**Definition 3.** *Let $(X, s_I, s_G, O)$ be a planning task, let $s \in S$ with $0 < h^+(s) < \infty$, let $x_0 \in X_{s_G}$, and let $t_0 = (s(x_0), c)$ be a relevant transition in $DTG_{x_0}$ with $o_0 := \mathrm{rop}(t_0)$.*

*A local dependency graph for $s$, $x_0$, and $o_0$, or local dependency graph in brief, is a graph $lDG = (V, A)$ with unique leaf vertex $x_0$, and where $x \in V$ and $(x, x') \in A$ if either: $x' = x_0$, $x \in X_{\mathrm{pre}_{o_0}}$, and $\mathrm{pre}_{o_0}(x) \neq s(x)$; or $x' \in V \setminus \{x_0\}$ and $(x, x')$ is an arc in $SG$.*

*A* global dependency graph *for* $x_0$ *and* $o_0$, or global dependency graph *in brief*, is a graph $gDG = (V, A)$ with unique leaf vertex $x_0$, and where $x \in V$ and $(x, x') \in A$ if either: $x' = x_0$ and $x_0 \neq x \in X_{\text{pre}_{o_0}}$; or $x' \in V \setminus \{x_0\}$ and $(x, x')$ is an arc in $SG$.

**Definition 4.** *Let* $(X, s_I, s_G, O)$, $s$, $t_0$, $o_0$, *and* $G = lDG$ *or* $G = gDG$ *be as in Definition 3. We say that* $G = (V, A)$ *is* successful *if all of the following holds:*

*(1) $G$ is acyclic.*

*(2) If $G = lDG$ then $s_G(x_0) \neq s(x_0)$, and there exists no transitive successor $x'$ of $x_0$ in $SG$ so that $x' \in X_{s_G}$ and $s_G(x') \neq s(x')$.*

*(3) We have that $t_0$ either:*

   *(a) has self-irrelevant side effect deletes; or*

   *(b) has replaceable side effect deletes; or*

   *(c) has recoverable side effect deletes.*

*(4) For $x \in V \setminus \{x_0\}$, all $DTG_x$ transitions either are irrelevant, or have self-irrelevant deletes, or are invertible and have irrelevant side effect deletes and no side effects on $V \setminus \{x_0\}$.*

**Lemma 5.** *Let* $(X, s_I, s_G, O)$ *be a planning task, and let* $s \in S$ *be a state with* $0 < h^+(s) < \infty$. *Say that* $x_0 \in X$ *and, for every* $o_0 = \text{rop}(s(x_0), c)$ *in* $DTG_{x_0}$ *where* $t_0 = (s(x_0), c)$ *is relevant,* $lDG_{o_0}$ *is a successful local dependency graph for* $s$, $x_0$, *and* $o_0$. *Then, for at least one of the* $o_0$, *there exist an optimal relaxed plan* $P^+(s)$ *for* $s$, *and a successful optimal rplan dependency graph* $oDG^+$ *for* $P^+(s)$, $x_0$, *and* $o_0$, *where* $oDG^+$ *is a sub-graph of* $lDG_{o_0}$.

*Proof.* Observe first that Definition 4 property (2) forces any relaxed plan $P^+(s)$ to move $x_0$, i.e., we have that $P^+(s, x_0)$ is non-empty. In particular, $P^+(s, x_0)$ takes a path in $DTG_{x_0}$ from $s(x_0)$ to $s_G(x_0)$. Let $\overrightarrow{q}$ be a shortest such path taken by $P^+(s, x_0)$, and let $o_0$ be the responsible operator of the first transition in $\overrightarrow{q}$. Clearly, this transition has the form $(s(x_0), c)$, i.e., $o_0$ is one of the operators $o_0$ in the claim. Lying on a shortest path from $s(x_0)$ to $s_G(x_0)$ in the sub-graph of $DTG_{x_0}$ taken by $P^+(s, x_0)$, the transition $(s(x_0), c)$ is not irrelevant. This can be seen with exactly the same argument as given in the proof to Lemma 3 for the transitions on the paths $\overrightarrow{q}$ constructed there, except that the endpoint is now a goal instead of an operator precondition.

Next, observe that any optimal $P^+(s)$ contains at most one operator $o$ with $x_0 \in X_{\text{pre}_o}$ and $\text{pre}_o(x_0) = s(x_0)$. This also follows from Definition 4 property (2): $x_0$ cannot become important for any non-achieved goal, i.e., no $P^+(s)$ operator outside $P^+(s, x_0)$ relies on a precondition on $x_0$. To see this, assume that such an operator $o$ does exist. Then, since $P^+(s)$ is optimal, there exists a "reason" for the inclusion of $o$. Precisely, $o$ must achieve at least one fact that is "needed" in the terms of Hoffmann and Nebel (2001b): a fact that is either in the goal or in the precondition of another operator $o'$ behind $o$ in $P^+(s)$. Iterating this argument for $o'$ (if necessary), we obtain a sequence $o = o_1, (x_1, c_1), o_2, (x_2, c_2), \ldots, o_n, (x_n, c_n)$ where $(x_n, c_n)$ is a goal fact not satisfied in $s$ and where $o_i$ achieves $(x_i, c_i)$ in $P^+(s)$. Obviously, $SG$ then contains a path from $x_0$ to $x_n$, and $x_n \in X_{s_G}$ and $s_G(x_n) \neq s(x_n)$, in contradiction to Definition 4 property (2). Thus such $o$ does not exist. With the same argument, it follows also that every operator in $P^+(s, x_0)$ either has no side effect used elsewhere in the relaxed plan, or has no precondition on $x_0$. Thus those operators in $P^+(s, x_0)$ that are preconditioned on $x_0$ serve only to transform $s(x_0)$ into $s_G(x_0)$. Of course, then, at most a single one of these operators relies on $s(x_0)$ or else $P^+(s)$ is not optimal.

Say in what follows that $lDG_{o_0} = (V, A)$. Denote by $(V', A')$ the result of backchaining by Definition 1 from $o_0$ with $P^+_{<0}(s)$. Definition 3 will include all variables and arcs included by Definition 1. To see this, just note that all arcs $(x, x')$ included by Definition 1 are due to relevant transitions. Hence $(V', A')$ is a sub-graph of $(V, A)$. In particular, since $(V, A)$ is acyclic, $(V', A')$ is acyclic as well.

Our next observation is that, assuming that Definition 4 condition (2) holds true, Definition 4 condition (3a) implies Definition 2 condition (2a), Definition 4 condition (3b) implies Definition 2 condition (2b), and Definition 4 condition (3c) implies Definition 2 condition (2c).

Consider first case (a) where $t_0$ has self-irrelevant side effect deletes. We show that $R_1^+ \cap C_0 = \emptyset$. Recall here the notations of Appendix A.2 – $C_0 = \{(x_0, s(x_0))\} \cup \text{ctx}(t_0)$, and $R_1^+$ is a super-set of the set of facts that we will need for the relaxed plan after removing $o_0$. For all variables except $x_0$, it is clear that there is no fact in this intersection: all facts in $\text{ctx}(t_0)$ are irrelevant or $o_0$-only relevant by prerequisite, and are thus not contained in $R_1^+$. Hence, $(x_0, s(x_0))$ remains as the only possible content of $R_1^+ \cap C_0$. We show in what follows that $(x_0, s(x_0)) \notin R_1^+$, and thus $(x_0, s(x_0)) \notin R_1^+ \cap C_0$ and the latter intersection is empty, as desired. Recall that $R_1^+$ denotes the union of $s_G$, the precondition of any $o_0 \neq o \in P^+(s)$, and the precondition of any operator which is the responsible operator for an induced transition in $oDTG_x^+$, with $x \in V \setminus \{x_0\}$. By Definition 4 condition (2), $(x_0, s(x_0)) \notin s_G$. As argued above, $o_0$ is the only operator in $P^+(s)$ that may be preconditioned on $s(x_0)$ and thus it is not in the precondition of any $o_0 \neq o \in P^+(s)$. Lastly, say that $p$ is a precondition of a responsible operator for an induced transition in $oDTG_x^+$, the corresponding original transition being $t$. Then, since inverse transitions do not introduce any new conditions, $p \in \text{cond}(t)$ and thus $p \in \text{pre}_{\text{rop}(t)}$ where, by definition, $\text{rop}(t) \in P^+_{<0}(s)$. But then, since $o_0 \neq \text{rop}(t) \in P^+(s)$, we have $(x_0, s(x_0)) \notin \text{pre}_{\text{rop}(t)}$, which implies that $p \neq (x_0, s(x_0))$. Thus $(x_0, s(x_0)) \notin R_1^+$ like we needed to show.

Consider now case (b) where $t_0$ has recoverable side effect deletes. To show Definition 2 condition (2b) for $o_0 = \text{rop}(t_0)$, all we need to prove is that $s(x_0)$ is not $oDG^+$-relevant, i.e., that $s(x_0) \notin R_1^+$. This was already shown above.

For case (c), $t_0$ has replaceable side effect deletes. Again, to show Definition 2 condition (2c) for $t_0$, all we need to prove is that $s(x_0)$ is not $oDG^+$-relevant.

Consider finally the conditions imposed on non-leaf variables $x \in V \setminus \{x_0\}$, i.e., Definition 4 condition (4) and Definition 2 condition (3). By Definition 4 condition (4), the $DTG_x$ transitions of every $x \in V \setminus \{x_0\}$ either are irrelevant, or have self-irrelevant deletes, or are invertible and have irrelevant side effect deletes and no side effects on $V \setminus \{x_0\}$. If a $DTG_x$ transitions is irrelevant then it cannot be in $oDTG_x^+$, thus the 2nd or 3rd case is true of the $oDTG_x^+$ transitions of every $x \in V' \setminus \{x_0\}$. This concludes the argument. □

**Theorem 3.** *Let $(X, s_I, s_G, O)$ be a planning task, and let $s \in S$ be a state with $0 < h^+(s) < \infty$. Say that $x_0 \in X$ so that, for every $o_0 = \text{rop}(s(x_0), c)$ in $DTG_{x_0}$ where $(s(x_0), c)$ is relevant, $lDG_{o_0}$ is a successful local dependency graph. Then $s$ is not a local* minimum, and $ed(s) \leq \max_{o_0} \text{cost}^{\text{D*}}(lDG_{o_0})$. If, for every $lDG_{o_0}$, we have Definition 4 condition (3a) or (3b), then $ed(s) \leq \max_{o_0} \text{cost}^{\text{D*}}(lDG_{o_0}) - 1$.

*Proof.* By Lemma 5, for some choice of $o_0 = \text{rop}(s(x_0), c)$ there exists an optimal relaxed plan $P^+(s)$ and a successful optimal rplan dependency graph $oDG^+ = (V', A')$ for $P^+(s)$, so that $oDG^+$ is a sub-graph of $lDG_{o_0}$ with the same unique leaf vertex $x_0$. We can apply Lemma 3 and obtain that $s$ is not a local minimum.

To see the other part of the claim, let $V^{**}$ be defined as in Section 6, i.e., $V^{**}$ is the subset of $V \setminus \{x_0\}$ for which all $DTG_x$ transitions either are irrelevant, or are invertible and have empty conditions, irrelevant side effect deletes, and no side effects on $V \setminus \{x_0\}$. Then, for each $DTG_x$ transition $t$ where $x \in V^{**}$, $t$ satisfies both the restriction required by Lemma 4 on $oDTG_x^+$ transitions – if $t$ is irrelevant, then it cannot be in $oDTG_x^+$, else it is invertible and has irrelevant side effect deletes and no side effects on $V \setminus \{x_0\}$ – and the restriction required by Lemma 4 on the other transitions – either irrelevant, or empty conditions and irrelevant side effect deletes. We can hence apply Lemma 4 to $oDG^+$, and obtain a (not necessarily monotone) path to an exit, with length bound $\text{cost}^{\text{d*}}(oDG^+)$ if $(s(x_0), c)$ has irrelevant side effect deletes or replaceable side effect deletes, and $\text{cost}^{\text{d*}}(oDG^+) + 1$ if $(s(x_0), c)$ has recoverable side effect deletes. It thus suffices to show that $\text{cost}^{\text{D*}}(lDG_{o_0}) \geq \text{cost}^{\text{d*}}(oDG^+)$. That, however, is obvious because $V \supseteq V'$, $\text{cost}^{\text{D*}}(x) \geq 0$ for all $x$, and $\text{maxPath}(DTG_x) \geq \text{diam}(oDTG_x^+)$ for all $x \in V'$. $\qquad\square$

**Theorem 4.** *Let $(X, s_I, s_G, O)$ be a planning task. Say that all global dependency graphs $gDG$ are successful. Then $S$ does not contain any local minima and, for any state $s \in S$ with $0 < h^+(s) < \infty$, $ed(s) \leq \max_{gDG} \text{cost}^{\text{D*}}(gDG)$. If, for every $gDG$, we have Definition 4 condition (3a) or (3b), then $ed(s) \leq \max_{gDG} \text{cost}^{\text{D*}}(gDG) - 1$.*

*Proof.* Let $s \in S$ be a state. We need to prove that $s$ is no local minimum. If $h^+(s) = 0$ or $h^+(s) = \infty$, there is nothing to show. Else, assume that the variables $X$ are topologically ordered according to the strongly connected components of $SG$, and let $x_0 \in X$ be the uppermost variable so that $x_0 \in X_{s_G}$ and $s_G(x_0) \neq s(x_0)$; obviously, such $x_0$ exists. Clearly, the only chance for $x_0$ to not satisfy Definition 4 condition (2) – "there exists no transitive successor $x'$ of $x_0$ in $SG$ so that $x' \in X_{s_G}$ and $s_G(x') \neq s(x')$" – is if there exists $x'$ in the same strongly connected $SG$ component, with $x' \in X_{s_G}$ (and $s_G(x') \neq s(x')$). But then, there exists a transition $t'$ in $DTG_{x'}$ with an outside condition eventually leading, by backwards chaining in $SG$, to $x_0$. Let $gDG'$ be the global dependency graph for $x'$ and $\text{rop}(t')$ (such a $gDG'$ exists because $x' \in X_{s_G}$). Since Definition 3 includes all transitive $SG$-predecessors of $x'$ pertaining to the conditions of $t'$, $gDG'$ includes $x_0$. But then, since $x_0$ and $x'$ lie in the same strongly connected component, Definition 3 eventually reaches $x'$. Thus $gDG'$ contains a cycle, in contradiction to the prerequisite. It follows that the strongly connected $SG$ component of $x_0$ contains only $x_0$, and thus Definition 4 condition (2) holds true.

Now, say that $o_0$ is responsible for a relevant transition of the form $(s(x_0), c)$ in $DTG_{x_0}$. Then there exists a local dependency graph $lDG$ for $s$, $x_0$, and $o_0$ so that $lDG$ is a sub-graph of $gDG$. This follows from the simple observation that Definition 3 will include, for $gDG$, all variables and arcs that it will include for $lDG$. (Note here that any precondition of $o_0$ on $x_0$, if present, is satisfied in $s$ because $o_0 = \text{rop}(s(x_0), c)$, and thus Definition 3 will not include $x_0$ as a predecessor for achieving $o_0$ preconditions in $lDG$.)

Obviously, given the above, $lDG$ is successful. Since this works for any choice of not-irrelevant $(s(x_0), c)$, we can apply Theorem 3. The claim follows directly from this and the fact that $\text{cost}^{\text{D*}}(gDG) \geq \text{cost}^{\text{D*}}(lDG)$. The latter is obvious because $\text{cost}^{\text{D*}}$ increases monotonically when adding additional variables. □

### A.4 Example Constructions

Our first example shows that, even within the scope of our basic result, operators are not necessarily respected by the relaxation, i.e., an operator may start an optimal real plan yet not occur in any optimal relaxed plan.

**Example 1.** *Consider the planning task in Figure 4. Variables are shown (in dark green) on the left hand side of their respective DTG. Circles represent variable values, and lines represent DTG transitions. Transitions with a condition are longer lines, with the condition inscribed below the line (in blue). For each variable, a dashed arrow indicates the value in the initial state $s_I$. Where a goal value is defined, this is indicated by a circled value. Where needed, we will refer to the operators responsible for a transition in terms of the respective variable followed by the indices of the start and end value. For example, the operator moving $x$ from $c_1$ to $c_2$ will be referred to as "x12". We abbreviate states $\{(x, c), (y, d)\}$ as $(c, d)$. We stick to these conventions throughout this section.*

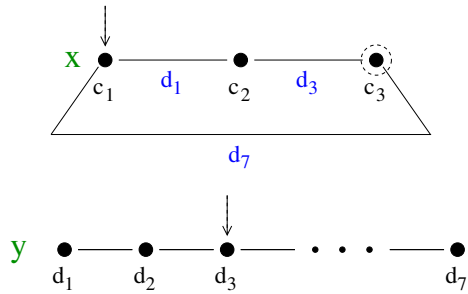

Figure 4: Planning task underlying Example 1.

As shown in Figure 4, the DTG of $x$ consists of three vertices whose connection requires the conditions $d_1$ and $d_3$, or alternatively $d_7$ as a shortcut. The domain of $y$ is a line of length 6 requiring no conditions.

Clearly, the support graph of this planning task is acyclic, and all transitions in all DTGs have no side effects and are invertible. However, operator $y34$ (for example) is not respected by the relaxation. To see this, note first that $h^+(s_I) = 4$: the only optimal relaxed plan is $\langle y32, y21, x12, x23 \rangle$ because the relaxed plan ignores the need to "move back" to $d_3$ for operator $x23$. On the other hand, the only optimal (real) plan for $s_I$ is $\langle y34, y45, y56, y67, x17 \rangle$. If we choose to use $y32$ instead, like the optimal relaxed plan does, then we end up with the sequence $\langle y32, y21, x12, y12, y23, x23 \rangle$ which is 1 step longer. Hence, in $s_I$, $y34$ starts an optimal plan, but does not start an optimal relaxed plan.

We next give three examples showing how local minima can arise in very simple situations generalizing our basic result only minimally. We consider, in this order: cyclic support graphs; non-invertible transitions; transitions with side effects.

**Example 2.** *Consider the planning task in Figure 5.*

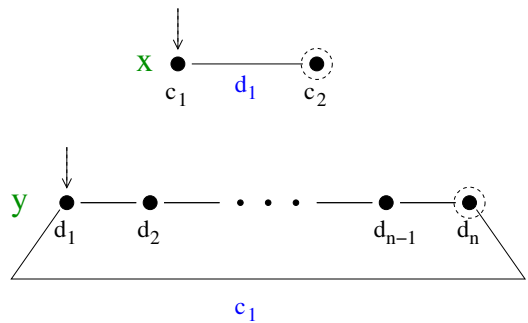

Figure 5: Planning task underlying Example 2.

*The DTG of $x$ is just two vertices whose connection requires the condition $d_1$. The domain of $y$ is a line of length $n$ requiring no conditions, with a shortcut between $d_1$ and $d_n$ that requires $c_1$ as condition. Clearly, all transitions in all DTGs have no side effects and are invertible. However, SG contains a cycle between $x$ and $y$ because they mutually depend on each other. We will show now that this mutual dependence causes the initial state $s_I = \{(x, c_1), (y, d_1)\}$ to be a local minimum, for $n \geq 5$. We abbreviate, as before, states $\{(x, c), (y, d)\}$ as $(c, d)$. We have $h^+(s_I) = 2$: the only optimal relaxed plan is $\langle x12, y1n \rangle$. Now consider the operators applicable to $s_I = (c_1, d_1)$:*

- *Execute $x12$, leading to $s_1 = (c_2, d_1)$ with $h^+(s_1) = 2$ due to $\langle x21, y1n \rangle$. From here, the only new state to be reached is via $y12$, giving $s_2 = (c_2, d_2)$ with $h^+(s_2) = 3$ due to $\langle y21, x21, y1n \rangle$. (Note here that $n - 2 \geq 3$ by prerequisite, so a relaxed plan composed of $yi(i + 1)$ operators also has $\geq 3$ steps.) We have $h^+(s_2) > h^+(s_I)$ so this way we cannot reach an exit on a monotone path.*

- *Execute $y12$, leading to $s_3 = (c_1, d_2)$ with $h^+(s_3) = 3$ due to $\langle y21, x12, y1n \rangle$. (Note here that $n - 2 \geq 3$ by prerequisite, so a relaxed plan moving $y$ by $ypp$ operators has $\geq 4$ steps.) Again, the path is not monotone.*

- *Execute $y1n$, leading to $s_4 = (c_1, d_n)$ with $h^+(s_4) = 2$ due to $\langle yn1, x12 \rangle$. From here, the only new state to be reached is via $yn(n-1)$, giving $s_5 = (c_1, d_{n-1})$ with $h^+(s_5) = 3$ due to $\langle y(n-1)n, yn1, x12 \rangle$. (Note here that $n-2 \geq 3$ by prerequisite, so a relaxed plan moving $y$ to $d_1$ via $d_{n-2}, \ldots, d_2$ has $\geq 3 + 2$ steps.) Again, the path is not monotone.*

*No other operators are applicable to $s_I$, thus we have explored all states reachable from $s_I$ on monotone paths. None of those states is an exit, proving that $s_I$ is a local minimum (as are $s_1$ and $s_4$). There is, in fact, only a single state $s$ with $h^+(s) = 1$, namely $s = (c_2, d_{n-1})$. Clearly, reaching $s$ from $s_I$ takes $n - 1$ steps: first apply $x12$, then traverse $d_2, \ldots, d_{n-2}$. So the exit distance of $s_I$ is $n - 3$, thus this distance is unbounded.*

In Section 9, the following modification of Example 2 is considered. We set $n := 2$, i.e., the domain of $y$ is reduced to the two values $d_1, d_2$; and we remove the line $d_2, \ldots, d_{n-2}$, i.e., $y$ can move only via what was previously the short-cut. This modified example falls into the SAS$^+$-PUBS tractable class identified by Bäckström and Klein (1991), and it still contains a local minimum (the example is unsolvable, though).

**Example 3.** *Consider the planning task in Figure 6.*

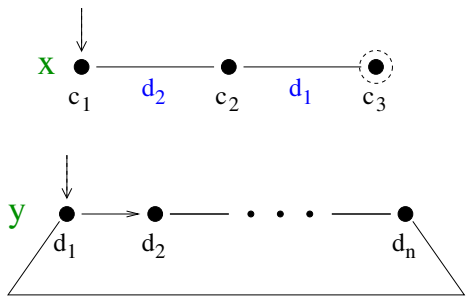

Figure 6: Planning task underlying Example 3. The arrow between $d_1$ and $d_2$ indicates that the respective DTG transition is directed, i.e., there exists no transition from $d_2$ to $d_1$.

*The DTG of $x$ is three vertices whose connection requires (starting from the initial value $c_1$) first condition $d_2$, then condition $d_1$. The domain of $y$ is a circle of length $n$ requiring no conditions, and being invertible except for the arc from $d_1$ to $d_2$.*

*Clearly, the support graph is acyclic and all transitions in all DTGs have no side effects. However, the non-invertible arc from $d_1$ to $d_2$ causes the initial state $s_I = (c_1, d_1)$ to be a local minimum for all $n \geq 3$. This is very easy to see. We have $h^+(s_I) = 3$ due to the only optimal relaxed plan $\langle y12, x12, x23 \rangle$. Note here that the relaxed plan does not have to "move $y$ back" because $(y, d_1)$ is still true after executing $y12$. Now, the operators applicable to $s_I$ are $y12$ and $y1n$. The latter, reaching the state $s_n = (c_1, d_n)$, immediately increases the value of $h^+$. This is because, with $n \geq 3$, $y1n$ does not get $y$ closer to $d_2$, while moving it farther away from $d_1$ (both of which need to be achieved). The shortest relaxed for $s_n$ is $\langle yn1, y12, x12, x23 \rangle$. Alternatively, say we apply $y12$ in $s_I$, reaching the state $s_2 = (c_1, d_2)$. We have $h^+(s_2) = n + 1$: we need to apply, in the relaxation, $x12$, $n - 1$ steps to complete the circle from $d_2$ back to $d_1$, and $x23$. Thus, for $n \geq 3$, $s_2$ has a larger $h^+$ value than $s_I$. It follows that $s_I$ is a local minimum. The nearest exit to $s_I$ is $s_{n-1} = (c_2, d_{n-1})$: $s_{n-1}$ has the relaxed plan $\langle y(n-1)n, yn1, x23 \rangle$ of length 3, and after applying $y(n-1)n$ we get $h^+$ value 2. Reaching $s_{n-1}$ from $s_I$ takes 1 step moving $x$ and $n - 2$ steps moving $y$. So the exit distance of $s_I$ is $n - 1$, thus this distance is unbounded.*

**Example 4.** *Consider the planning task in Figure 7.*

*The DTG of $x$ consists of two kinds of transitions. First, there is a line $c_1, \ldots, c_n$ of transitions requiring no conditions. Second, there are direct links, called short-cuts in what follows, between $c_n$ and every other $c_i$, conditioned on value $d_1$ of $y$. The DTG of $y$ contains just two vertices that are connected unconditionally. Moving from $d_1$ to $d_2$ has the side-effect $c_n$. (That side-effect is responsible for the "towards-$c_n$ direction" of the short-cuts in the DTG of $x$.)*

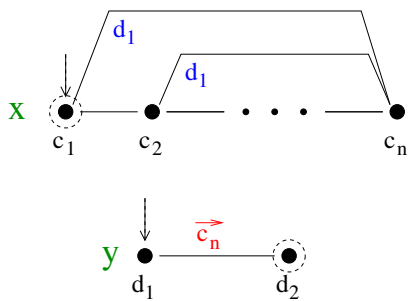

Figure 7: Planning task underlying Example 4. The (red) inscription $c_n$ above the line between $d_1$ and $d_2$ indicates that the transition from $d_1$ to $d_2$ has the side effect $c_n$.

The support graph is acyclic. Its only arc goes from $y$ to $x$, due to the short-cuts in the DTG of $x$, and due to the operator $y12$ which has an effect on $x$ and a precondition on $y$. The transitions are all invertible; in particular each short-cut has both, a direction towards $c_n$ and vice versa. However, the side-effect of $y12$ causes the initial state $s_I = (c_1, d_1)$ to be a local minimum for all $n \geq 3$.

We have $h^+(s_I) = 1$ due to the only optimal relaxed plan $\langle y12 \rangle$. Note here that the relaxed plan does not care about the side effect of $y12$, because $c_1$ is still true afterward. Now, if we apply any operator in $s_I$ that leaves $c_1$, then clearly we increase $h^+$ by 1: no matter what move we make, the relaxed plan must include both $y12$ and a move back to $c_1$. The only other available option in $s_I$ is to apply $y12$. We get the state $s_1 = (c_n, d_2)$. There, $h^+(s_1) = 2$ as well, because the relaxed plan needs to re-achieve $c_1$. Since $n \geq 3$, doing so via the unconditional sequence $c_n, \ldots, c_1$ takes $\geq 2$ steps. The only alternative is to use the short-cut $xn1$ from $c_n$ to $c_1$; doing so involves applying $y21$ in the first place, giving us a relaxed plan of length 2. Hence all direct successors of $s_I$ have a heuristic value $> 1$, and so $s_I$ is a local minimum. Note also that the exit distance of $s_I$ grows with $n$. The nearest exit is a state from which the goal can be reached in a single step. Clearly, the only such state is $(c_2, d_2)$. The shortest path to that state, from $s_I$, applies $y12$ and then moves along the unconditional line $c_n, \ldots, c_2$, taking $1 + (n-2) = n - 1$ steps.

We next show that the exit path constructed using "short-cuts", leading to the improved bound $\mathrm{cost}^{\mathrm{d*}}$ instead of $\mathrm{cost}^{\mathrm{d}}$, may be non-monotone, and that the improved bound may indeed under-estimate the length of a shortest monotone exit path.

**Example 5.** *Consider the planning task in Figure 8.*

*In this example, the only optimal relaxed plan for the initial state moves $z$ along the path $e_0, \ldots, e_{2n}$ – note here that all these values are needed for moving $y$ – then moves $y$ to $d_{2k+2n}$, then moves $x$ to $c_1$. This gives a total of $h^+(s_I) = 2n + (2k + 2n) + 1 = 4n + 2k + 1$ steps.*

*The only operators applicable to $s_I$ move $z$. If we move along the line $e_0, \ldots, e_{2n}$, then $h^+$ remains constant: we always need to include the moves back in order to achieve the own goal of $z$. Once we reach $e_{2n}$, we can move $y$ one step, then need to move $z$ back, etc. During all these moves, up to the state where $y = d_{2k+2n}$, as long as $z$ stays within* $e_0, \ldots, e_{2n}$, $h^+$ remains constant. To see this, observe first that of course it suffices for a relaxed plan to reach once, with $z$, all the values on this line, taking $2n$ moves wherever we are on the line; the moves for $y$ are as before. Second, observe that indeed all these moves are needed: wherever $y$ is on the line $d_0, \ldots, d_{2k+2n}$, it needs to move to $d_{2k+2n}$ in order to suit $x$, and it needs to move to $d_0$ to suit its own goal. Every value in $e_0, \ldots, e_{2n}$ appears as a condition of at least one of these $y$ moves. Thus, from $s_I$, the nearest exit reached this way is the state $s$ where $y = d_{2k+2n}$ and $z = e_{2n}$: there, we can move $x$ to $c_1$ which decreases $h^+$ to $4n + 2k$. The length of the exit path $\overrightarrow{o}$ we just described, from $s_I$ to $s$, obviously is $2k * (2n + 1) + 2n * 2 = 4kn + 2k + 4n$.

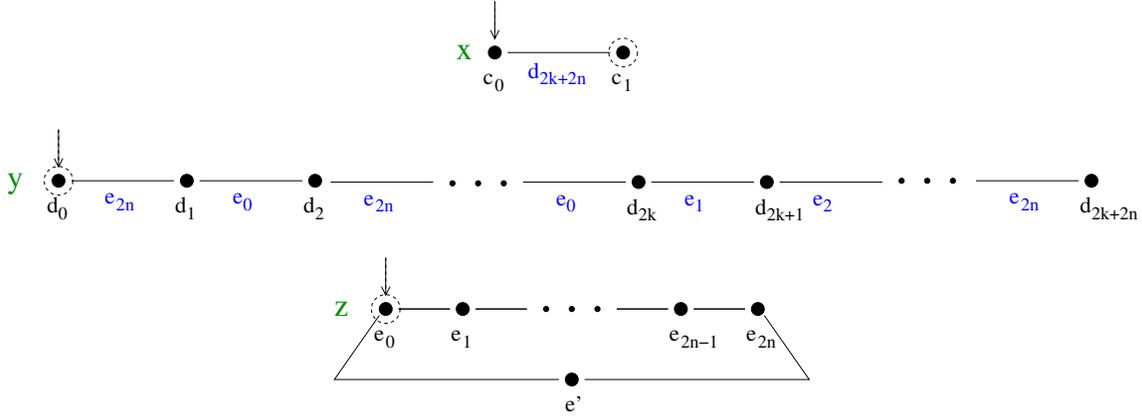

Figure 8: Planning task underlying Example 5.

What happens if we move $z$ to $e'$? Consider first that we do this in $s_I$. Then $h^+$ increases to $4n + 2k + 2$: we need to reach all values on the line $e_0, \ldots, e_{2n}$, which from $e'$ takes one step more. The same argument applies for any state traversed by $\overrightarrow{o}$, because, as argued, in any such state we still need to reach all values on the line $e_0, \ldots, e_{2n}$. Thus $\overrightarrow{o}$ is the shortest monotone path to an exit.

The only optimal rplan dependency graph $oDG^+$ for $s_I$ is the entire $SG$, and $oDTG_z^+$ contains all of $DTG_z$ except $e'$. The only global dependency graph $gDG$ is the entire $SG$.

Clearly, in $s_I$, the next required value to reach for any variable is $e_{2n}$, so the construction in the proof to Theorem 2 will first try to reach that value. When using "short-cuts" as accounted for by $\text{cost}^{d*}(.)$, the exit path constructed will move to $e_{2n}$ via $e'$ rather than via the line $e_0, \ldots, e_{2n}$, and thus as claimed this exit path is not monotone.

Finally, consider the bound returned by $\text{cost}^{d*}(oDG^+)$. Obviously, $\text{cost}^{d*}(oDG^+) = \text{cost}^{D*}(gDG)$. We obtain the bound $(-1) + \text{cost}^{d*}(oDG^+) = (-1) + 1[\text{cost}^{d*}(x)] + 1 * (2k + 2n)[\text{cost}^{d*}(x) * \text{diam}(oDTG_y^+)] + (2k + 2n) * (n + 1)[\text{cost}^{d*}(y) * \text{diam}(DTG_z)]$. Note here that $\text{diam}(DTG_z) = n + 1$ because $DTG_z$ is a circle with $2n + 2$ nodes. Overall, we have $(-1) + \text{cost}^{d*}(oDG^+) = (2k + 2n) * (n + 2) = 2kn + 4k + 2n^2 + 4n$. For sufficiently large $k$, this is less than $4kn + 2k + 4n$, as claimed. In detail, we have $4kn + 2k + 4n > 2kn + 4k + 2n^2 + 4n$ iff $2kn - 2k > 2n^2$ iff $kn - k > n^2$ iff $k > \frac{n^2}{n-1}$. This holds, for example, if we set $n := 2$ and $k := 5$.

The reader will have noticed that Example 5 is very contrived. The reason why we need such a complicated unrealistic example is that $\text{cost}^d$, and with that $\text{cost}^{d*}$, contains two sources of over-estimation, cf. the discussion in Section 5. In particular, every move of non-leaf variables is supposed to take a whole $oDTG^+/DTG$ diameter. To show that $cost^{d*}$ is not in general an upper bound on the length of a monotone exit path, we thus need the presented construction around $k$ so that its under-estimation – considering $diam(DTG_z)$ instead of $diam(oDTG_z^+)$ – outweighs this over-estimation. Importantly, constructing examples where the "short-cuts" temporarily increase $h^+$ (but $cost^{d*}$ nevertheless delivers an upper bound on monotone exit path length) is much easier. All that needs to happen is that, for whatever reason, we have a variable $z$ like here, where the currently required value ($e_{2n}$ in Example 5) is reached in $oDTG_z^+$ values along an unnecessarily long path all of whose values are needed in the relaxed plan. This happens quite naturally, e.g., in transportation domains if the same vehicle needs to load/unload objects along such a longer path.

We now demonstrate that, in a case where our analyses apply, exit distance may be exponential.

**Example 6.** *Consider the planning task in Figure 9.*

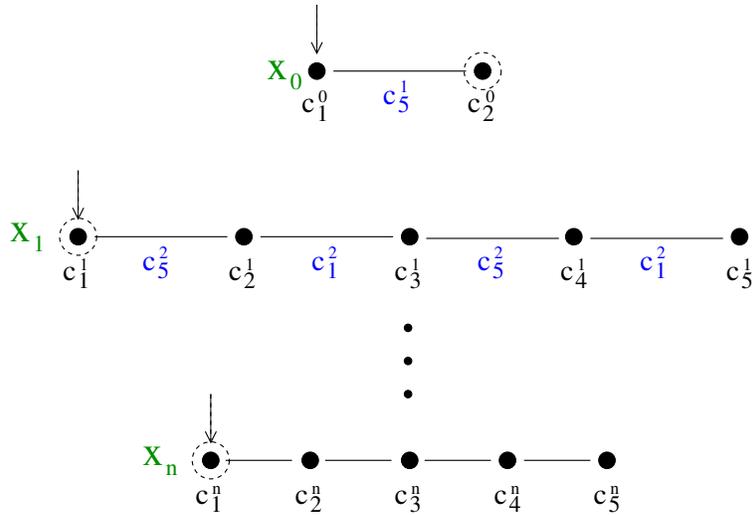

Figure 9: Planning task underlying Example 6.

*The DTG of $x_0$ is two vertices whose connection is conditioned on $c_5^1$. For all other variables $x_i$, we have five vertices on a line, alternatively requiring the last vertex $c_5^{i+1}$ of $x_{i+1}$ and the first vertex $c_1^{i+1}$ of $x_{i+1}$. Clearly, the only optimal rplan dependency graph $oDG^+$ for $s_I$, and the only global dependency graph $gDG$ for the task is the full support graph $SG$. This is acyclic, and all transitions are invertible and have no side effects, thus our analyses apply.*

*What are $h^+(s_I)$ and $ed(s_I)$? For a relaxed plan, we need to move $x_0$ to $c_2^0$. Due to the conditioning, for each variable both "extreme" values – left and right hand side – are required so we need 4 moves for each $x_i$ with $1 \leq i \leq n$. Thus $h^+(s_I) = 1 + 4n$.*

*Now, consider any state $s$ where $s(x_0) = c_1^0$. To construct a relaxed plan, obviously we still need 1 move for $x_0$. We also still need 4 moves for each other variable. Consider $x_1$. If $s(x_1) = c_1^1$ then we need to move it to $c_5^1$ in order to be able to move $x_0$. If $s(x_1) = c_2^1$ then we need to move it to $c_5^1$ in order to be able to move $x_0$, and to $c_1^1$ for its own goal, and so forth. In all cases, all four transitions must be taken in the relaxed plan. Due to the conditioning, recursively the same is true for all other variables. Thus, $h^+(s) = 1 + 4n$.*

This means that the nearest exit is a state $s'$ where $x_0$ has value $c_1^0$ and $x_1$ has value $c_5^1$: in $s'$, we can move $x_0$ and afterward, definitely, $4n$ steps suffice for a relaxed plan. What is the distance to a state $s'$? We need to move $x_1$ four times. Let's denote this as $d(x_1) = 4$. Each move requires 4 moves of $x_2$, so $d(x_2) = 16$. The sequence of moves for $x_2$ "inverses direction" three times. At these points, $x_3$ does not need to move so $d(x_3) = (d(x_2) - 3) * 4$. Generalizing this, we get $d(x_{i+1}) = [d(x_i) - (\frac{d(x_i)}{4} - 1)] * 4 = 3d(x_i) + 4$, so the growth over $n$ is exponential.

Obviously, Example 6 also shows that plan length can be exponential in cases where Theorem 4 applies. We remark that Example 6 is very similar to an example given by Domshlak and Dinitz (2001). The only difference is that Domshlak and Dinitz's example uses different conditions for transitions to the left/to the right, which enables them to use smaller DTGs with only 3 nodes. In our setting, we cannot use different conditions because we need the transitions to be invertible. This causes the "loss" of exit path steps in those situations where the next lower variable "inverses direction" and thus relies on the same outside condition as in the previous step. Indeed, for DTGs of size 3, this loss of steps results in a polynomially bounded exit distance. The recursive formula for $d(x_i)$ becomes $d(x_{i+1}) = [d(x_i) - (\frac{d(x_i)}{2} - 1)] * 2 = d(x_i) + 2$, resulting in $ed(s_I) = n^2 + n$. On the other hand, $\text{cost}^{\text{d*}}$ and $\text{cost}^{\text{D*}}$ still remain exponential in this case, because they do not consider the loss incurred by inversing directions. Precisely, it is easy to see that $\text{cost}^{\text{d*}}(oDG^+) = \text{cost}^{\text{D*}}(gDG) = 1 + \sum_{i=1}^{n} 2^i = 2^{n+1} - 1$. This proves that these bounds can over-estimate by an exponential amount.

The next example shows that the exit path constructed (implicitly) by our analyses may be exponentially longer than an optimal plan for the task.

**Example 7.** *Consider the planning task in Figure 10.*

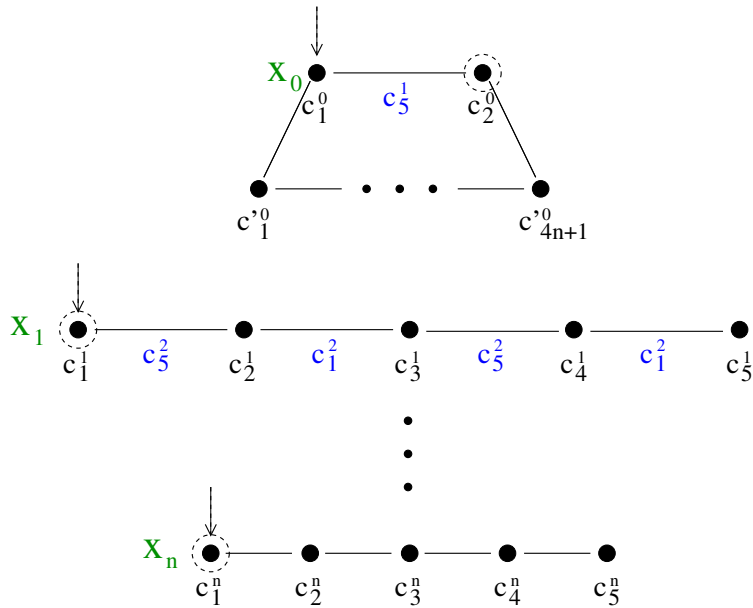

Figure 10: Planning task underlying Example 7.

*In this example, the only optimal relaxed plan for the initial state is the same as in Example 6, because the "alternative" route via $c'_{01}, \ldots, c'_{0(4n+1)}$ takes $1 + 4n + 1 = 4n + 2 > 4n + 1$ steps. Thus the exit path constructed remains the same, too, with length exponential in n. However, the length of the shortest plan is $4n + 2$.*

Note in Example 7 that the observed weakness – being guided into the "wrong" direction – is caused by a weakness of optimal relaxed planning, rather than by a weakness of our analysis. The relaxation overlooks the fact that moving via $x_1, \ldots, x_n$ will incur high costs due to the need to repeatedly undo and re-do conditions achieved beforehand. Note also that, in this example too, we get an exponential over-estimation of exit distance.

We finally show that feeding Theorem 2 with non-optimal relaxed plans does not give any guarantees:

**Example 8.** *Consider the planning task in Figure 11.*

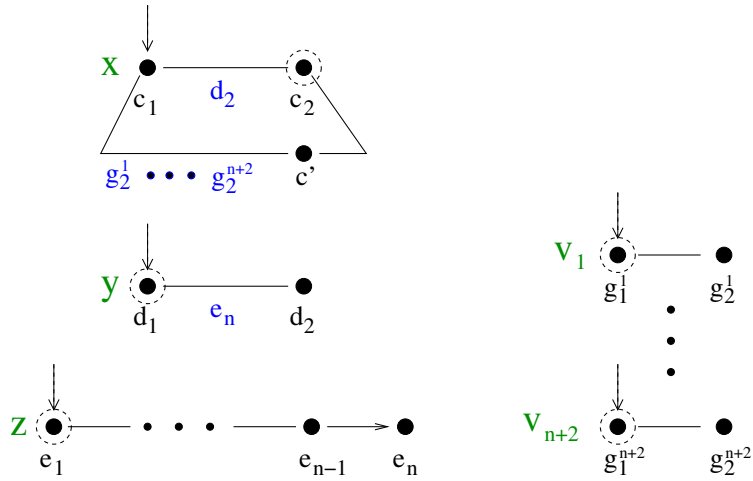

Figure 11: Planning task underlying Example 8. The arrow between $e_{n-1}$ and $e_n$ indicates that the respective DTG transition is directed, i.e., there exists no transition from $e_n$ to $e_{n-1}$.

*There are two ways to achieve the goal $c_2$: either via moving $y$ and $z$, or by moving $v_1, \ldots, v_{n+2}$. The only optimal relaxed plan chooses the former option, giving $h^+(s_I) = n+1$. As soon as $n \geq 3$, however, the only parallel-optimal relaxed plan $P^+(s_I)$ chooses the latter option because moving $y$ and $z$ results in $n + 1$ sequential moves, whereas $v_1, \ldots, v_{n+2}$ can be moved in parallel, giving parallel length 3.*

*Consider what happens to $h^+$ in either of the options. If we move $z$, then $h^+$ remains constant because we need to move $z$ back into its own goal. As soon as we reach $z = e_n$, $h^+ = \infty$ because the last transition is uni-directional and we can no longer achieve the own goal of $z$. Thus there is no exit path, and in particular no monotone exit path, via this option.*

*Say we move $v_1, \ldots, v_{n+2}$ instead. In the first move (whichever $v_i$ we choose), $h^+$ increases because the shortest option is to undo this move and go via $y$ and $z$: this takes $n+2$ steps whereas completing the $v_i$ moves and going via $c'$ takes $(n+1) + 2 = n+3$ steps.*

*Thus there is no monotone exit path via this option either, and $s_I$ is a local minimum. After completing the $n+2$ moves of $v_i$ and moving to $x = c'$, we have $h^+ = (n+2)+1$ due to the shortest relaxed plan that moves back all $v_i$ and moves to $x = c_2$. To reduce this heuristic value to the initial value $h^+(s_I) = n+1$, we need to execute a further 2 of these steps. The state we have then reached has a better evaluated neighbor, so the exit distance is $n+5$.*

*Consider now the effect of feeding Theorem 2 with the parallel-optimal plan $P^+(s_I)$. Clearly, the optimal rplan dependency graph $oDG^+$ constructed for $P^+(s_I)$ consists of $x$ and all the $v_i$ variables, but does not include $y$ nor $z$. Thus the theorem applies, and it wrongly concludes that $s_I$ is not a local minimum. The exit distance bound computed is $(-1) + \text{cost}^{d*}(oDG^+) = (-1) + 1[\text{cost}^{d*}(x)] + \sum_{i=1}^{n+2}(1*1)[\text{cost}^{d*}(x) * \text{diam}(DTG_{v_i})] = n+2$. This is less than the actual distance $ed(s_I) = n+5$, and thus this result is also wrong.*

Say we modify Example 8 by making the last transition of $z$ undirected, but making one of the $v_i$ transitions unidirectional to the right. Then the $v_1, \ldots, v_{n+2}$ option leads into a dead end, whereas the $y, z$ option succeeds. In particular, Theorem 2 does not apply to $oDG^+$ constructed for the parallel-optimal relaxed plan $P^+(s_I)$, and thus this is an example where using non-optimal relaxed plans results in a loss of information.

### A.5 Benchmark Performance Guarantees

We give definitions of the 7 domains mentioned in Propositions 1 and 2. For each domain, we explain why the respective property claimed holds true. In most of the domains, we assume some static properties as are used in PDDL to capture unchanging things like the shape of the road network in a transportation domain. We assume in what follows that such static predicates have been removed prior to the analysis, i.e., prior to testing the prerequisites of Theorem 4.

**Definition 5.** *The Logistics domain is the set of all planning tasks $\Pi = (\mathcal{V}, \mathcal{O}, s_I, s_G)$ whose components are defined as follows. $\mathcal{V} = P \cup V$ where $P$ is a set of "package-location" variables $p$, with $\mathcal{D}_p = L \cup V$ where $L$ is some set representing all possible locations, and $V$ is a set of "vehicle-location" variables $v$, with $\mathcal{D}_v = L_v$ for a subset $L_v \subseteq L$ of locations. $\mathcal{O}$ contains three types of operators: "move", "load", and "unload", where $move(v, l1, l2) = (\{v = l1\}, \{v = l2\})$ for $l1 \neq l2$, $load(v, l, p) = (\{v = l, p = l\}, \{p = v\})$, and $unload(v, l, p) = (\{v = l, p = v\}, \{p = l\})$. $s_I$ assigns an arbitrary value to each of the variables, and $s_G$ assigns an arbitrary value to some subset of the variables.*

Every global dependency graph $gDG$ in Logistics either has a package $p$ as the leaf variable $x_0$, or has a vehicle variable $v$ as the leaf variable $x_0$. In the latter case $gDG$ consists of only $x_0$, with no arcs. In the former case, $o_0$ is preconditioned on a single vehicle $v$ only, leading to a single non-leaf variable $v$. In both cases, $gDG$ is acyclic, all involved transitions have no side effects, and all involved transitions are invertible. Thus we can apply Theorem 4. We have $\text{cost}^{D*}(gDG) = 1 + 1 * 1$ for packages and $\text{cost}^{D*}(gDG) = 1$ for vehicles, thus overall we obtain the correct bound 1.

**Definition 6.** *The Miconic-STRIPS domain is the set of all planning tasks $\Pi = (\mathcal{V}, \mathcal{O}, s_I, s_G)$ whose components are defined as follows. $\mathcal{V} = O \cup D \cup B \cup S \cup \{e\}$ where $|O| = |D| = |B| = |S|$ and: $O$ is a set of "passenger-origin" variables $o$, with $\mathcal{D}_o = L$ where $L$* *is some set representing all possible locations (floors); $D$ is a set of "passenger-destination" variables $d$ with $\mathcal{D}_d = L$; $B$ is a set of "passenger-boarded" variables $b$ with $\mathcal{D}_b = \{1,0\}$; $S$ is a set of "passenger-served" variables $s$ with $\mathcal{D}_s = \{1,0\}$; $e$ is the "elevator-location" variable with $\mathcal{D}_e = L$. $\mathcal{O}$ contains three types of operators: "move", "board", and "depart", where $move(l1,l2) = (\{e = l1\}, \{e = l2\})$ for $l1 \neq l2$, $board(l,i) = (\{e = l, o_i = l\}, \{b_i = 1\})$, and $depart(l,i) = (\{e = l, d_i = l, b_i = 1\}, \{b_i = 0, s_i = 1\})$. $s_I$ assigns arbitrary locations to the variables $O$, $D$, and $e$, and assigns $0$ to the variables $B$ and $S$. $s_G$ assigns $1$ to the variables $S$.*

Passenger-origin and passenger-destination variables are static, i.e., not affected by any operator. Thus the common pre-processes will remove these variables, using them only to statically prune the set of operators that are reachable. We assume in what follows that such removal has taken place.

Every global dependency graph $gDG$ in Miconic-STRIPS has a passenger-served variable $s_i$ as the leaf variable $x_0$. This leads to non-leaf variables $b_i$ and $e$, with arcs from $e$ to both other variables and from $b_i$ to $s_i$. Clearly, $gDG$ is acyclic. The transitions of $e$ are all invertible and have no side effects. The transition $(0,1)$ of $b_i$ (is not invertible since departing has a different condition on $e$ but) has an irrelevant own-delete – $b_i = 0$ does not occur anywhere in the goal or preconditions – and has no side effects and thus irrelevant side effect deletes. The transition $(1,0)$ of $b_i$ (is not invertible but) is irrelevant – $b_i = 0$ doesn't occur anywhere. The transition $(0,1)$ of the leaf variable $s_i$ has self-irrelevant side effect deletes – $b_i = 1$ occurs only in the precondition of the transition's own responsible operator $rop(0,1) = depart(l_d, i)$. Hence we can apply Theorem 4. This delivers the bound $\text{cost}^{\text{D}*}(gDG) - 1 = -1 + 1[s_i] + (1*1)[\text{cost}^{\text{D}*}(s_i) * \max\text{Path}(DTG_{b_i})] + (2*1)[(\text{cost}^{\text{D}*}(s_i) + \text{cost}^{\text{D}*}(b_i)) * \text{diam}(DTG_e)] = 3$.

**Definition 7.** *The Simple-TSP domain is the set of all planning tasks $\Pi = (\mathcal{V}, \mathcal{O}, s_I, s_G)$ whose components are defined as follows. $\mathcal{V} = \{p\} \cup V$ where: $p$ is the "position" variable, with $\mathcal{D}_p = L$ where $L$ is some set representing all possible locations; and $V$, with $|V| = |L|$, is a set of "location-visited" variables $v$, with $\mathcal{D}_v = \{1,0\}$. $\mathcal{O}$ contains a single type of operators: $move(l1,l2) = (\{p = l1\}, \{p = l2, v_{l2} = 1\})$ for $l1 \neq l2$. $s_I$ assigns an arbitrary value to $p$ and assigns $0$ to the variables $V$. $s_G$ assigns $1$ to the variables $V$.*

Every global dependency graph $gDG$ in Simple-TSP has a location-visited variable $v_i$ as the leaf variable $x_0$. This leads to the single non-leaf variable $p$. Clearly, $gDG$ is acyclic. Every transition $(0,1)$ of $v_i$ considered, induced by $o_0 = move(l1, li)$, has replaceable side effect deletes. Any operator $o = move(l1, x)$ can be replaced by the equivalent operator $move(li, x)$ unless $x = li$. In the latter case, we have $o_0 = o$ which is excluded in the definition of replaceable side effect deletes. Every transition $(l1, l2)$ of $p$ clearly is invertible; it has the irrelevant side effect delete $v_{l2} = 0$; its side effect is only on $v_{l2}$ which is not a non-leaf variable of $gDG$. Hence we can apply Theorem 4. This delivers the bound $\text{cost}^{\text{D}*}(gDG) - 1 = -1 + 1[v_i] + (1*1)[\text{cost}^{\text{D}}(v_i) * \text{diam}(DTG_p)] = 1$.

We consider an extended version of the Movie domain, in the sense that, whereas the original domain version considers only a fixed range of snacks (and thus the state space is constant across all domain instances), we allow to scale the number of different snacks.[25]

---

25. The original domain version allows to scale the number of operators adding the same snack. All these operators are identical, and can be removed by trivial pre-processes.

**Definition 8.** *The Movie domain is the set of all planning tasks* $\Pi = (\mathcal{V}, \mathcal{O}, s_I, s_G)$ *whose components are defined as follows.* $\mathcal{V} = \{c0, c2, re\} \cup H$. *Here,* $c0$ *is the "counter-at-zero" variable, with* $\mathcal{D}_{c0} = \{1, 0\}$; $c2$ *is the "counter-at-two-hours" variable, with* $\mathcal{D}_{c2} = \{1, 0\}$; $re$ *is the "movie-rewound" variable, with* $\mathcal{D}_{re} = \{1, 0\}$; $H$ *are "have-snack" variables* $h$ *with* $\mathcal{D}_h = \{1, 0\}$. $\mathcal{O}$ *contains four types of operators: "rewindTwo", "rewindOther", "resetCounter", and "getSnack", where* $rewindTwo = (\{c2 = 1\}, \{re = 1\})$, $rewindOther = (\{c2 = 0\}, \{re = 1, c0 = 0\})$, $resetCounter = (\emptyset, \{c0 = 1\})$, *and* $getSnack(i) = (\emptyset, \{h_i = 1\})$. $s_I$ *assigns an arbitrary value to all variables.* $s_G$ *assigns the* $re$, $c0$, *and* $H$ *variables to 1.*

Note that, depending on the value of the static variable $c2$, the operator set will be different: if $s_I(c2) = 1$ then $rewindOther$ is removed, if $s_I(c2) = 0$ then $rewindTwo$ is removed. We refer to the former as case (a) and to the latter as case (b).

Every global dependency graph $gDG$ consists of a single (leaf) variable. The transitions of each $h$ variable have no side effects and thus have irrelevant side effect deletes. The transition $(0, 1)$ of $c0$ has no side effects and thus has irrelevant side effect deletes. The transition $(1, 0)$ of $c0$ is irrelevant. For case (a), the transition $(0, 1)$ of $re$ has no side effects and thus has irrelevant side effect deletes so we can apply Theorem 4. For case (b), the transition $(0, 1)$ of $re$ has the side effect $c0 = 0$. Observe that (1) this fact itself is irrelevant; and (2) that the only $\psi \in \text{ctx}(0, 1)$ is $\{c0 = 1\}$, and $o := resetCounter$ satisfies $\emptyset = \text{pre}_o \subseteq (\text{prev}_{\text{rop}(0,1)} \cup \text{eff}_{\text{rop}(0,1)}) = \{re = 1, c0 = 0\}$, $\{c0 = 1\} = \text{eff}_o \subseteq \psi = \{c0 = 1\}$, and $\{c0 = 1\} = \text{eff}_o \supseteq \{(y, d) \mid (y, d) \in \psi, (y, d) \in s_G \cup \bigcup_{\text{rop}(c,c') \neq o' \in O} \text{pre}_{o'}\} = \{c0 = 1\}$. Thus the transition has recoverable side effect deletes, and again we can apply Theorem 4. In case (a), for all $gDG$s the bound $\text{cost}^D(gDG) - 1$ applies. Obviously, $\text{cost}^D(gDG) = 1$ and thus we obtain the correct bound 0. In case (b), the bound $\text{cost}^D(gDG)$ applies, and again $\text{cost}^D(gDG) = 1$ so we obtain the correct bound 1.

**Definition 9.** *The Ferry domain is the set of all planning tasks* $\Pi = (\mathcal{V}, \mathcal{O}, s_I, s_G)$ *whose components are defined as follows.* $\mathcal{V} = C \cup \{f, e\}$ *where:* $C$ *is a set of "car-location" variables* $c$, *with* $\mathcal{D}_c = L \cup \{f\}$ *where* $L$ *is some set representing all possible locations;* $f$ *is the "ferry-location" variable with* $\mathcal{D}_f = L$; $e$ *is the "ferry-empty" variable with* $\mathcal{D}_e = \{1, 0\}$. $\mathcal{O}$ *contains three types of operators: "sail", "board", and "debark", where* $sail(l1, l2) = (\{f = l1\}, \{f = l2\})$ *for* $l1 \neq l2$, $board(l, c) = (\{f = l, c = l, e = 1\}, \{c = f, e = 0\})$, *and* $debark(l, c) = (\{f = l, c = f\}, \{c = l, e = 1\})$. $s_I$ *assigns 1 to variable* $e$, *assigns an arbitrary value to variable* $f$, *and assigns an arbitrary value other than* $f$ *to the variables* $C$. $s_G$ *assigns an arbitrary value* $\neq f$ *to (some subset of) the variables* $C$ *and* $f$.

Let $s$ be an arbitrary reachable state where $0 < h^+(s) < \infty$, and let $P^+(s)$ be an arbitrary optimal relaxed plan for $s$. Then we can always apply Theorem 2. To show this, we distinguish three cases: (a) $s(e) = 1$, $o_0 = board(l, c)$ is the first board operator in $P^+(s)$, and we set $x_0 = c$; (b) $s(e) = 0$, $o_0 = debark(l, c)$ is the first debark operator in $P^+(s)$, and we set $x_0 = c$; (c) $P^+(s)$ contains no board or debark operator and we set $o_0$ to be the first operator, $sail(l1, l2)$, in $P^+(s)$, with $x_0 = f$. Obviously, exactly one of these cases will hold in $s$. Let $oDG^+ = (V, A)$ be the sub-graph of $SG$ including $x_0$ and the variables/arcs included as per Definition 1. Let $t_0$ be the transition taken by $o_0$.

In case (a), obviously we can reorder $P^+(s)$ so that either $board(l, c)$ is the first operator in $P^+(s)$, or all its predecessors are $sail$ operators. $oDG^+$ then either (1) includes no new (non-leaf) variables at all, or (2) includes only $f$. As for $f$, clearly all its transitions are invertible and have no side effects. The transition $t_0$ has the own effect $(c, f)$ deleting $(c, l)$ which clearly is not needed in the rest of $P^+(s)$. It has the side effect $e = 0$ deleting $e = 1$. That latter fact may be needed by other board operators in $P^+(s)$. However, necessarily $P^+(s)$ contains an operator of the form $debark(l', c)$, which is applicable after $board(l, c)$ and a sequence of moves that $P^+(s)$ must contain from $l$ to $l'$; $debark(l', c)$ recovers $e = 1$. Thus the $oDG^+$-relevant deletes of $t_0$ are $P_{>0}^+(s)$-recoverable. In case (b), similarly we can reorder $P^+(s)$ so that either (1) $debark(l, c)$ is the first operator in $P^+(s)$, or (2) all its predecessors are $sail$ operators. The transition $t_0$ has the own effect $(c, l)$ deleting $(c, f)$ which clearly is not needed in the rest of $P^+(s)$; it has the side effect $e = 1$ deleting $e = 0$ which clearly is not needed in the rest of $P^+(s)$. Thus, again, the $oDG^+$-relevant deletes of $t_0$ are $P_{>0}^+(s)$-recoverable (the recovering sub-sequence of $P_{>0}^+(s)$ being empty because no recovery is required). In case (c), finally, $oDG^+$ contains only $f$, $t_0$ has no side effects, and its own delete $(f, l1)$ is not needed anymore (in fact, in this case $l2$ must be the goal for $f$, and $P^+(s)$ contains only the single operator $o_0$). Hence, in all cases, we can apply Theorem 2. $\mathrm{cost}^{\mathrm{d}*}(oDG^+) = 1$ in cases (a1), (b1), and (c) so there we get the bound 0. $\mathrm{cost}^{\mathrm{d}*}(oDG^+) = 1 + \mathrm{diam}(DTG_f) = 2$ in cases (a2) and (b2) so there we get the bound 1.

**Definition 10.** *The Gripper domain is the set of all planning tasks* $\Pi = (\mathcal{V}, \mathcal{O}, s_I, s_G)$ *whose components are defined as follows.* $\mathcal{V} = \{ro, f_1, f_2\} \cup B$. *Here,* $ro$ *is the "robot-location" variable, with* $\mathcal{D}_{ro} = \{L, R\}$; $f_1, f_2$ *are "gripper-free" variables, with* $D_{f_1} = D_{f_2} = \{1, 0\}$; *and* $B$ *are "ball-location" variables, with* $\mathcal{D}_b = \{L, R, 1, 2\}$. $\mathcal{O}$ *contains three types of operators: "move", "pickup", and "drop", where* $move(l1, l2) = (\{ro = l1\}, \{ro = l2\})$ *for* $l1 \neq l2$, $pickup(g, b, l) = (\{ro = l, b = l, f_g = 1\}, \{b = g, f_g = 0\})$, *and* $drop(g, b, l) = (\{ro = l, b = g\}, \{b = l, f_g = 1\})$. $s_I$ *assigns* $L$ *to* $ro$, *assigns* $1$ *to* $f_1$ *and* $f_2$, *and assigns* $L$ *to the variables* $B$. $s_G$ *assigns* $R$ *to the variables* $B$.

Let $s$ be an arbitrary reachable state where $0 < h^+(s) < \infty$, and let $P^+(s)$ be an arbitrary optimal relaxed plan for $s$. Then we can always apply Theorem 2. We distinguish two cases: (a) there exists $b \in B$ so that $s(b) = g$ for $g \in \{1, 2\}$, $o_0 = drop(g, b, R)$, and we set $x_0 = b$; (b) there exists no $b \in B$ so that $s(b) = g$ for $g \in \{1, 2\}$, $o_0 = pickup(g, b, L)$ for some $b \in B$ is in $P^+(s)$, and we set $x_0 = b$. Obviously, exactly one of these cases will hold in $s$. Let $oDG^+ = (V, A)$ be the sub-graph of $SG$ including $x_0$ and the variables/arcs included as per Definition 1. Let $t_0$ be the transition taken by $o_0$.

In case (a), obviously we can reorder $P^+(s)$ so that either $drop(g, b, R)$ is the first operator in $P^+(s)$, or its only predecessor is $move(L, R)$. $oDG^+$ then either (1) includes no new (non-leaf) variables at all, or (2) includes only $ro$. As for $ro$, clearly all its transitions are invertible and have no side effects. The transition $t_0$ has the own effect $(b, R)$ deleting $(b, g)$ which clearly is not needed in the rest of $P^+(s)$; it has the side effect $f_g = 1$ deleting $f_g = 0$ which clearly is not needed in the rest of $P^+(s)$. Thus the $oDG^+$-relevant deletes of $t_0$ are $P_{>0}^+(s)$-recoverable. In case (b), similarly we can reorder $P^+(s)$ so that either (1) $pickup(g, b, L)$ is the first operator in $P^+(s)$, or (2) its only predecessor is $move(R, L)$. The transition $t_0$ has the own effect $(b, g)$ deleting $(b, L)$ which clearly is not needed in the rest of $P^+(s)$. It has the side effect $f_g = 0$ deleting $f_g = 1$; that latter fact may be needed by other pickup operators in $P^+(s)$. However, necessarily $P^+(s)$ contains the operators $move(L, R)$ and $drop(g, b, R)$, which are applicable after $board(l, c)$; $drop(g, b, R)$ recovers $f_g = 1$. Thus, again, the $oDG^+$-relevant deletes of $t_0$ are $P_{>0}^+(s)$-recoverable. Hence, in both cases, we can apply Theorem 2. $\text{cost}^{\text{d}*}(oDG^+) = 1$ in cases (a1) and (b1), so there we get the bound 0. $\text{cost}^{\text{d}*}(oDG^+) = 1 + \text{diam}(ro) = 2$ in cases (a2) and (b2) so there we get the bound 1.

**Definition 11.** *The Transport domain is the set of all planning tasks* $\Pi = (\mathcal{V}, \mathcal{O}, s_I, s_G)$ *whose components are defined as follows.* $\mathcal{V} = P \cup VE \cup C$ *where:* $P$ *is a set of "package-location" variables* $p$, *with* $\mathcal{D}_p = L \cup VE$ *where* $L$ *is some set representing all possible locations;* $VE$ *is a set of "vehicle-location" variables* $v$, *with* $\mathcal{D}_v = L$; *and* $C$ *is a set of "vehicle-capacity" variables* $c_v$, *with* $\mathcal{D}_{c_v} = \{0, \ldots, K\}$ *where* $K$ *is the maximum capacity.* $\mathcal{O}$ *contains three types of operators: "drive", "pickup", and "drop", where:* $drive(v, l1, l2) = (\{v = l1\}, \{v = l2\})$ *for* $(l1, l2) \in R$ *where* $G^R = (L, R)$ *is an undirected graph of roads over* $L$; *pickup*$(v, l, p, c) = (\{v = l, p = l, c_v = c\}, \{p = v, c_v = c - 1\})$, *and* $drop(v, l, p, c) = (\{v = l, p = v, c_v = c\}, \{p = l, c_v = c + 1\})$. $s_I$ *assigns an arbitrary value in* $L$ *to each of the variables* $P \cup VE$, *and assigns* $K$ *to the variables* $C$. $s_G$ *assigns an arbitrary value in* $L$ *to some subset of the variables* $P \cup VE$.

Note here the use of numbers and addition/subtraction. These are, of course, not part of the planning language we consider here. However, they can be easily encoded (on the finite set of number $\{0, \ldots, K\}$) via static predicates. After pre-processing, in effect the resulting task will be isomorphic to the one obtained by the simple arithmetic above, which we thus choose to reduce notational clutter.

Let $s$ be an arbitrary reachable state where $0 < h^+(s) < \infty$. Then there exists an optimal relaxed plan $P^+(s)$ for $s$ so that we can apply Theorem 2. We distinguish three cases: (a) there exists $p \in P$ so that $s(p) = v$ for $v \in VE$, $o_0 = drop(v, l, p, c)$ where $s(c_v) = c$ is in $P^+(s)$, and we set $x_0 = p$; (b) there exists no $p \in P$ so that $s(p) = v$ for $v \in VE$, $o_0 = pickup(v, l, p, K)$ for some $p \in P$ is in $P^+(s)$, and we set $x_0 = p$; (c) $P^+(s)$ contains no drop or pickup operator and we set $o_0$ to be the first operator, $drive(v, l1, l2)$, in $P^+(s)$, with $x_0 = v$. Obviously, we can choose $P^+(s)$ so that exactly one of these cases will hold in $s$ (the choice of $P^+(s)$ is arbitrary for (b) and (c), but in (a) there may exist optimal relaxed plans where $s(c_v) \neq c$). Let $oDG^+ = (V, A)$ be the sub-graph of $SG$ including $x_0$ and the variables/arcs included as per Definition 1. Let $t_0$ be the transition taken by $o_0$.

In case (a), obviously we can reorder $P^+(s)$ so that either $o_0 = drop(v, l, p, c)$ is the first operator in $P^+(s)$, or all its predecessors are $drive$ operators. $oDG^+$ then either (1) includes no new (non-leaf) variables at all, or (2) includes only $v$. As for $v$, clearly all its transitions are invertible and have no side effects. The transition $t_0$ has the own effect $(p, v)$ deleting $(p, l)$ which clearly is not needed in the rest of $P^+(s)$. It has the side effect $c_v = c+1$ deleting $c_v = c$. That latter fact may be needed by other operators in $P^+(s)$, either taking the form $drop(v, l', p', c)$ or the form $pickup(v, l', p', c)$. Clearly, if $P^+(s)$ contains these operators then we can replace them with $drop(v, l', p', c + 1)$ and $pickup(v, l', p', c + 1)$ respectively – the value $(c_v, c + 1)$ will be true at their point of (relaxed) execution. Thus we can choose $P^+(s)$ so that the $P^+(s)$-relevant deletes of $t_0$ are $P^+(s)$-recoverable on $V \setminus \{x_0\}$. In case (b), similarly we can reorder $P^+(s)$ so that either (1) $o_0 = pickup(v, l, p, K)$ is the first operator in $P^+(s)$, or (2) all its predecessors are $drive$ operators. The transition $t_0$ has the own effect $(p, v)$ deleting $(p, l)$ which clearly is not needed in the rest of $P^+(s)$. It has the side effect $c_v = K - 1$ deleting $c_v = K$. That latter fact may be needed by other operators in $P^+(s)$, taking the form $pickup(v, l', p', K)$. However, necessarily $P^+(s)$ contains an operator of the form $drop(v, l', p, c')$. If $c' \neq K - 1$ then we can replace this operator with $drop(v, l', p, K - 1)$ since, clearly, the value $(c_v, K - 1)$ will be true at the point of (relaxed) execution. Now, $drop(v, l', p, K - 1)$ is applicable after $pickup(v, l, p, K)$ and a sequence of drive operators that $P^+(s)$ must contain from $l$ to $l'$; $drop(v, l', p, K - 1)$ recovers $c_v = K$. Thus, again, we can choose $P^+(s)$ so that the $P^+(s)$-relevant deletes of $t_0$ are $P^+(s)$-recoverable on $V \setminus \{x_0\}$. In case (c), finally, $oDG^+$ contains only $v$, $t_0$ has no side effects, and its own delete $(v, l1)$ is not needed anymore. Hence, in all cases, we can apply Theorem 2. $\text{cost}^{\text{d*}}(oDG^+) = 1$ in cases (a1), (b1), and (c) so there we get the bound 0. $\text{cost}^{\text{d*}}(oDG^+) = 1 + \min(\text{diam}(oDTG_v^+), \text{diam}(DTG_v))$ in cases (a2) and (b2) so there the bound is at most the diameter of the road map $G^R$.

When ignoring action costs, the Elevators domain of IPC 2008 is essentially a variant of Transport. The variant is more general in that (a) each vehicle (each elevator) may have its own maximal capacity, and (b) each vehicle can reach only a subset of the locations, i.e., each vehicle has an individual road map. On the other hand, Elevators is more restricted than Transport in that (c) each vehicle road map is fully connected (every reachable floor can be navigated to directly from every other reachable floor), and (d) goals exist only for packages (passengers, that is), not for vehicles. Even when ignoring restrictions (c) and (d), it is trivial to see that the arguments given above for Transport still hold true. Therefore, whenever $s$ is a reachable state with $0 < h^+(s) < \infty$, there exists an optimal relaxed plan $P^+(s)$ for $s$ so that we can apply Theorem 2. As before, the bound is at most the diameter of the road map. Due to (c), this diameter is 1.